\documentclass[10pt,twocolumn,letterpaper]{article}

\usepackage{iccv}
\usepackage{times}
\usepackage{epsfig}
\usepackage{graphicx}
\usepackage{amsmath}
\usepackage{amssymb}
\usepackage{algorithm}
\usepackage{algorithmic}
\usepackage{caption}

\usepackage[pagebackref=true,breaklinks=true,letterpaper=true,colorlinks,bookmarks=false]{hyperref}

\iccvfinalcopy 

\def\iccvPaperID{5864} 

\ificcvfinal\fi

\begin{document}

\title{Out-of-domain GAN inversion via Invertibility Decomposition for Photo-Realistic Human Face Manipulation}


\author{Xin Yang$^{1, 2, 3}$ \quad Xiaogang Xu$^{4, 5}$ \quad Yingcong Chen$^{1, 2, 3}$\footnotemark[1]\\
$^1$ HKUST (Guangzhou) \quad
$^2$ HKUST  \quad 
$^3$ HKUST (Guangzhou) - SmartMore Joint Lab  \quad  \\
$^4$ CUHK \quad $^5$ SmartMore\\
 {\tt \small xin.yang@connect.ust.hk} \quad {\tt \small xgxu@cse.cuhk.edu.hk}  \quad {\tt \small yingcongchen@ust.hk}
}

\twocolumn[{
\maketitle

\begin{figure}[H]
    \vspace{-4em}
    \hsize=\textwidth
    \centering
    \begin{tabular}{@{}c@{}c@{}c@{}c@{}c@{}c@{}}
        \resizebox{0.15\textwidth}{!}{~}&
        \resizebox{0.15\textwidth}{!}{~}&
        \resizebox{0.15\textwidth}{!}{~}&
        \scriptsize{Input  \textbf{$\rightarrow$ +Smile}} & 
        \scriptsize{(a) HyperStyle~\cite{alaluf2021HyperStyle}} &
        \scriptsize{(b) HFGI$_{e4e}$~\cite{wang2021high}}
        \\
        \multicolumn{6}{@{}c@{}}{\includegraphics[width=0.9\textwidth]{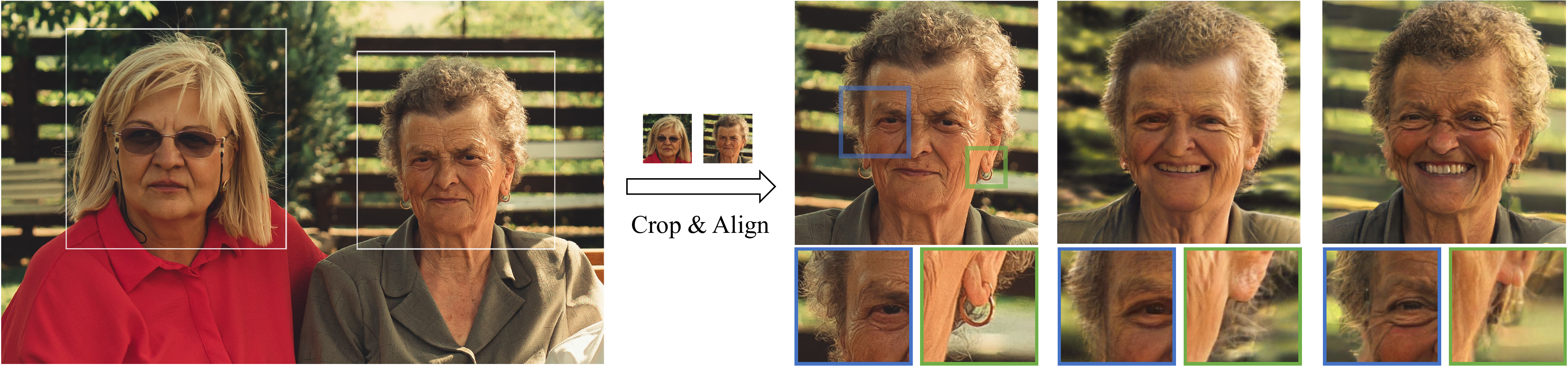}}
        \\
        \resizebox{0.15\textwidth}{!}{~}&
        \resizebox{0.15\textwidth}{!}{~}&
        \resizebox{0.15\textwidth}{!}{~}&
        \scriptsize{(c) Ours$_{e4e}$} & 
        \scriptsize{(d) SAM~\cite{parmar2022spatially}} &
        \scriptsize{(e) DiffCAM~\cite{song2022editing}}
        \\
        \multicolumn{6}{@{}c@{}}{\includegraphics[width=0.9\textwidth]{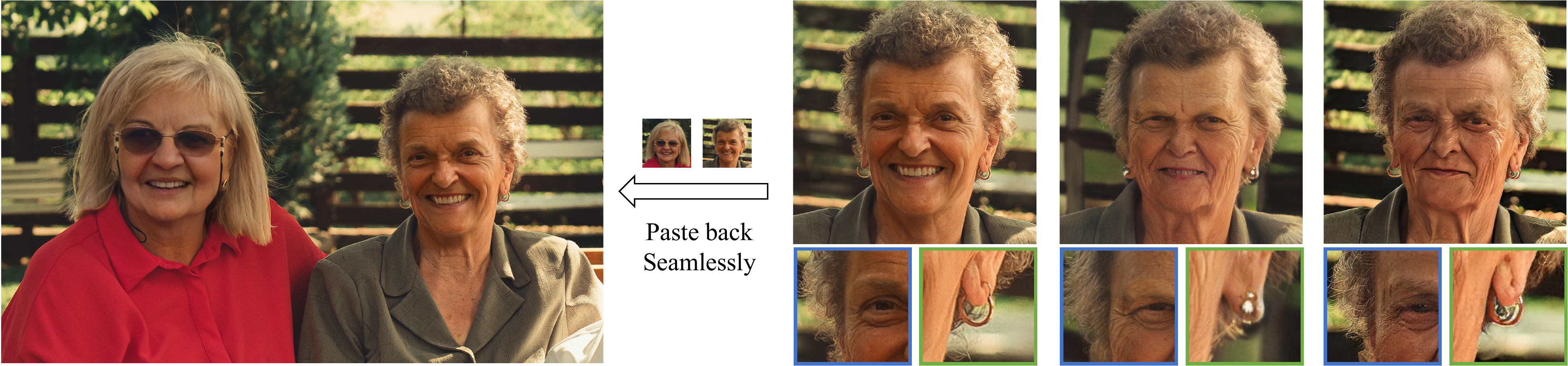}}
        \\
        \resizebox{0.15\textwidth}{!}{~}&
        \resizebox{0.15\textwidth}{!}{~}&
        \resizebox{0.15\textwidth}{!}{~}&
        \resizebox{0.15\textwidth}{!}{~}& 
        \resizebox{0.15\textwidth}{!}{~}&
        \resizebox{0.15\textwidth}{!}{~}
        \vspace{-1.7em}
    \end{tabular}
    \caption{\textbf{Out-of-domain GAN inversion and attribute manipulation.} Our method maintains out-of-domain objects in the input image (e.g., background, earrings) and generates high-fidelity attribute-edited results. Please zoom in for detail.}
    \label{fig:teaser}
    \vspace{-0.3em}
\end{figure}
}]

\ificcvfinal\thispagestyle{empty}\fi


\begin{abstract}
The fidelity of Generative Adversarial Networks (GAN) inversion is impeded by Out-Of-Domain (OOD) areas (e.g., background, accessories) in the image.
Detecting the OOD areas beyond the generation ability of the pre-trained model and blending these regions with the input image can enhance fidelity.
The ``invertibility mask" figures out these OOD areas, and existing methods predict the mask with the reconstruction error. However, the estimated mask is usually inaccurate due to the influence of the reconstruction error in the In-Domain (ID) area.
In this paper, we propose a novel framework that enhances the fidelity of human face inversion by designing a new module to decompose the input images to ID and OOD partitions with invertibility masks.
Unlike previous works, our invertibility detector is simultaneously learned with a spatial alignment module. We iteratively align the generated features to the input geometry and reduce the reconstruction error in the ID regions. Thus, the OOD areas are more distinguishable and can be precisely predicted.
Then, we improve the fidelity of our results by blending the OOD areas from the input image with the ID GAN inversion results.
Our method produces photo-realistic results for real-world human face image inversion and manipulation. 
Extensive experiments demonstrate our method's superiority over existing methods in the quality of GAN inversion and attribute manipulation. 
Our code is available at: \hyperlink{https://github.com/AbnerVictor/OOD-GAN-inversion}{https://github.com/AbnerVictor/OOD-GAN-inversion}
\end{abstract}

\section{Introduction}
\label{sec:intro}
In recent years, there have been efforts to apply generative models, e.g., StyleGAN2~\cite{karras2020analyzing}, for face image editing~\cite{abdal2019image2stylegan,abdal2020image2stylegan++,wu2021stylespace,song2022editing, alaluf2021HyperStyle, wang2021high, parmar2022spatially, tov2021designing, alaluf2021restyle, richardson2021encoding, xuyao2022, roich2021pivotal, nitzan2022mystyle, wei2022e2style} and restoration~\cite{wang2021towards,he2022gcfsr,zhu2022blind,yang2021gan}.
The basis of these applications is the GAN inversion.
The typical inversion strategy~\cite{richardson2021encoding, tov2021designing} is to train encoders to encode the face images into the latent of the pre-trained generator and reconstruct the images via the generator.
However, such approaches can not result in artifact-free and precise reconstruction due to the inevitable information loss when translating the high-resolution image into the limited GAN latent space.

 Some improved methods were proposed to further finetune the generator for image-specific reconstruction without losing editability~\cite{roich2021pivotal,nitzan2022mystyle}. Although these finetuning-based techniques improve the inversion accuracy of identity and style, still, the generator can hardly handle the reconstruction of out-of-domain contents, e.g., the complex background, accessories, and hair.
Meanwhile, some works~\cite{wang2021high,wang2021towards, parmar2022spatially} strengthen the potential of out-of-domain inversion capability of pre-trained GAN models by modulating the generator features with the input features extracted from source images.
Such methods suffer from the fidelity-editability trade-off~\cite{wang2021high, shannon1959coding, tishby2015deep, he2022gcfsr, wang2021towards, yang2021gan} since the feature modulation operation breaks the GAN priors. The larger latent space increases the reconstruction quality but undermines the editability of the framework.

Furthermore, some recent works~\cite{song2022editing, parmar2022spatially} propose to disentangle the target image into different spatial areas. They refine the regions with low invertibility to improve the inversion fidelity. Such low invertibility regions are the parts that cannot be reconstructed well with the generator, so-called Out-Of-Domain (OOD) areas. 
E.g., Song et al.~\cite{song2022editing} estimate a manipulation-aware mask with an attribute classifier. However, they ignore the geometrical misalignment between the inverted and the original images and apply a deghosting module for result refinement, which leads to undesired artifacts in their results (Fig.~\ref{fig:teaser} (e)).
Parmar et al.~\cite{parmar2022spatially} train an invertibility mask prediction module with perceptual supervision. The invertibility masks are then used as guidance for GAN feature modulation. However, their prediction of invertibility is noisy and inconsistent with the face semantic parts.
Therefore, they adopt a segmentation model for refinement, but the invertibility in the same semantic area (e.g., occlusions on the face) could be inconsistent. Also, they manually design thresholds to filter out the OOD areas, which is hard to optimize for small objects (Fig.~\ref{fig:teaser} (d)).

In summary, existing invertibility estimation methods mainly adopt the reconstruction error as the reference to judge the OOD regions. However, they ignore that reconstruction errors also come from the In-Domain (ID) areas. Consequently, their predicted mask is noisy and unreliable.

In this paper, we propose a novel strategy for photo-realistic GAN inversion by decomposing the input images into OOD and ID areas with invertibility masks. We focus on the high-resolution ($1024^2$ pixels) GAN inversion on human face images and the downstream applications (e.g., attribute editing).
Our basic idea is to reduce the reconstruction error of the ID areas and thus highlight the error of OOD regions. The reconstruction error of the ID area comes from both the textural and geometrical misalignment between the input image and the generated image. Although previous works improve the textural accuracy in the reconstruction by predicting or optimizing a better latent vector $w$, the geometrical misalignment is rarely discussed, which we believe is also important for invertibility estimation. Hence, we design an invertibility detector learned with an optical flow prediction module to reduce the influence of geometrical misalignment. The optical flow is computed between the features of the encoder and the generator, which is then applied to warp the generated features to alleviate their misalignment with the input features.
Compared with feature modulation~\cite{wang2021high, parmar2022spatially, xuyao2022}, such warping will not break the fidelity of the generated textures.
Along the training, the reconstruction error of the ID area will be minimized, and the invertibility mask prediction will be gradually focused on the OOD regions.
The overall procedure needs no extra labels for the mask or flows.

Based on the invertibility prediction, we design an effective approach to composite the generated content with the out-of-domain input feature for a photo-realistic generation with high fidelity.
Our framework consists of three major parts: the encoder, the Spatial Alignment and Masking Module (SAMM), and the generator. 
First, we extract features from the input image and predict its latent vector with a pre-trained image-to-latent encoder~\cite{tov2021designing, alaluf2021restyle}. Second, we feed the latent vector into a pre-trained StyleGAN2~\cite{karras2020analyzing} model for content generation, acquiring generated features.
Third, we estimate the optical flow and the invertibility mask between the input and the generated features at multiple resolutions. 
Then, we warp the generated features with the flow, aiming to minimize the reconstruction error of ID regions.
Finally, we composite the input image with the generated content according to the invertibility mask. 


Since only the spatial operation, i.e., warping, is applied to the generated features, we maintain their editability with existing GAN editing methods. 
Combined with the artifact-free and precise inversion effects, our method has excellent superiority in reconstruction accuracy and editing fidelity over existing approaches.
In this paper, we adopt StyleGAN2 as the backbone for experiments, and extensive experimental results demonstrate that our method outperforms current state-of-the-art methods with higher reconstruction fidelity and better visual quality.


In summary, our contributions are listed as follows:
\begin{enumerate}
    \item We propose a novel framework for out-of-domain GAN inversion on human face images by aligning and blending the generated image with the input image via optical flow and invertibility mask prediction.

    \item We investigate the GAN invertibility with a novel Spatial Alignment and Masking Module, which is a new solution for invertibility decomposition.
    
    \item Our proposed framework can produce photo-realistic results in both reconstruction and editing tasks. Experiments show that our framework outperforms existing methods in reconstruction accuracy and visual fidelity.
\end{enumerate}

\section{Related works}
\noindent\textbf{GAN inversion.} The process of GAN inversion involves encoding a real-world image into a semantic-disentangled latent space before reconstructing the image with a GAN generator.
It enables various downstream applications, e.g., face editing with labels~\cite{song2022editing, alaluf2021HyperStyle, wang2021high} or texts~\cite{patashnik2021styleclip,gal2022stylegan,abdal2022clip2stylegan}.

To tackle the non-trivial translation between the image and latent vector, some efforts have been made to design more suitable latent space and better encoders. Abdal et al.~\cite{abdal2019image2stylegan, abdal2020image2stylegan++} analyzes the extended GAN latent space ($W+$) for better inversion and attribute manipulation of real images. Based on the $W+$ latent space, later works further improve the inversion accuracy with hierarchical encoding~\cite{richardson2021encoding, hu2022style}, progressive training~\cite{tov2021designing} and iterative prediction strategies~\cite{alaluf2021restyle,wei2022e2style}.
More recently, Bai et al.~\cite{bai2022high} investigated the padding space of the StyleGAN generator to increase the invertibility of the pre-trained GAN model. Roich et al.~\cite{roich2021pivotal} and Nitzan et al.~\cite{nitzan2022mystyle} propose strategies to first encode images into latent vectors, then finetune the pre-trained generator for specific images or people. More recently, Alaluf et al.~\cite{alaluf2021HyperStyle} adopt a hypernetwork~\cite{ha2016hypernetworks} to modulate the generator kernel for better GAN inversion.
However, such methods can hardly invert the out-of-domain contents such as image-specific backgrounds or accessories.

\noindent\textbf{The distortion-editability trade-off.} To improve the fidelity of GAN inversion, some works~\cite{wang2021high, wang2021towards, xuyao2022, he2022gcfsr, zhu2022blind, Yang2021GPEN} add extra connections between the encoder and generator to further extend the latent space, but such methods fall into a tricky dilemma of balancing the distortion-editability tradeoff~\cite{wang2021high, shannon1959coding}. A larger latent space helps increase the reconstruction precision but decreases the editability in the generation due to the increasing semantic entanglement problem. Recently, Parmar et al.~\cite{parmar2022spatially} proposed the SAM with an invertibility prediction module guided by LPIPS~\cite{zhang2018perceptual} loss to predict the spatial invertibility map for the input image. Nevertheless, their invertibility predictions are noisy and inconsistent with the input image, which needs to be smoothed with a pre-trained segmentation network. 
Furthermore, since SAM is an optimization-based method, it takes a long inference time to produce a high-fidelity result. 
Moreover, Song et al.~\cite{song2022editing} propose DiffCAM to refocus the GAN inversion on the attributes to be manipulated. Instead of inversing the whole image, DiffCAM aims to find the edited area and blend the input image with the edited one. They adopt a semantic classifier (i.e., the facial attributes) as the prior for a rough editing mask prediction. However, due to the misalignment between the generated content and the input image, they suffer from undesired ghosting artifacts in their composition results. Thus, DiffCAM also needs to employ an extra ghosting removal module, while it can not fix the artifacts thoroughly.

Different from existing GAN inversion methods, we simultaneously improve the ID reconstruction via geometric alignment and OOD reconstruction via image blending with the invertibility mask.
In our experiments, we compare the baseline methods' inversion and attribute manipulation performance with our framework. Our method outperforms existing works in reconstruction quality (Fig.~\ref{fig:inversion}) and can produce photo-realistic face editing results (Fig.~\ref{fig:editing_compare}) with off-the-shelf GAN manipulation approaches~\cite{shen2020interpreting, patashnik2021styleclip, harkonen2020ganspace}.



\section{Method}
\begin{figure}[t]
    \centering
    \includegraphics[width=1.0\linewidth]{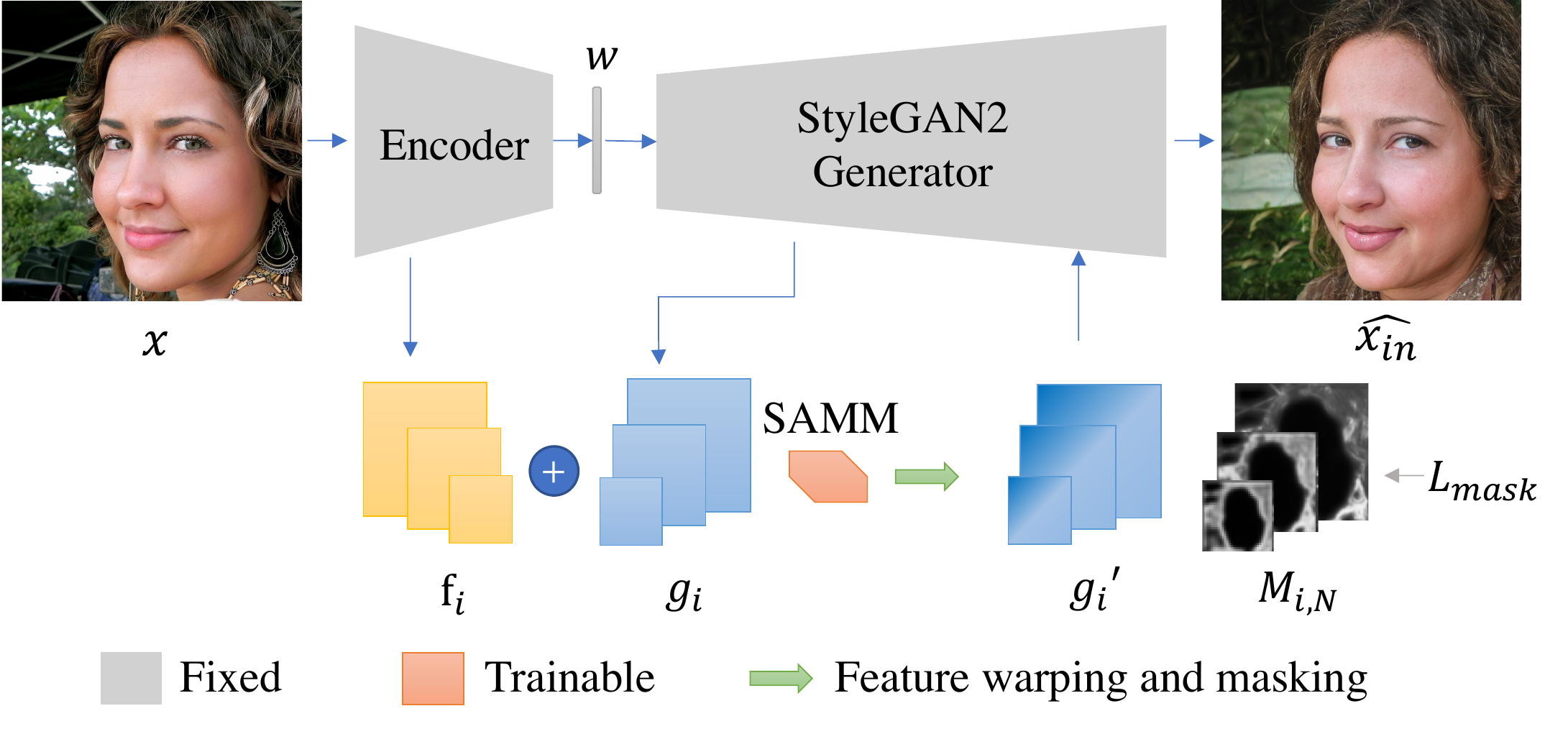}
    \vspace{-1em}
    \caption{\textbf{An overview of our framework.} Our framework begins by extracting features from the input image and aligning the generated features to the input features, which improves the in-domain reconstruction accuracy and eases the invertibility decomposition. Additionally, we predict the invertibility mask for feature and RGB space blending, which enhances the out-of-domain reconstruction quality. }
    \label{fig:overview}
    \vspace{-1em}
\end{figure}

In this paper, we propose a new framework for out-of-domain GAN inversion on human faces. The proposed framework can produce a precise invertibility mask to achieve excellent input and generated image composition.
The overview of our framework is shown in Fig.~\ref{fig:overview} and Fig.~\ref{fig:blending}. 
In the subsequent sections, we discuss the framework's overview in Sec.~\ref{sec:overview}, our proposed Spatial Alignment and Masking Module (SAMM) in Sec.~\ref{sec:SAMM}, the inversion and editing pipeline using our proposed modules in Sec.~\ref{sec:blending}, and our training strategy in Sec.~\ref{sec:training}.


\subsection{Overview}
\label{sec:overview}

\noindent\textbf{Traditional GAN inversion.} The essential part of our framework is the estimation of the GAN invertibility on a pre-trained StyleGAN2~\cite{karras2020analyzing}.
The StyleGAN2 generator is a multi-scale Convolutional-Neural-Network (CNN), where the results are generated gradually with enlarged spatial resolutions.
Meanwhile, the image generation is controlled by the latent vector $w$ in the generation process. Recent works~\cite{richardson2021encoding, tov2021designing, alaluf2021restyle, wang2021high, parmar2022spatially, song2022editing} adopt the pre-trained StyleGAN2 generator and aim to inverse the input RGB image into the well-disentangled $W^+ \in R^{512 \times 18}$ latent space proposed in \cite{richardson2021encoding} with an encoder $E$.
Given an input image $x \in R^{3 \times 1024 \times 1024}$, the GAN inversion of $x$ is to first encode $x$ into a latent vector $w \in W^+$ with the encoder $E$, then reconstruct the image $\hat{x}$ with the generator $G$, that:
\begin{equation}
    \hat{x} = G(E(x)).
    \label{eqn:gan_inversion}
\end{equation}

\noindent\textbf{GAN inversion with OOD decomposition.} Previous work~\cite{feng2021understanding} analyzed the GAN training strategy and discovered that $W$ space is a low-rank approximation for the target domain of generator $G$. 
Thus, it is difficult to reconstruct $\hat{x}$ without distortion. 
To solve this drawback, we tackle GAN inversion from the perspective of invertibility decomposition. Different from ~\cite{wang2021high, xuyao2022}, we intend to first decompose $x$ into the invertible partition $x_{in}$ and the OOD partition $x_{out}$ (e.g., accessories, tattoos) with an invertibility mask $m$, where
\begin{equation}
\label{eqn:decomp}
\begin{split}
    x_{out} = x \cdot m, ~~
    x_{in} = x \cdot (1-m).
\end{split}
\end{equation}
Then, we can reconstruct or edit $x_{in}$ via GAN inversion and preserve the OOD contents $x_{out}$ for better fidelity, that
\begin{gather}
    \label{eqn:decomp_gan_inversion}
    \widehat{x_{in}} =~ G(E(x)), \\
    \label{eqn:blending}
    \hat{x} =~ x_{out} + \widehat{x_{in}} \cdot (1 - m).
\end{gather}
The proposed framework consists of three major parts, i.e, the encoder $E$, the generator $G$, and SAMM. 
With the pre-trained and fixed $E$ and $G$, the SAMM estimates GAN invertibility by minimizing the reconstruction error of $\hat{x}$. 
At the training stage, we adopt the pre-trained image-to-latent encoder $E$ from \cite{tov2021designing} and train the SAMM for image-to-image reconstruction. 
Please refer to the following sections for details.

\subsection{Spatial Alignment and Masking Module}
\label{sec:SAMM}
\noindent\textbf{The limitation of previous invertibility estimation.} Previous methods for invertibility mask prediction mainly adopt the reconstruction error as the supervision~\cite{parmar2022spatially}. However, the predictions are usually inaccurate since the subtle ID reconstruction error, which should not be considered as the OOD content, disturbs the prediction of invertibility.
 
We find these ID errors come from the textural and geometrical misalignment between $x$ and $\widehat{x_{in}}$. While the textural misalignment (e.g., eyes or skin color) can be corrected via latent space optimization~\cite{parmar2022spatially, alaluf2021HyperStyle, alaluf2021restyle, nitzan2022mystyle, roich2021pivotal}, it is hard to reconstruct delicate local structures (e.g., the boundary of the face or hair) in $x$ with $w$, especially when the output resolution is high (e.g., $1024^2$ and above). In this paper, we mainly focus on minimizing the geometrical misalignment with spatial operations such as warping.




\begin{figure}[t]
    \centering
    \includegraphics[width=0.9\linewidth]{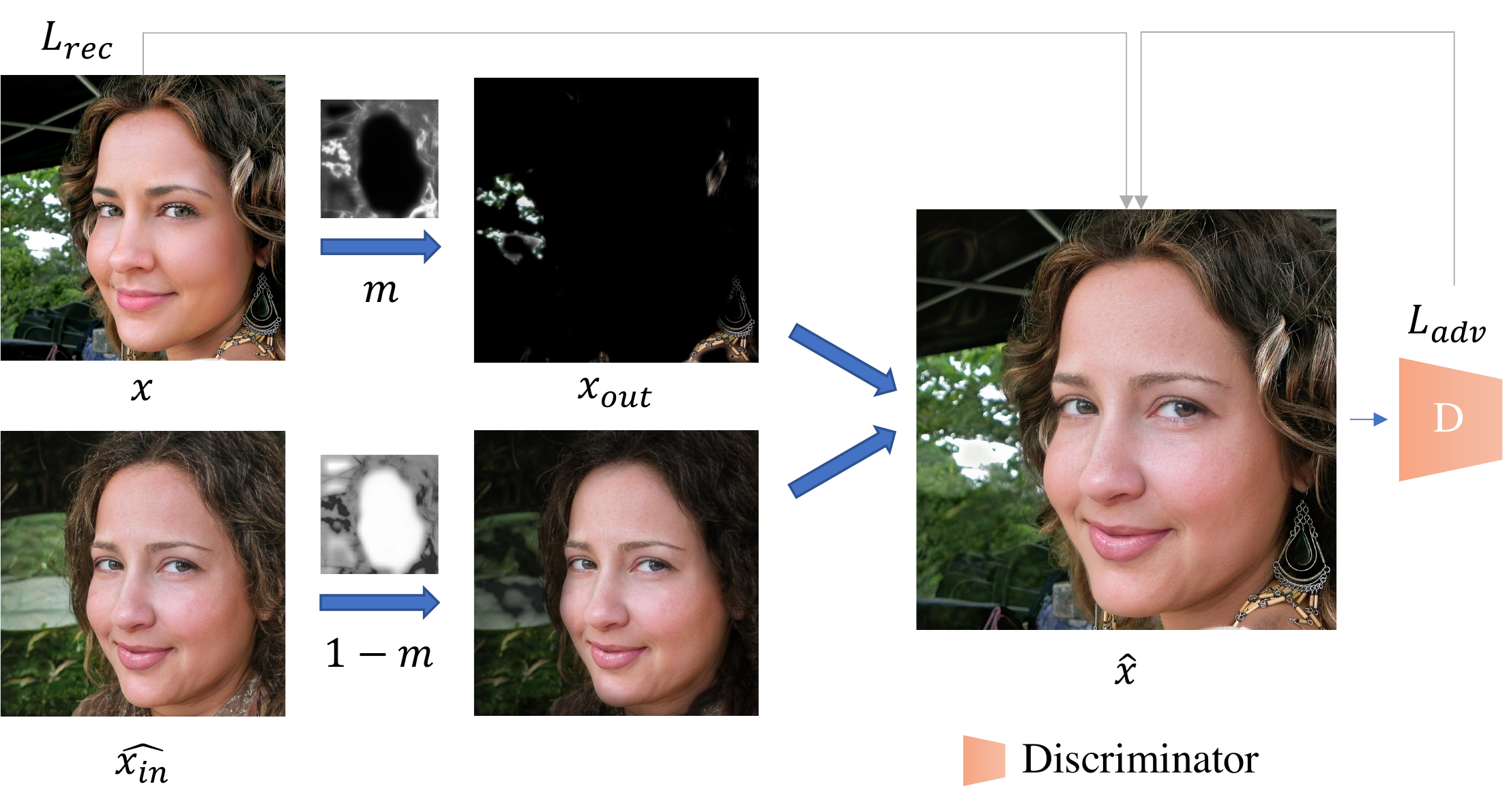}
    \vspace{-0.2em}
    \caption{\textbf{The invertibility decomposition.} With the predicted invertibility mask $m$, we blend the in-domain GAN inversion $\widehat{x_{in}}$ with the out-of-domain partition $x_{out}$ to produce the final result $\hat{x}$.}
    \label{fig:blending}
    \vspace{-1em}
\end{figure}

\noindent\textbf{Our SAMM for invertibility estimation.} 
Inspired by previous works done by Collins et al.~\cite{Collins20} and Chong et al.~\cite{chong2021stylegan} where they discovered that simple spatial operations in the GAN feature space maintain the realistic textures in the results.
We also find that because $w$ controls the generation by modulating the variance of the generated feature maps channel-wisely, simple spatial warping in the feature space of $G$ will not compromise the editability of the framework.
Thus, we train SAMM with an optical flow prediction module to predict the flow between $x$ and $\widehat{x_{in}}$ in the ID regions. Then, we warp the generated features with grid sampling ($\mathcal{GS}$) to reduce the geometrical error in $\widehat{x_{in}}$. 

\noindent\textbf{The details of SAMM.} 
First, we encode the input image into the latent vectors $w$ and the input features $f_i$ ($i \in \{1, ..., L\}$). 
Then, feature map $g_i$ is produced in the $i$-th layer of generation. During the generation, we feed both $g_i$ and $f_i$ into the SAMM for flow and mask prediction. Meanwhile, we do the prediction in an iterative manner ($N$ iterations). The pseudo-code of SAMM is shown in Algorithm~\ref{alg:samm}.

\begin{algorithm}
    \caption{The cycle optical flow and invertibility mask prediction with SAMM in the $i$-th layer of generation.}
    \label{alg:samm}
    \begin{algorithmic}[1]
        \STATE Initialization: $j = 1$, $g_{i, 0} = g_i$ and $\Delta_{i}^x, \Delta_{i}^y = 0$
        \FOR{$j = \{1, ..., N\}$}
            \STATE $\delta_{i}^x, \delta_{i}^y, m_{i} = {\rm SAMM}_i (f_i \oplus g_{i, (j-1)})$
            \STATE $\Delta_{i}^x \leftarrow \Delta_{i}^x + \delta_{i}^x$,~ $\Delta_{i}^y \leftarrow \Delta_{i}^y + \delta_{i}^y$   
            \STATE \textbf{if} $j = 1$ \textbf{then} $M_{i} \leftarrow m_{i}$
            \STATE \textbf{else} $M_{i} \leftarrow m_{i} \cdot M_{i} + M_{i} \cdot (1 - M_{i})$
            \STATE \textbf{endif}
            \STATE $g_{i,j} = \mathcal{GS} (g_i, \Delta_{i}^x, \Delta_{i}^y) \cdot M_{i} + g_i \cdot (1 - M_{i})$
        \ENDFOR
    \end{algorithmic}
\end{algorithm}
\vspace{-1em}

\subsection{Blending}
\label{sec:blending}

\noindent\textbf{Generate $\widehat{x_{in}}$.} Because $g_{i, N}$ is resampled from $g_i$, it maintains the local semantic attributes of $g_i$ and is spatially aligned to $f_i$. The aligned feature $g_{i, N}$ is then fed into the next styled-convolution layer $G_i$ and $g_{i+1} = G_i(g_{i, N}, w)$. And finally, $\widehat{x_{in}} = {\rm toRGB}(g_L)$, where $g_L$ is the last layer's feature map and ${\rm toRGB}$ is the output convolution in $G$. We visualize the generation process of $\widehat{x_{in}}$ in Fig.~\ref{fig:overview}.
In this paper, we align $g_i$ and $f_i$ in multiple resolutions of $32^2, 64^2, 128^2$, and $256^2$. Also, we set $N=2$ in most of our experiments.
For more details on our architecture, please refer to our supplement. 

\begin{figure}[t]
    \centering
    \begin{tabular}{@{}c@{\hspace{2mm}}c@{\hspace{2mm}}c@{}}
         \scriptsize{Input} & \scriptsize{Inversion} & \scriptsize{Narrow eyes}
        \\
         \includegraphics[width=0.3\linewidth]{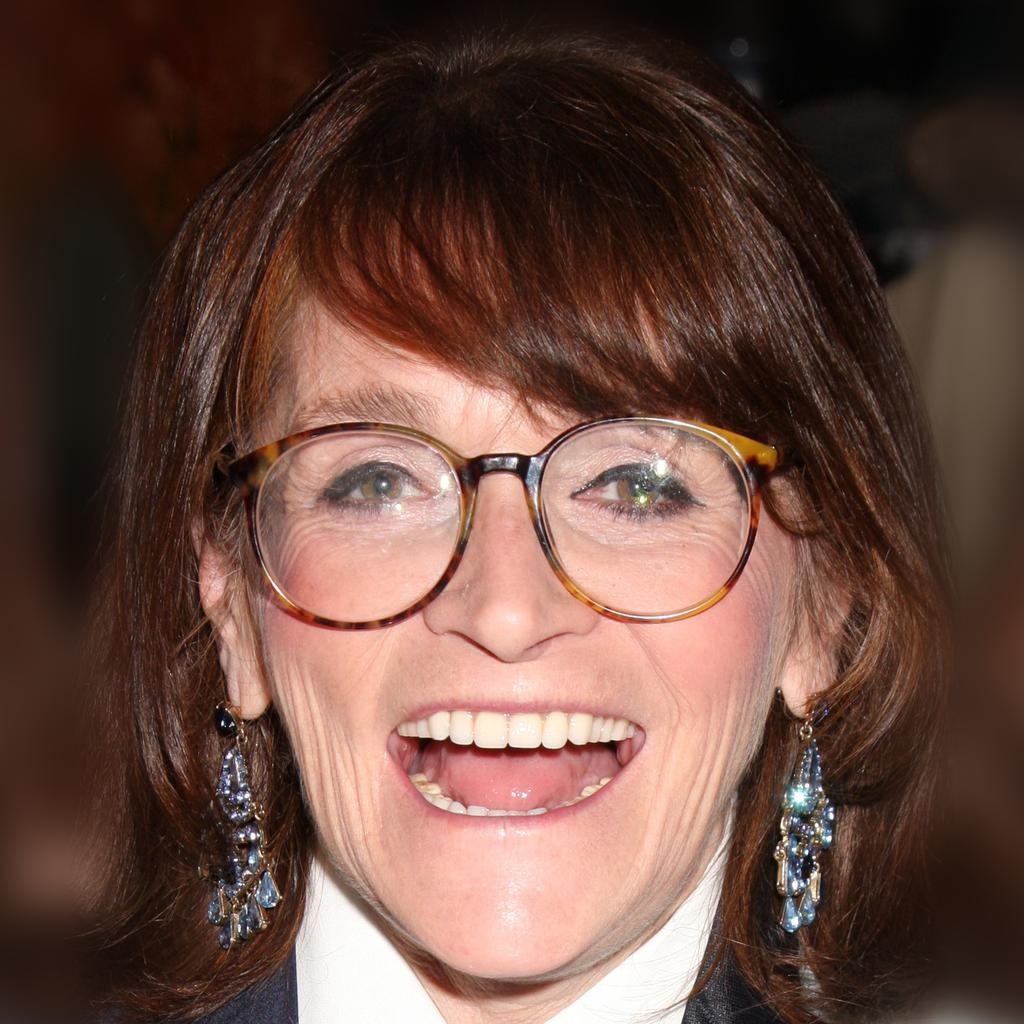} &
         \includegraphics[width=0.3\linewidth]{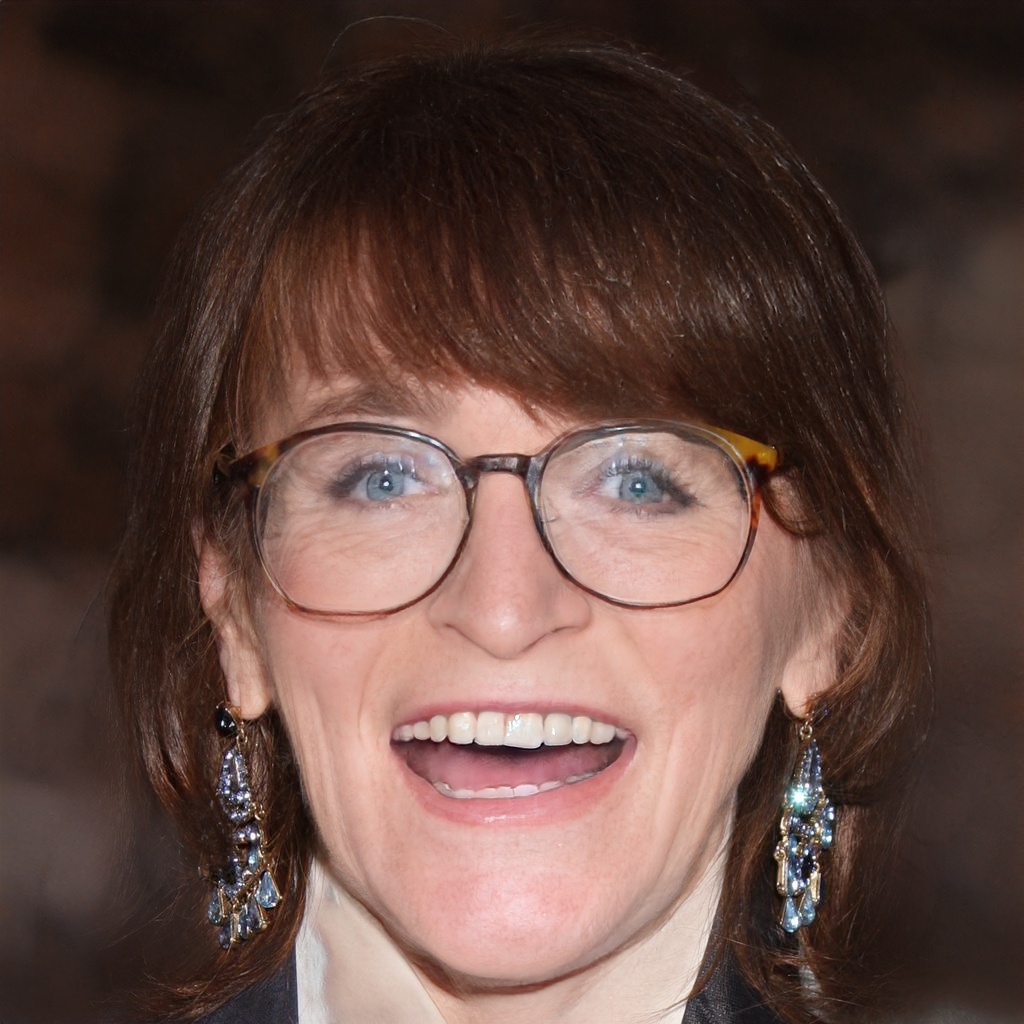} &
         \includegraphics[width=0.3\linewidth]{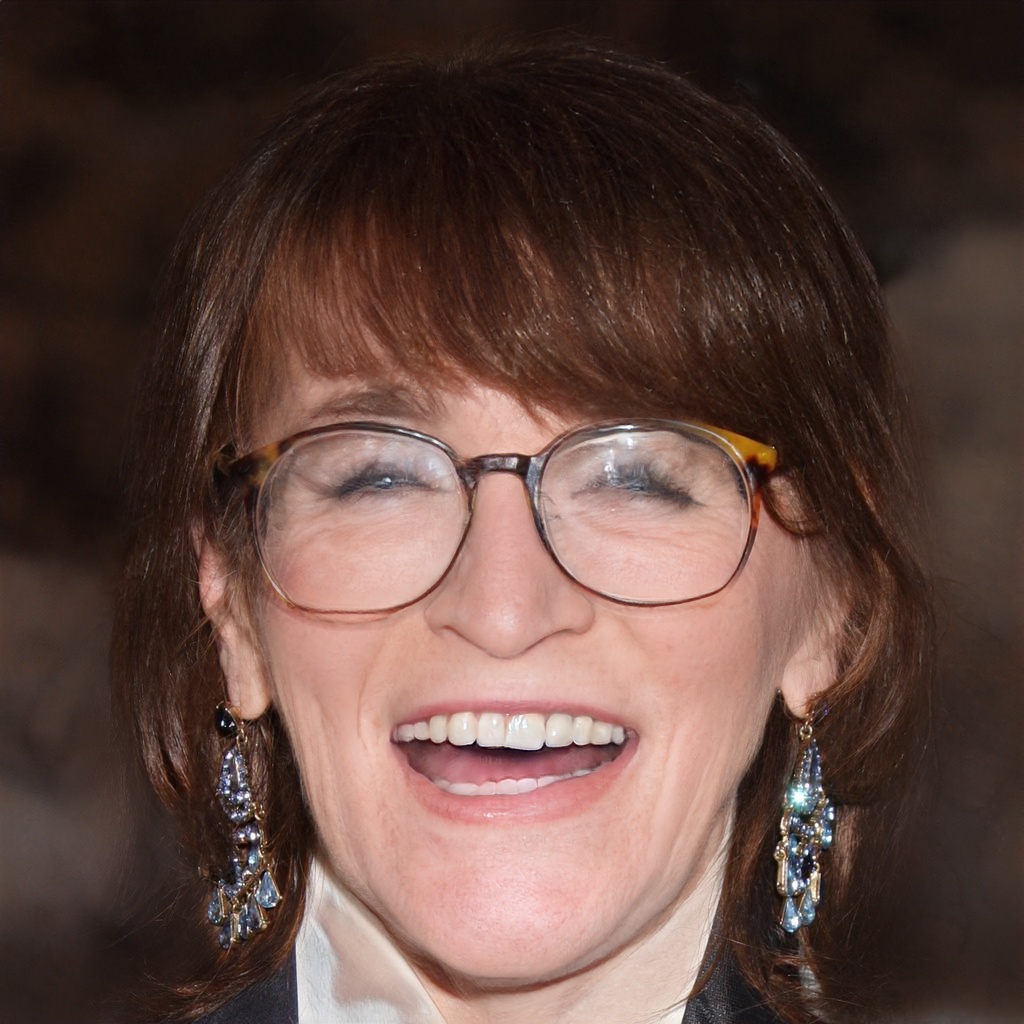}
        \\
         \multicolumn{3}{@{}c@{}}{\scriptsize{Predicted masks}}
         \\
         \multicolumn{3}{@{}c@{}}{\includegraphics[width=0.93\linewidth]{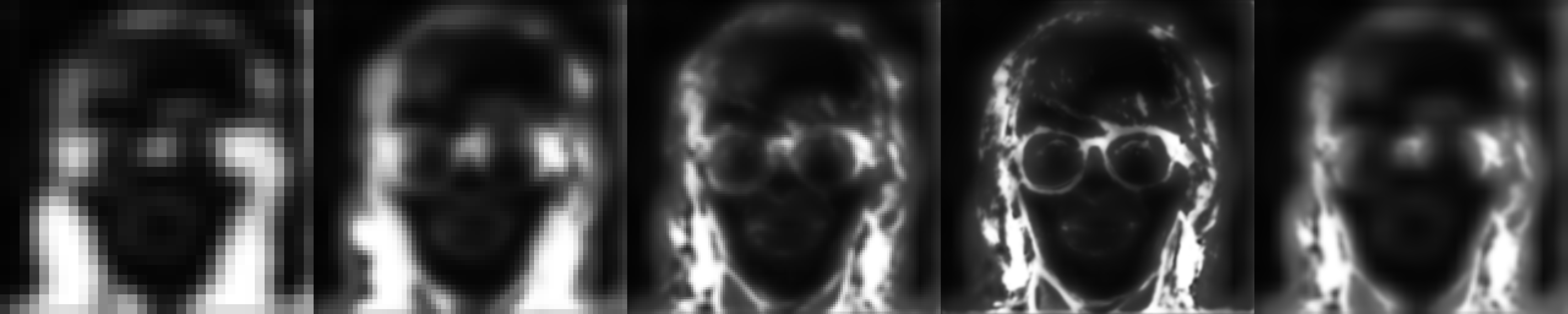}}  
    \end{tabular}
    \vspace{-1em}
    \caption{The first row shows our GAN inversion and editing result with our method. The second row shows the predicted masks $M_{i,N}$ in the resolution of $32^2, 64^2, 128^2, 256^2$ and $m$ in the resolution of $1024^2$, respectively. }
    \label{fig:mask}
    \vspace{-1.5em}
\end{figure}

\noindent\textbf{Generate $m$.} In order to blend the inversed result $\widehat{x_{in}}$ and $x_{out}$ in the RGB domain (Eqn.~\eqref{eqn:blending}) for a photo-realistic result, it is crucial to find a blending mask $m$ which precisely distinguish the OOD contents only.
As is shown in Fig.~\ref{fig:mask}, we observe that when we set $\phi_{area}$ to a small value in the training stage, the high-intensity values in $M_{i,N}$ primarily gather around the OOD area (e.g., earrings, glasses). The mask of different resolutions focuses on slightly different areas because the target texture of each layer in the source generator is different. Usually, the lower layers of the source generator focus on larger structures, such as the rough shape of the face, and the higher layers focus on detailed structures, such as the skin texture and accessories, which is also observed in \cite{richardson2021encoding, tov2021designing, alaluf2021HyperStyle}. 

Therefore, we could gather $M_{i,N}$ in each layer of generation for a final blending mask $m$ during inference. Instead of directly upscale and merging all $M_{i,N}$ to $m$, we consider that consistent high-intensity area as the OOD area. We adopt a merging function to sequentially merge $M_{i, N}$, $i\in \{1,...,L \}$, as shown in Algorithm~\ref{alg:mask_gathering}.

\begin{algorithm}
    \caption{Gathering the masks.}
    \label{alg:mask_gathering}
    \begin{algorithmic}[1]
        \STATE Initialization: $M_{1,N} = \Uparrow(M_{1, N})$
        \FOR{$i \in \{2, ..., L\}$}
            \STATE $\widetilde{M_{i, N}} \leftarrow \widetilde{M_{(i-1), N}} \cdot (\Uparrow(M_{i, N}) - \widetilde{M_{(i-1), N}} + 1)$
        \ENDFOR
    \STATE Update: $m \leftarrow \widetilde{M_{L, N}}$
    \end{algorithmic}
\end{algorithm}

Where $\Uparrow$ means up-sampling to the output resolution of the generator, i.e., $1024^2$ for the pre-trained StyleGAN2 generator on face images. 

\noindent\textbf{Generate $\hat{x}$.} Our final output of GAN inversion result $\hat{x}$ is produced following Eqn.~\eqref{eqn:decomp_gan_inversion}. The blending process is visualized in Fig.~\ref{fig:blending}, where we blend $\widehat{x_{in}}$ and $x_{out}$ with $m$.

\subsection{Training Objectives}
\label{sec:training}
In Sec.~\ref{sec:SAMM}, we propose the SAMM to align the generated feature $g$ to $f$, where the invertibility is defined by $m$.
In this section, we demonstrate the loss functions to train SAMM.
First, we assume that
the subtle misalignment in $\widehat{x_{in}}$ can be fixed with simple spatial operations in Algorithm.~\ref{alg:samm}. 
Hence, we train our framework by minimizing the reconstruction loss $L_{rec}$ on $\hat{x}$. Here we adopt the VGG perceptual loss~\cite{johnson2016perceptual} $L_{per}$, the MSE loss and the ArcFace~\cite{deng2018arcface} identity loss $L_{id}$ as our reconstruction objectives, that
\begin{equation}
    L_{rec} = L_{per}(x, \hat{x}) + {\rm MSE}(x, \hat{x}) + L_{id}(x, \hat{x}),
\end{equation}
please refer to our supplement for detail definition of $L_{per}$ and $L_{id}$. Also, to make $\widehat{x_{in}}$ look realistic, we keep the vanilla adversarial loss $L_{adv}$ for GAN model~\cite{karras2020analyzing} training. Here we skip the definition of $L_{adv}$ for simplicity. Refer to our supplement for more details.

\begin{figure*}[t]
    \centering
    \begin{tabular}{@{}c@{\hspace{1mm}}c@{\hspace{1mm}}c@{\hspace{1mm}}c@{\hspace{1mm}}c@{\hspace{1mm}}c@{\hspace{1mm}}c@{\hspace{1mm}}c@{}}
        \scriptsize{Ground truth} & 
        \scriptsize{e4e~\cite{tov2021designing}}  &
        \scriptsize{HFGI$_{e4e}$~\cite{wang2021high}}  &
        \scriptsize{HyperStyle~\cite{alaluf2021HyperStyle}, iter=5}  &
        \scriptsize{SAM~\cite{parmar2022spatially}, iter=10}  &
        \scriptsize{FeatureStyle~\cite{xuyao2022}} &
        \scriptsize{Ours$_{e4e}$} &
        \scriptsize{Ours$_{ReStyle}$}
        \\
        \includegraphics[width=0.12\linewidth]{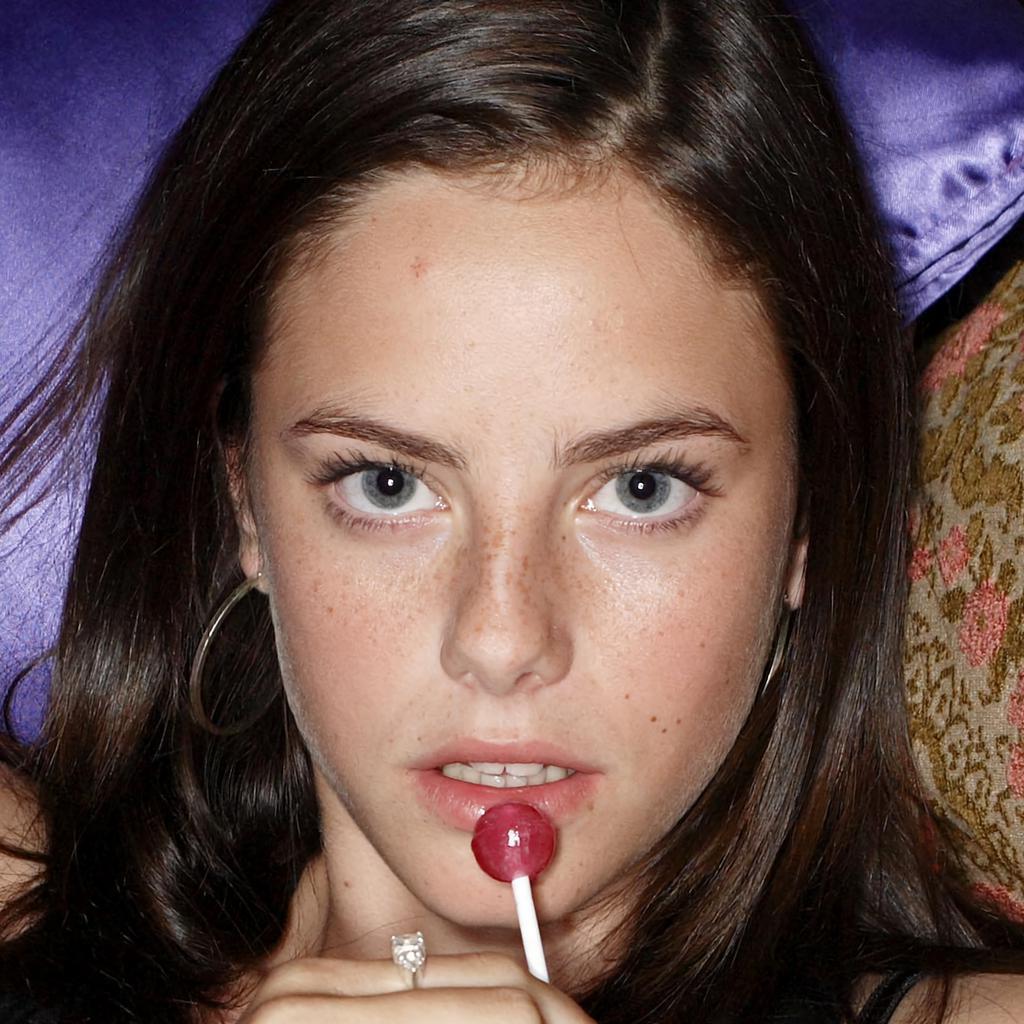} &
        \includegraphics[width=0.12\linewidth]{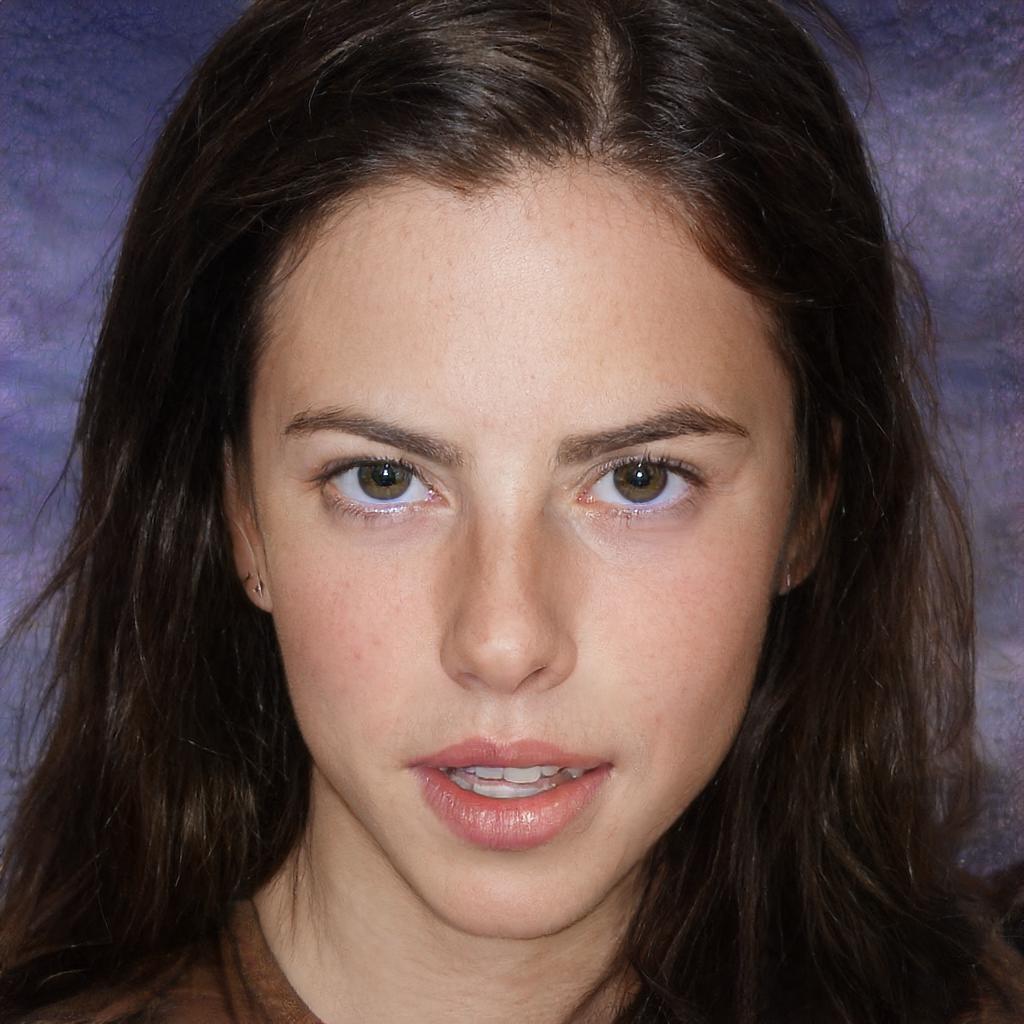} &
        \includegraphics[width=0.12\linewidth]{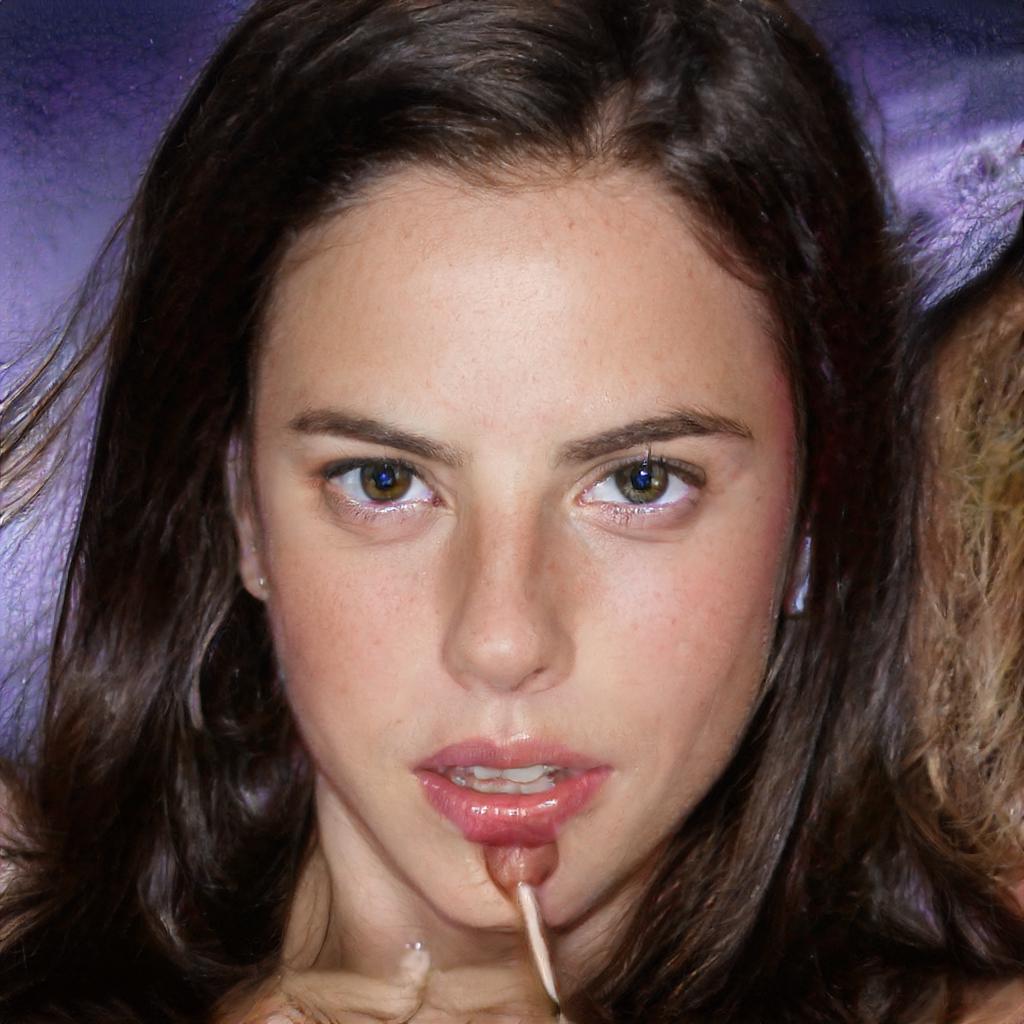} &
        \includegraphics[width=0.12\linewidth]{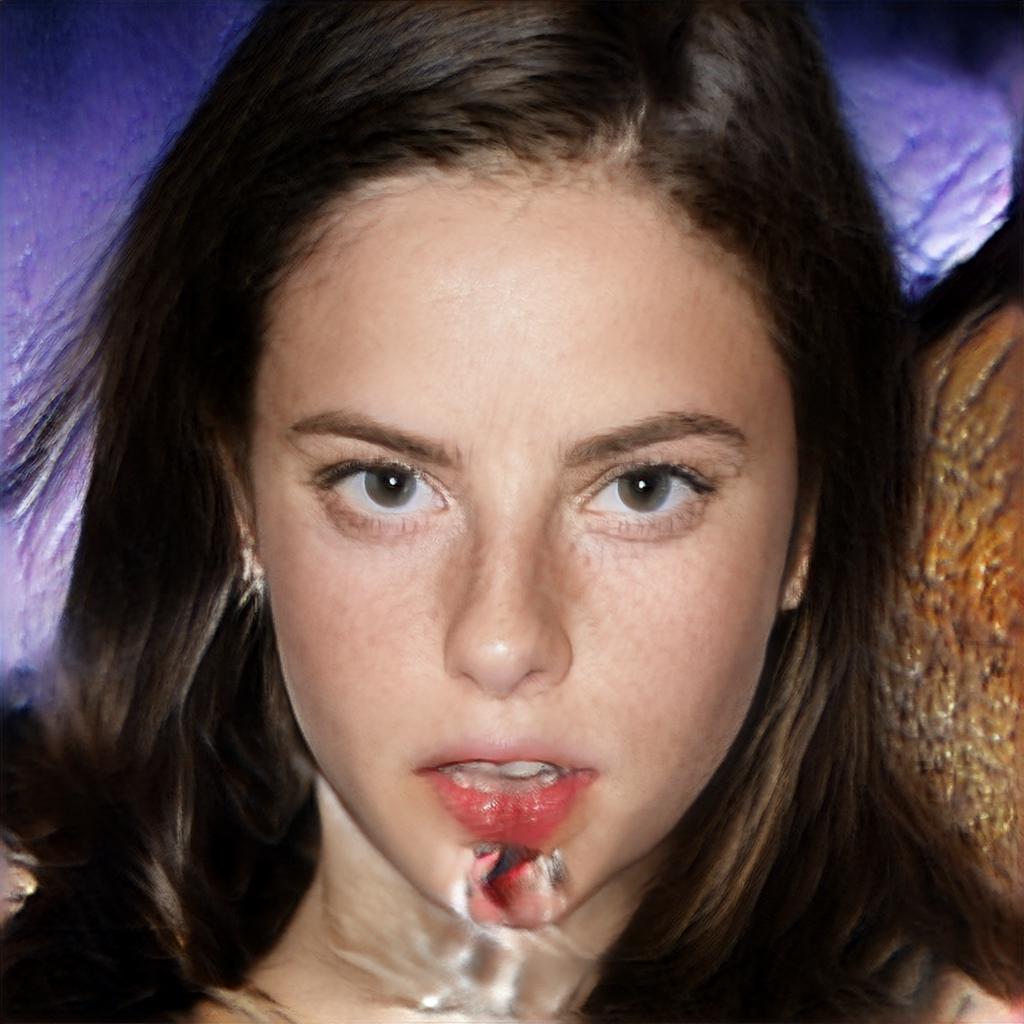} &
        \includegraphics[width=0.12\linewidth]{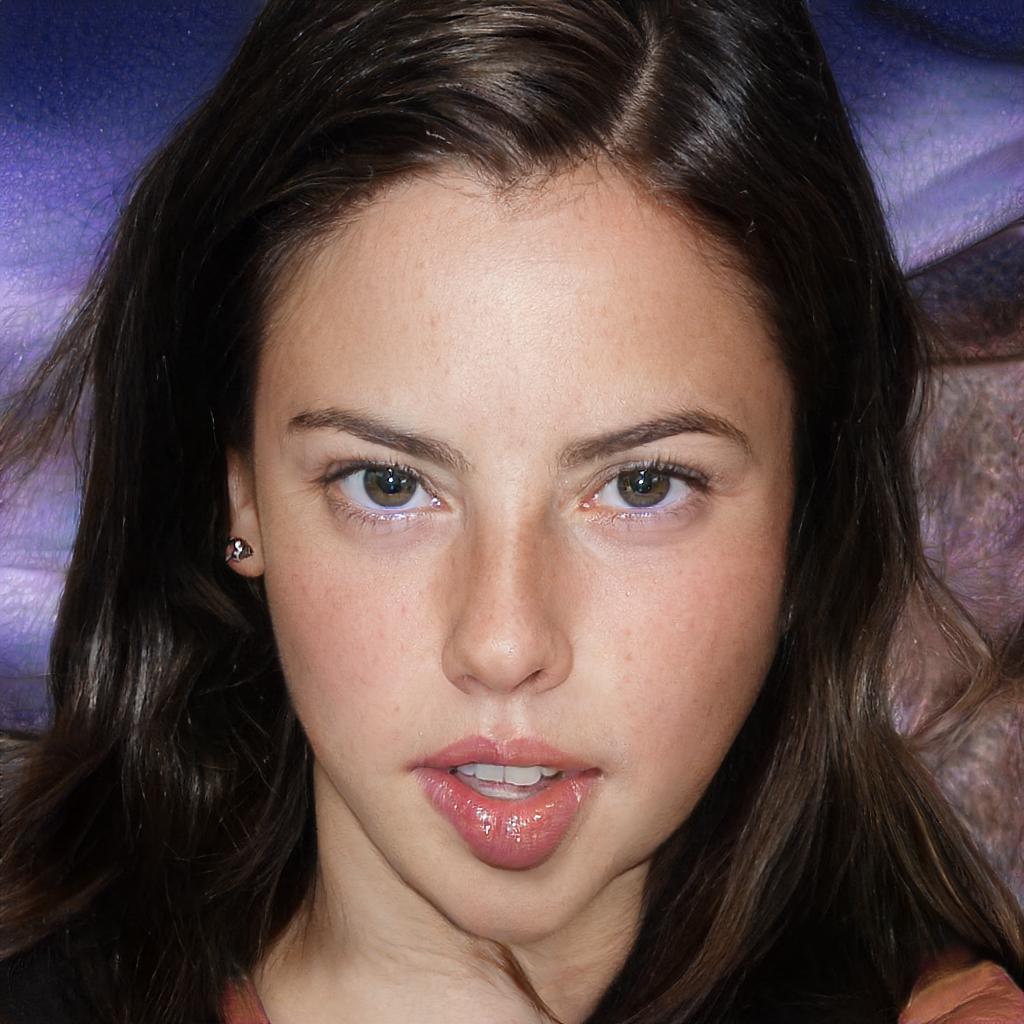} &
        \includegraphics[width=0.12\linewidth]{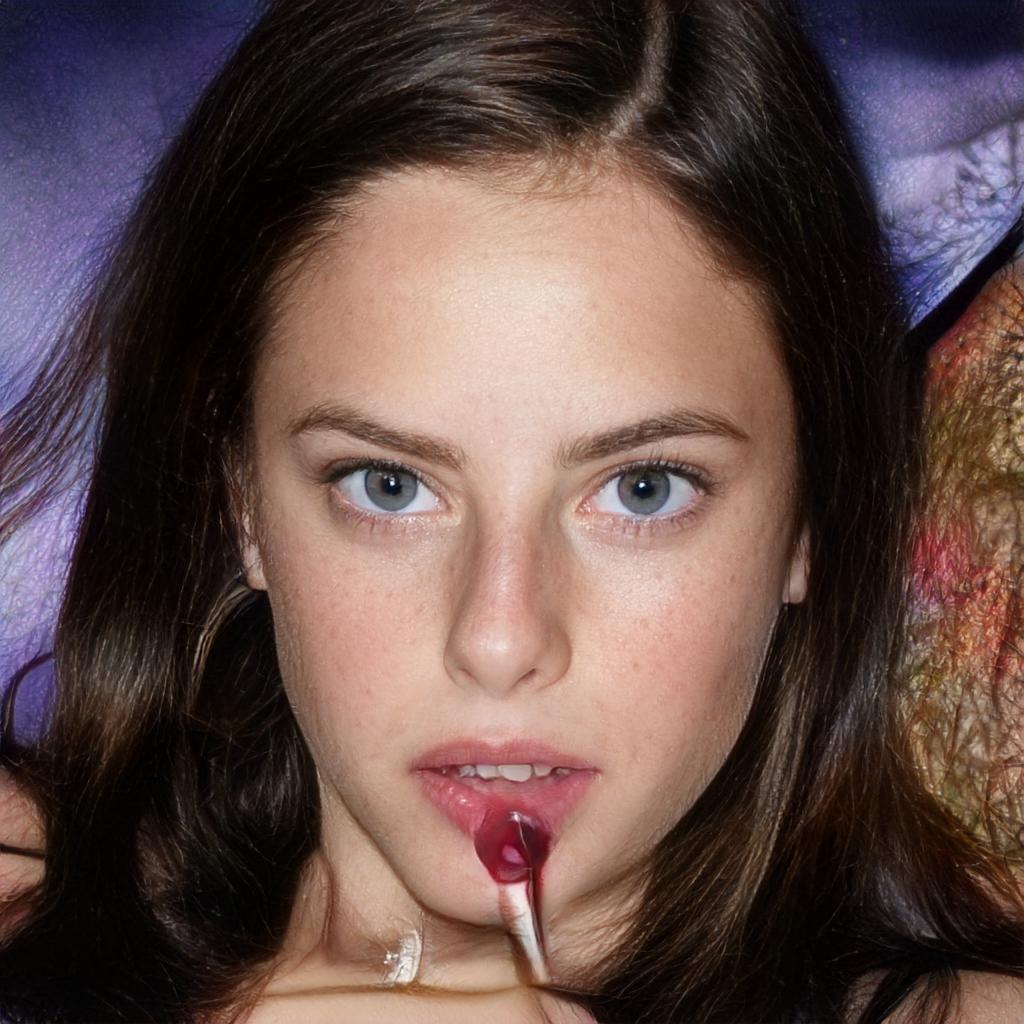} &
        \includegraphics[width=0.12\linewidth]{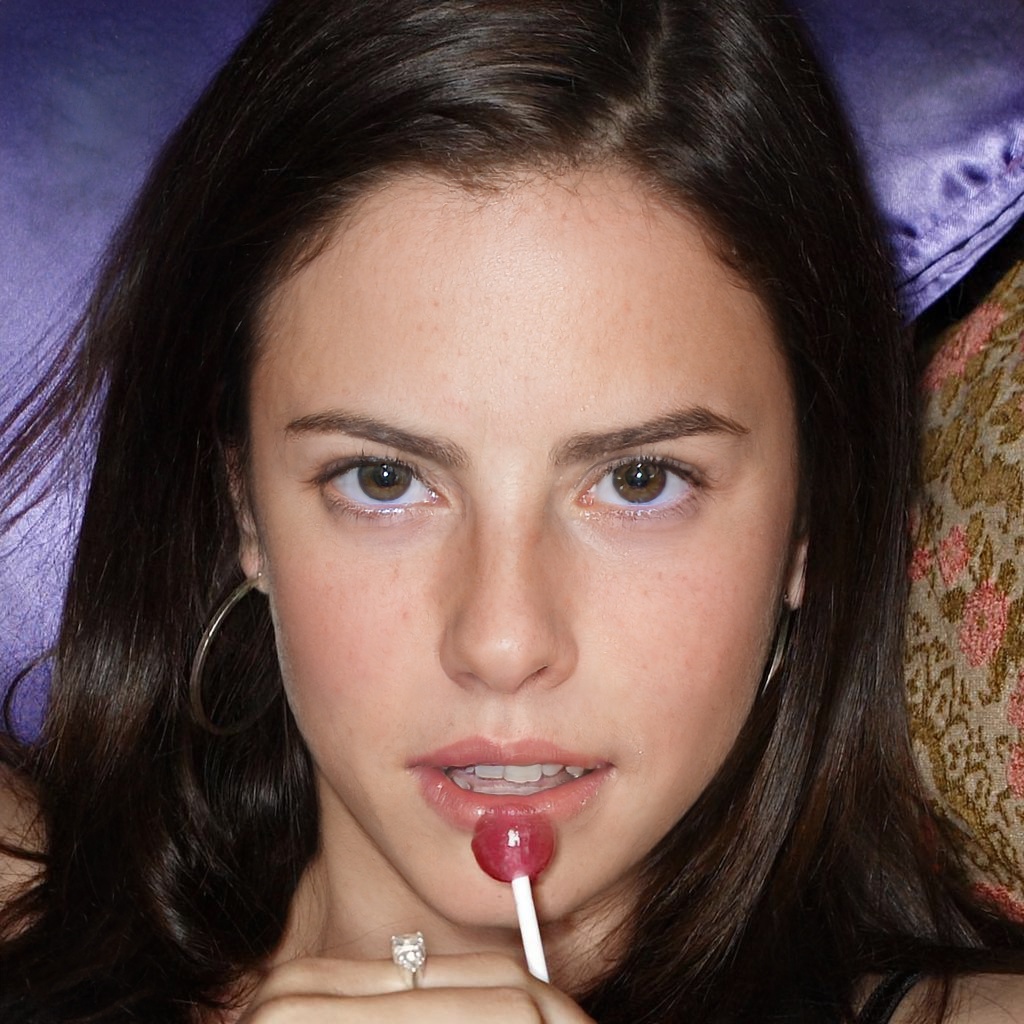} &
        \includegraphics[width=0.12\linewidth]{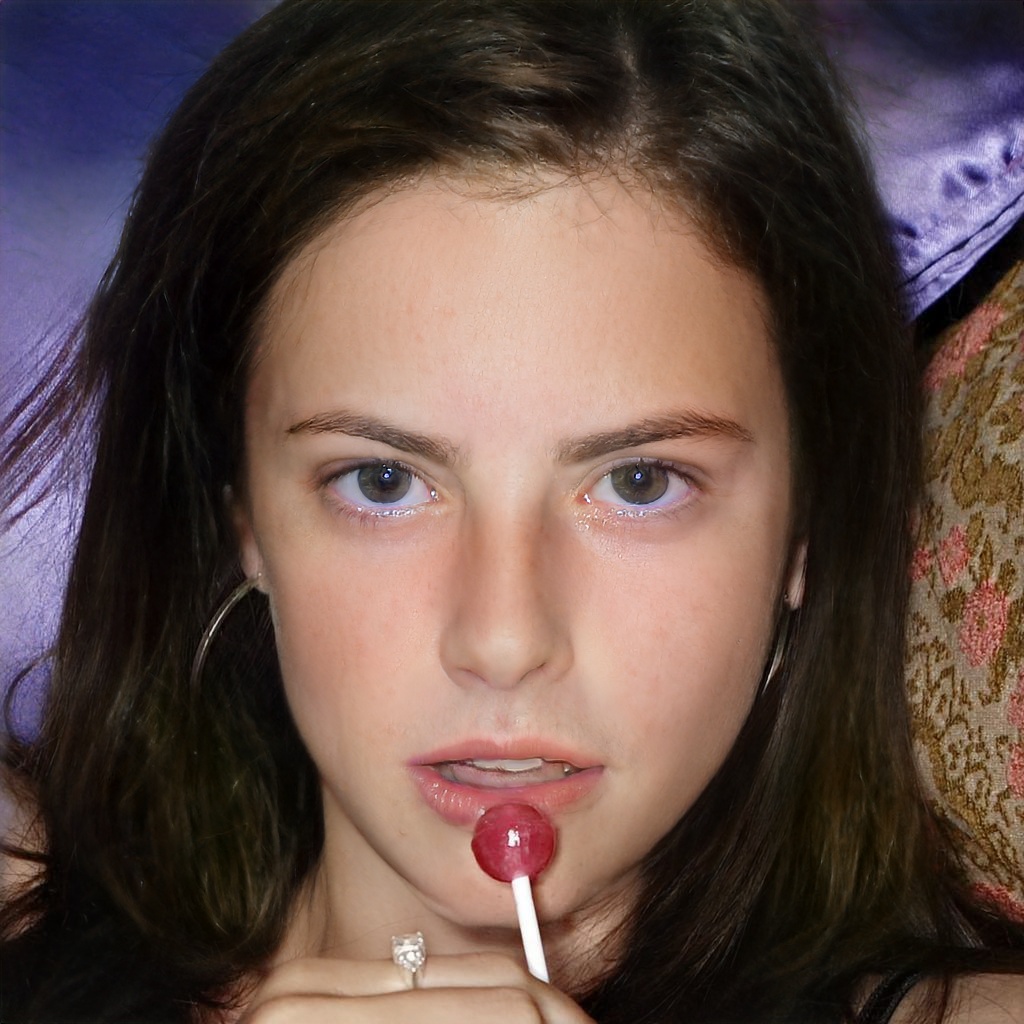}
        \\
        \includegraphics[width=0.12\linewidth]{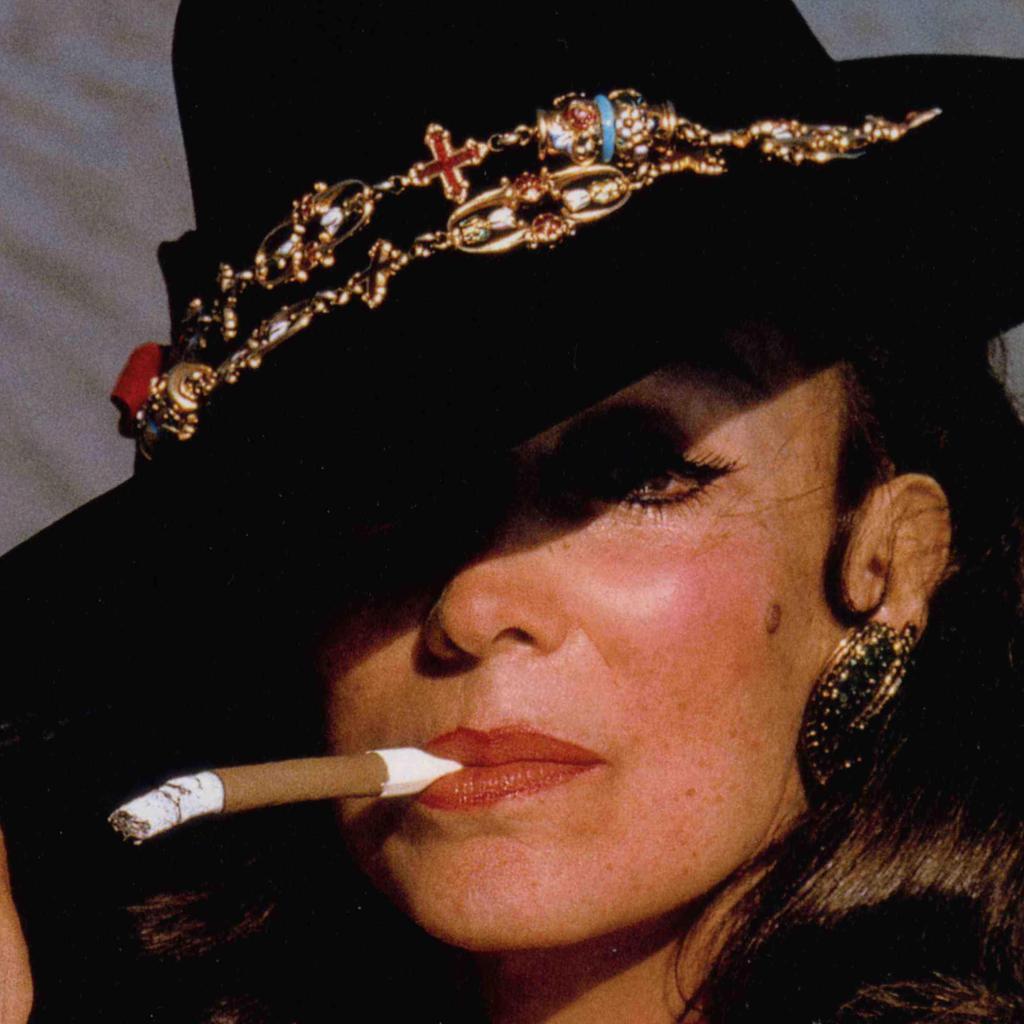} &
        \includegraphics[width=0.12\linewidth]{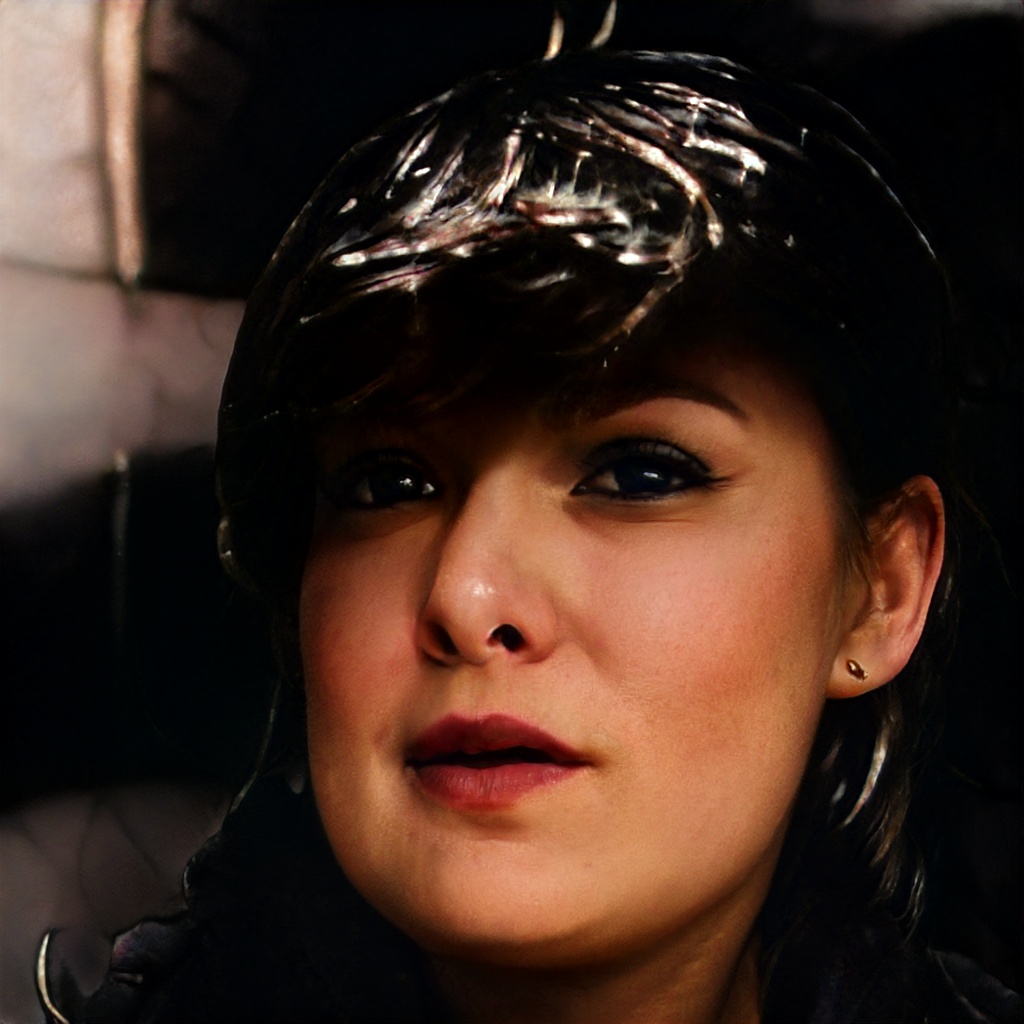} &
        \includegraphics[width=0.12\linewidth]{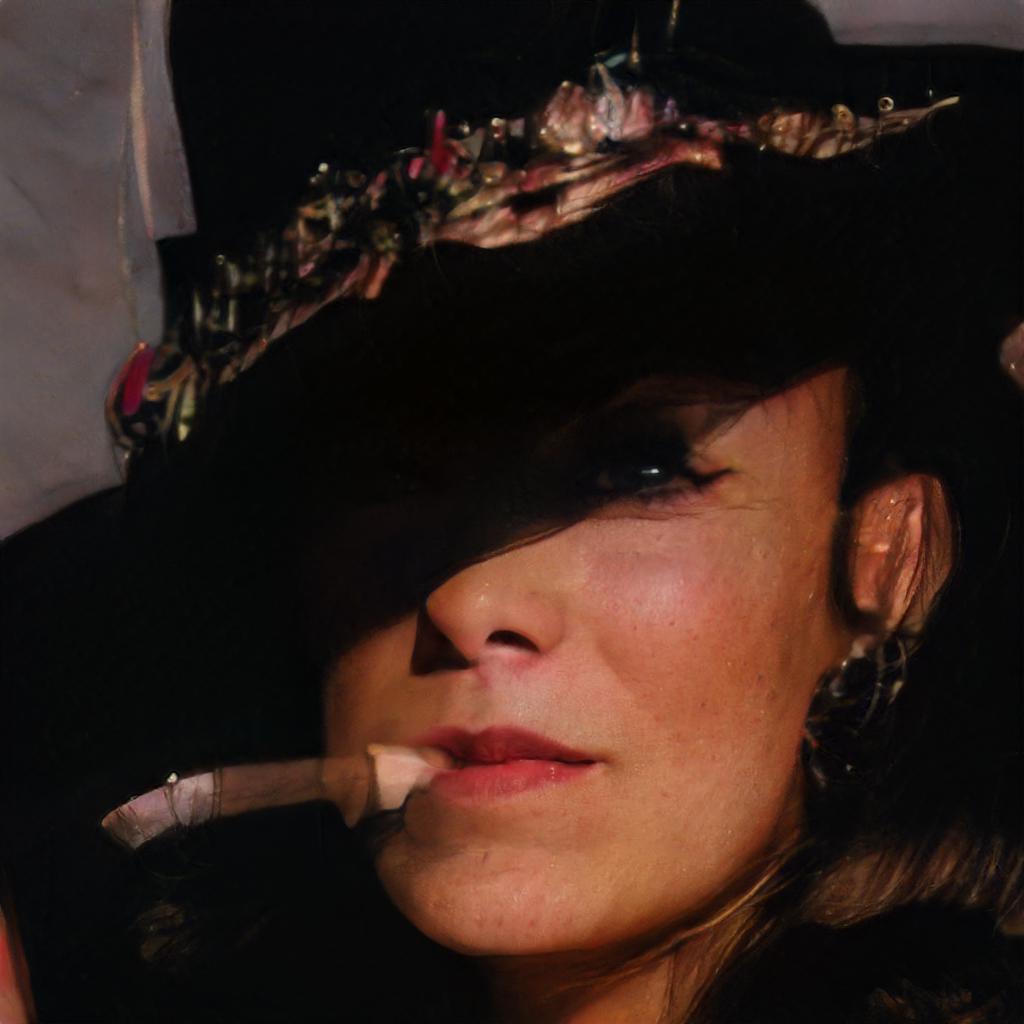} &
        \includegraphics[width=0.12\linewidth]{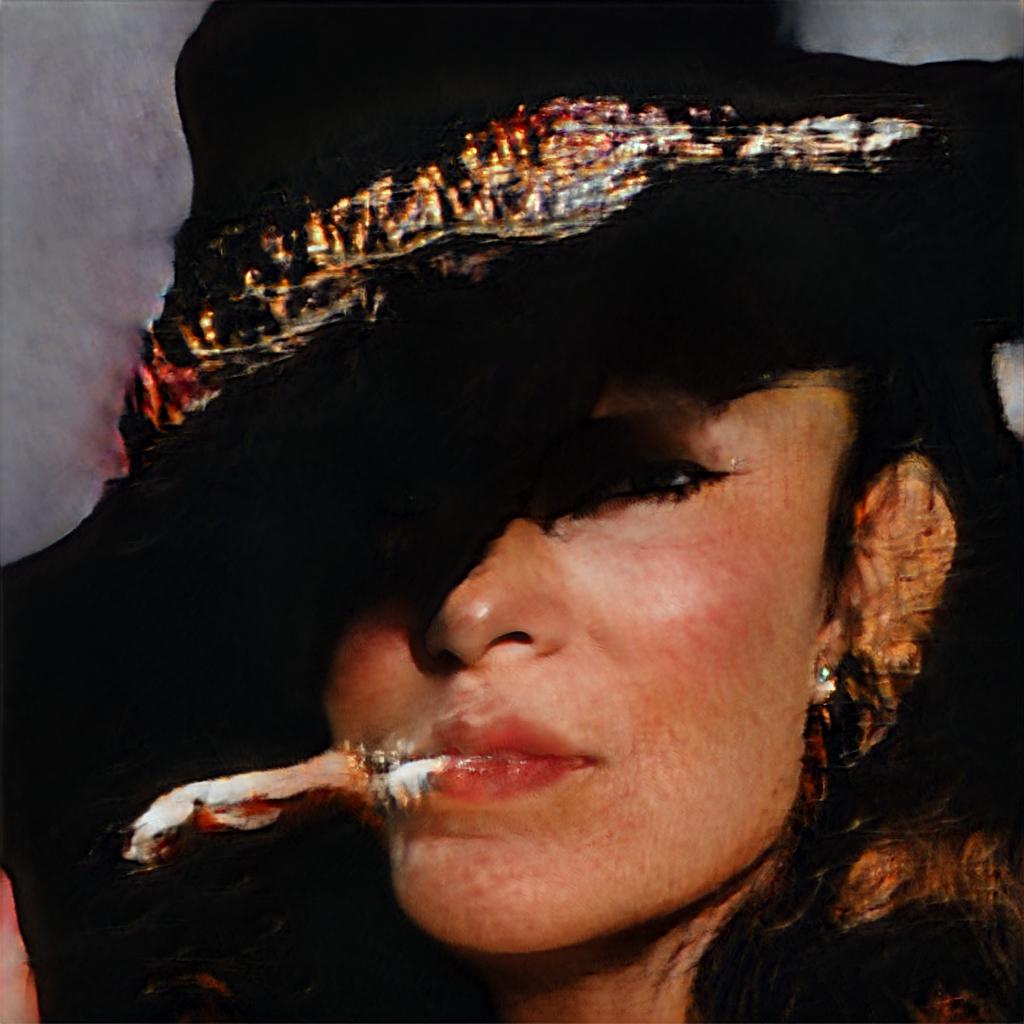} &
        \includegraphics[width=0.12\linewidth]{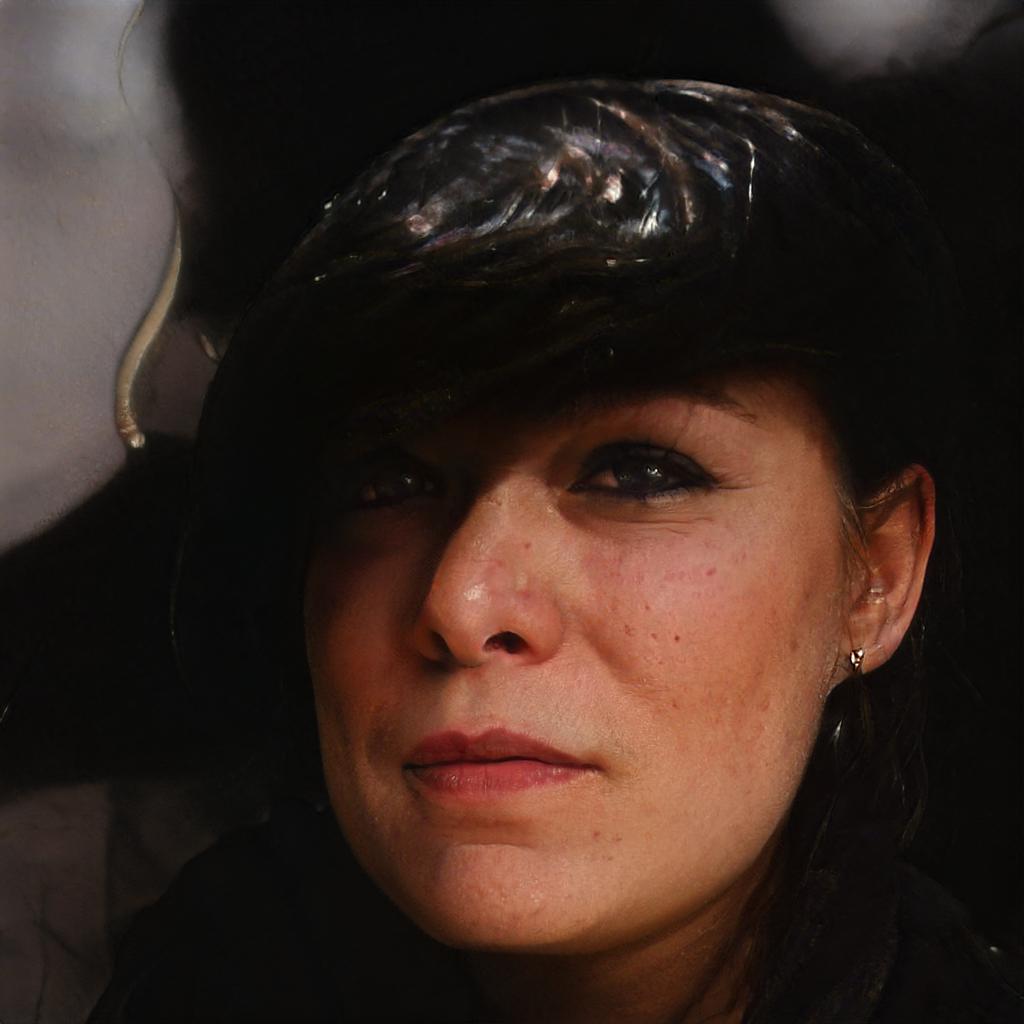} &
        \includegraphics[width=0.12\linewidth]{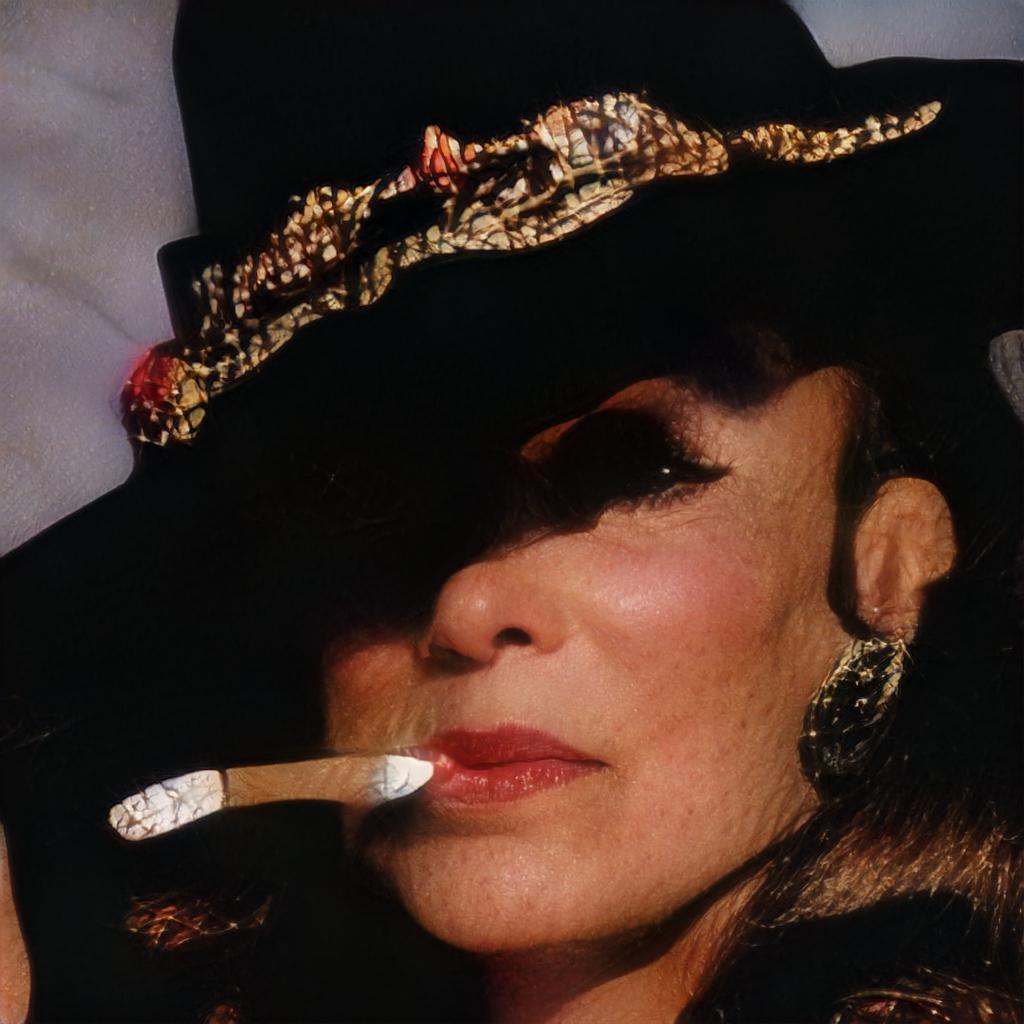} &
        \includegraphics[width=0.12\linewidth]{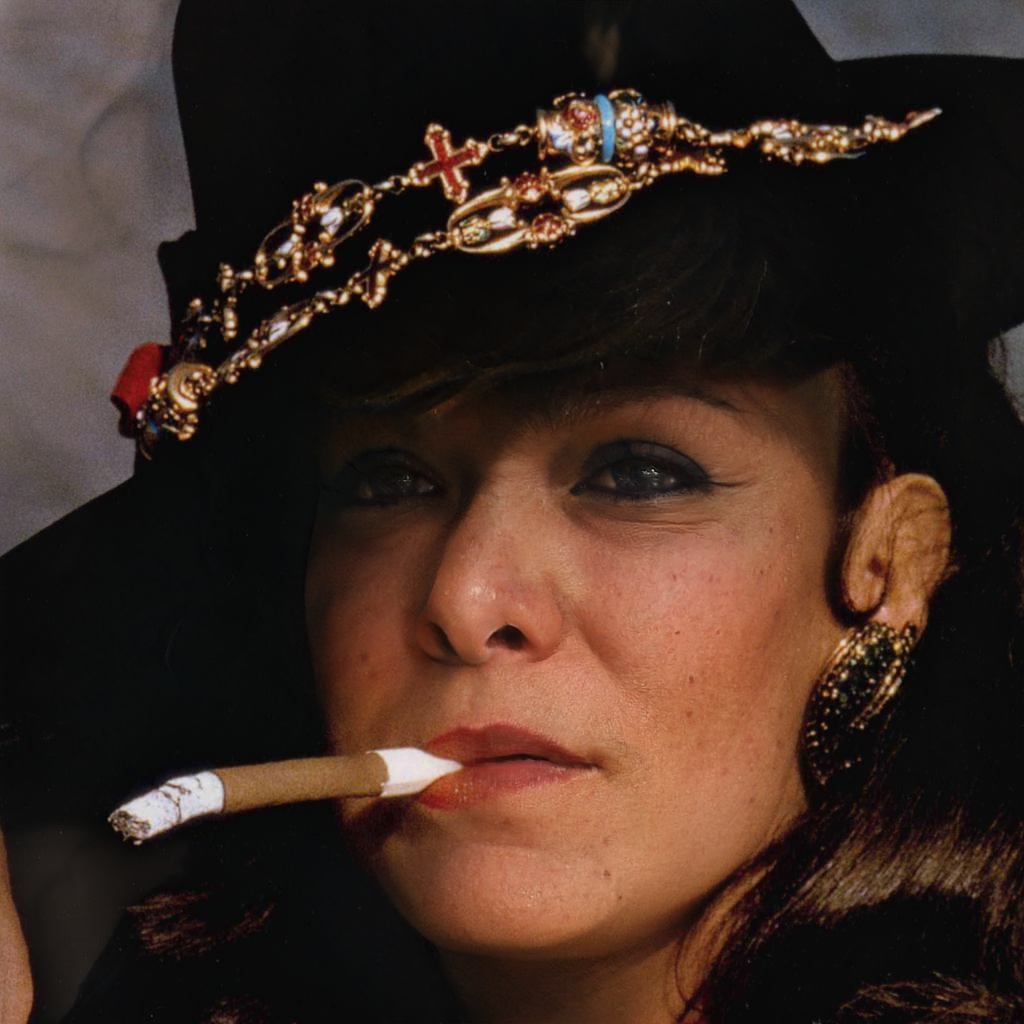} &
        \includegraphics[width=0.12\linewidth]{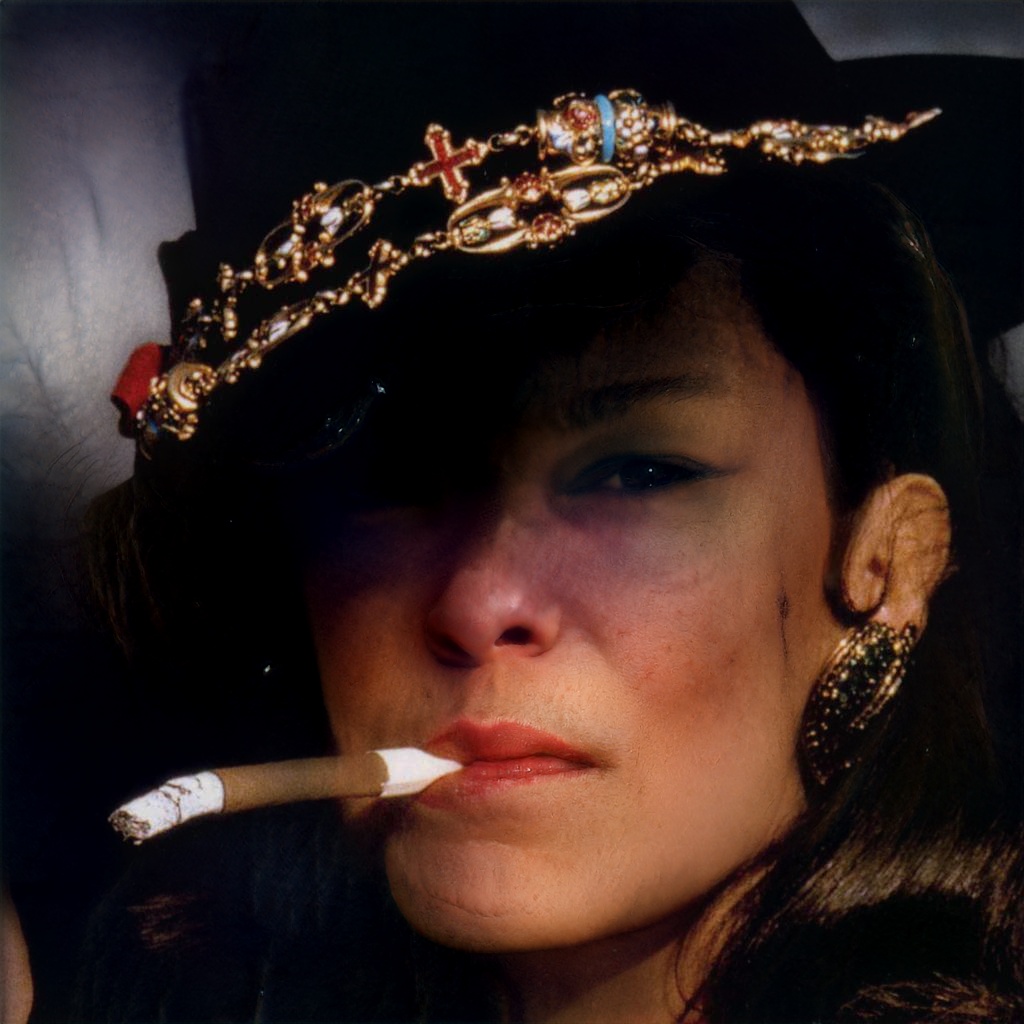}
        \\
        \includegraphics[width=0.12\linewidth]{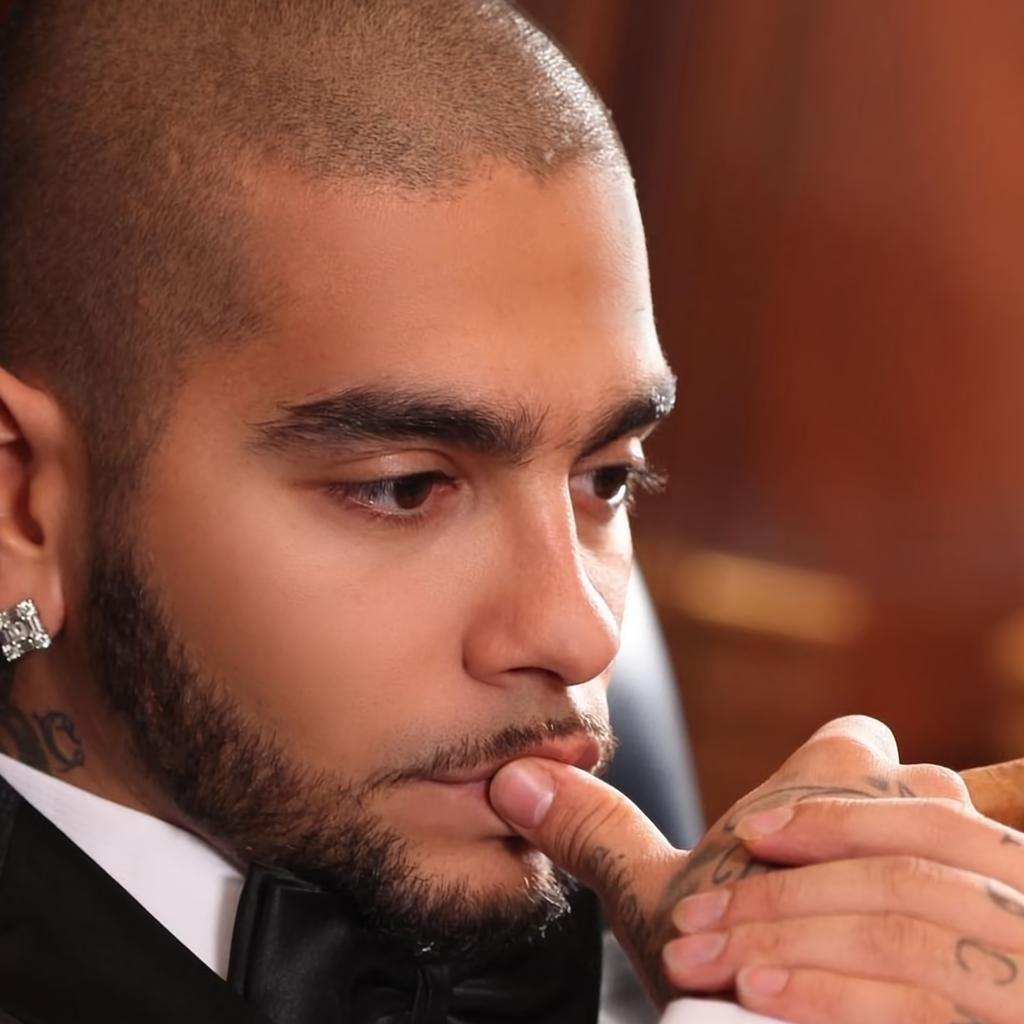} &
        \includegraphics[width=0.12\linewidth]{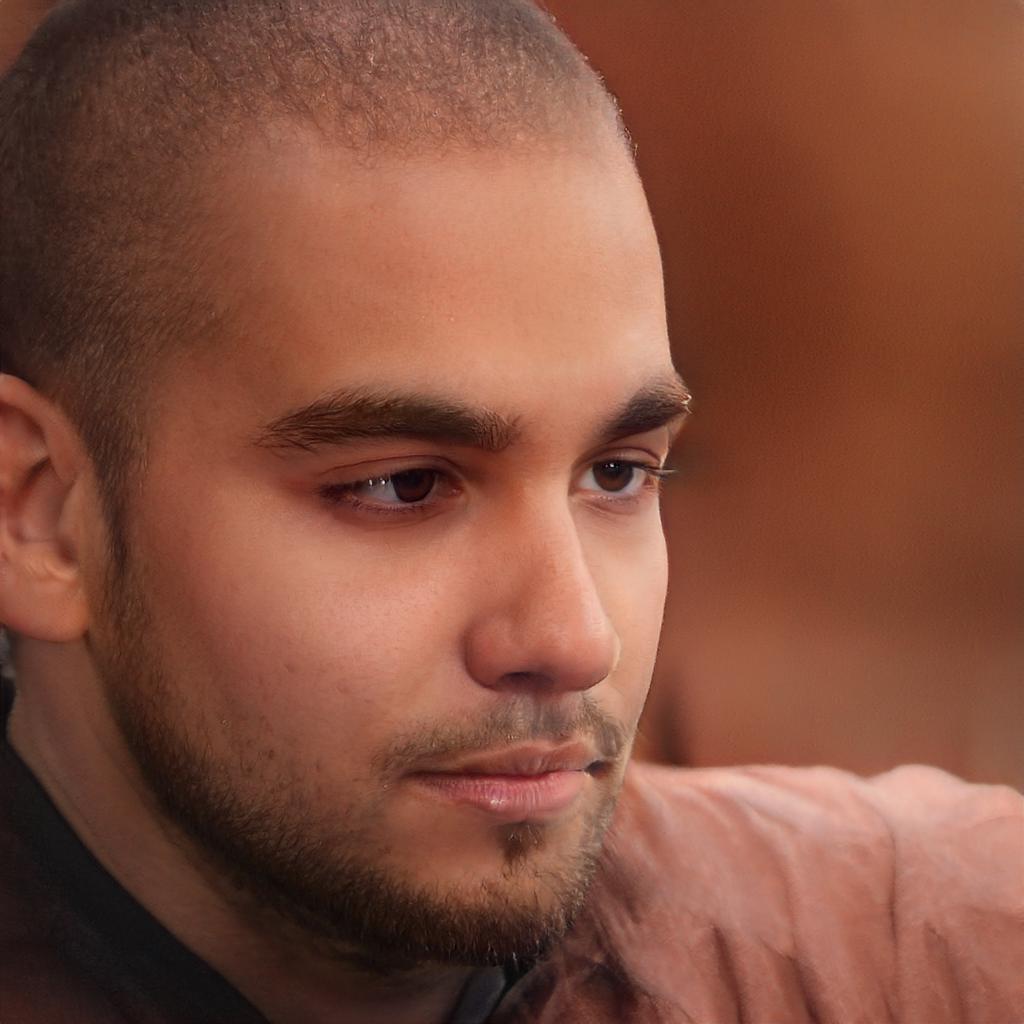} &
        \includegraphics[width=0.12\linewidth]{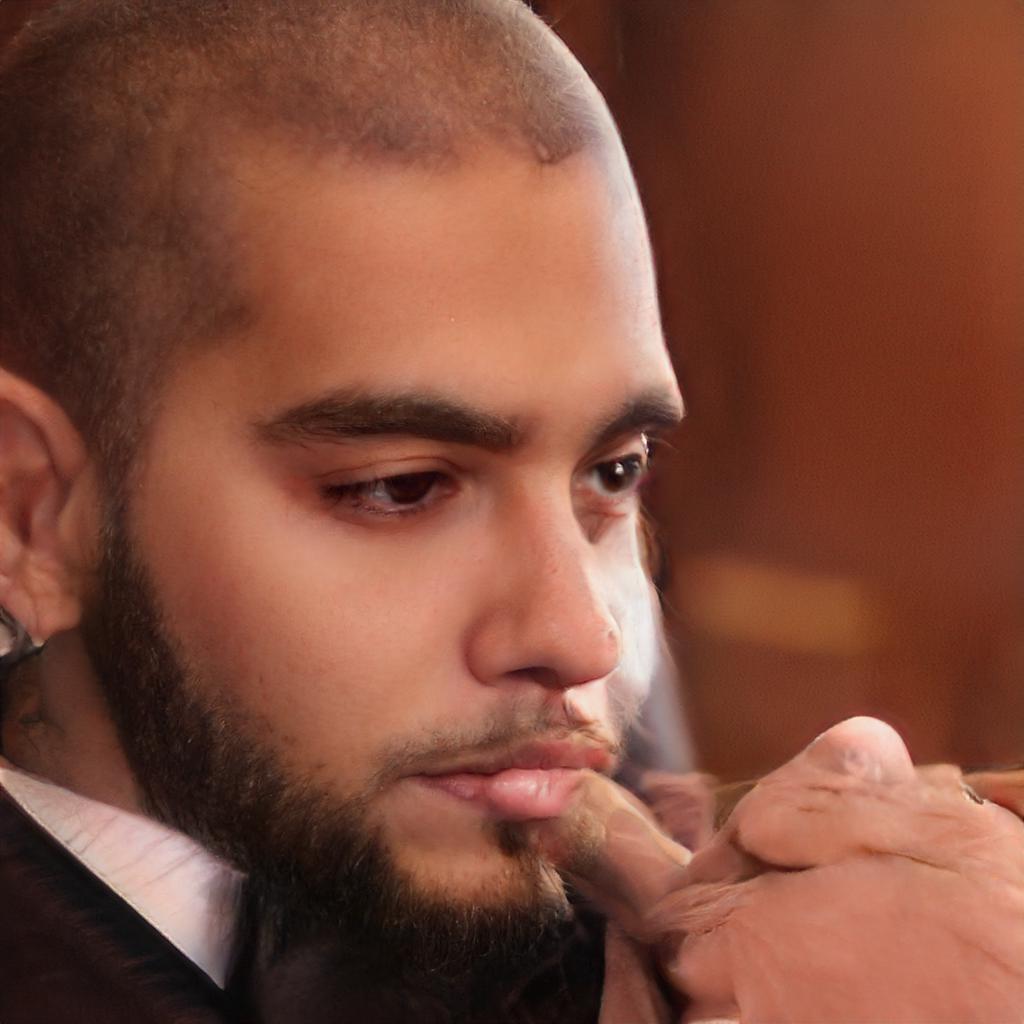} &
        \includegraphics[width=0.12\linewidth]{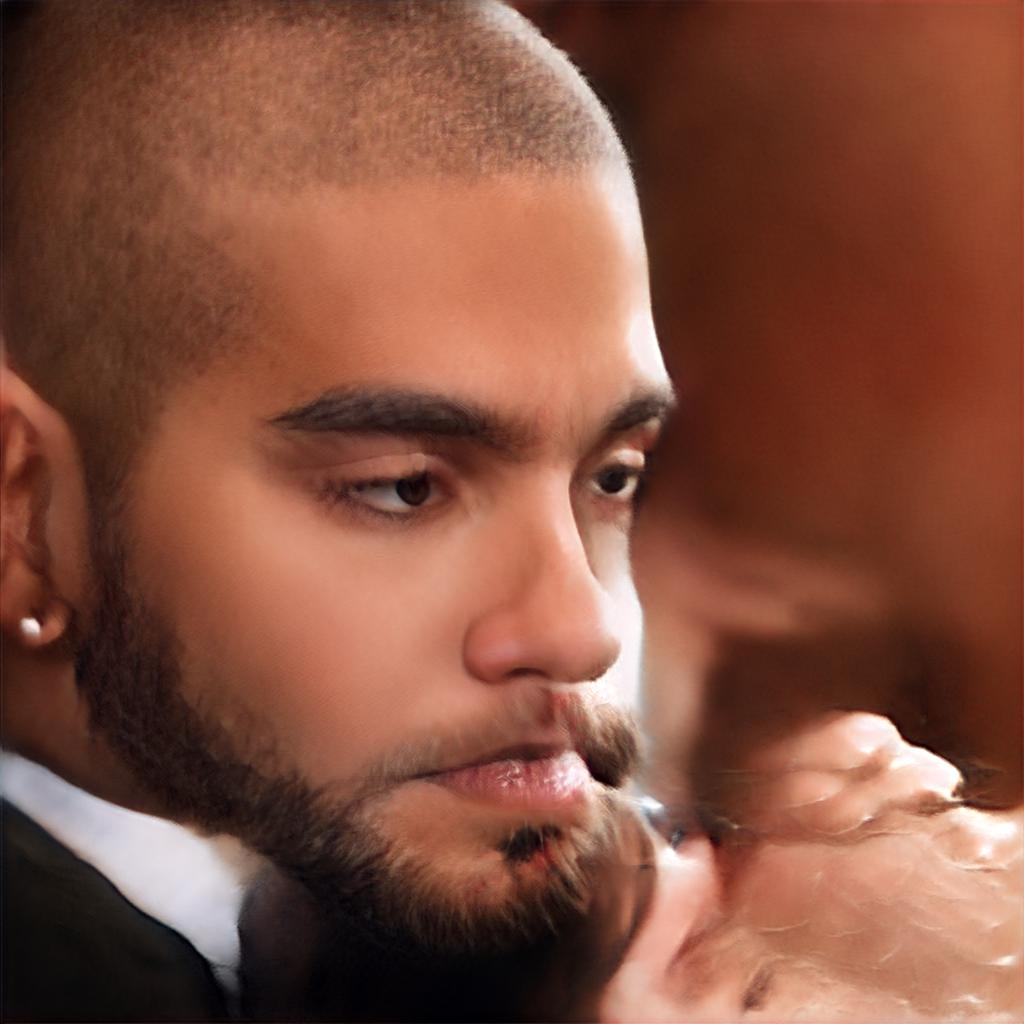} &
        \includegraphics[width=0.12\linewidth]{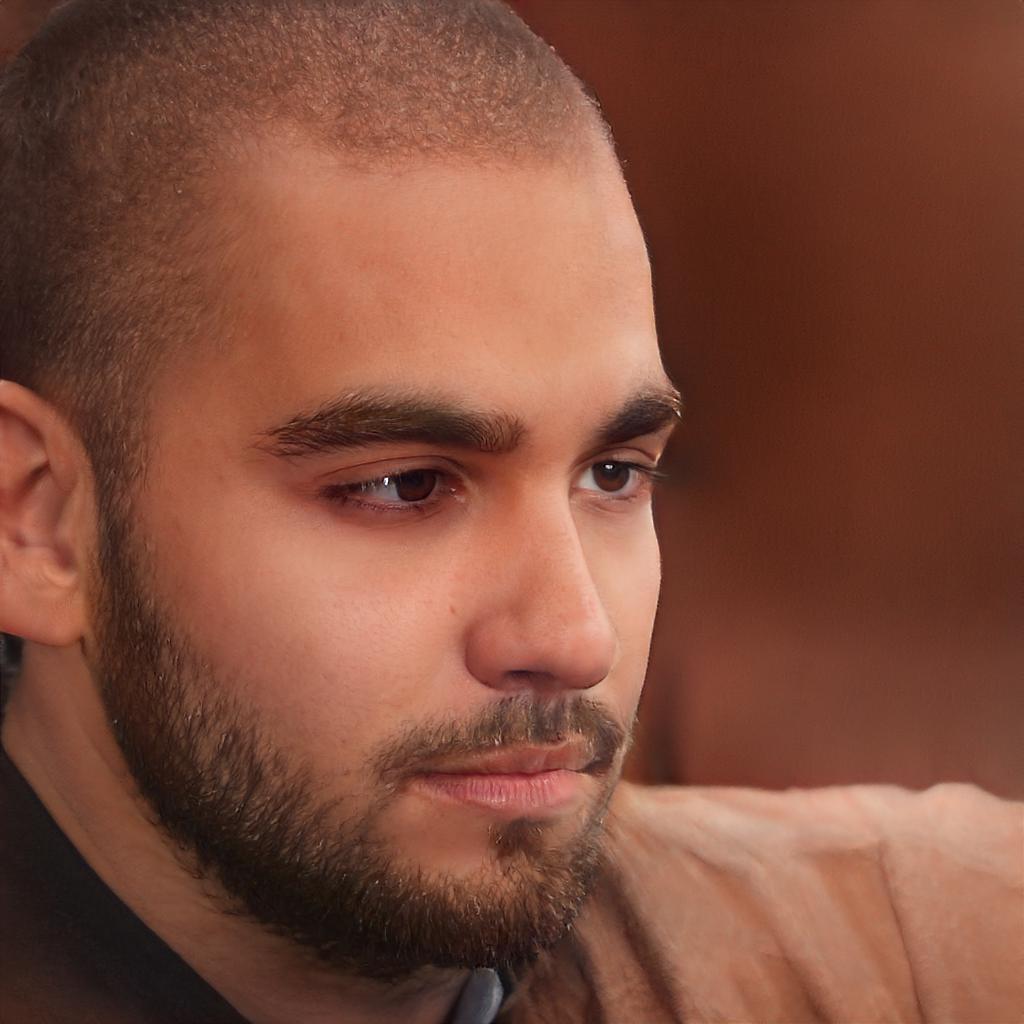} &
        \includegraphics[width=0.12\linewidth]{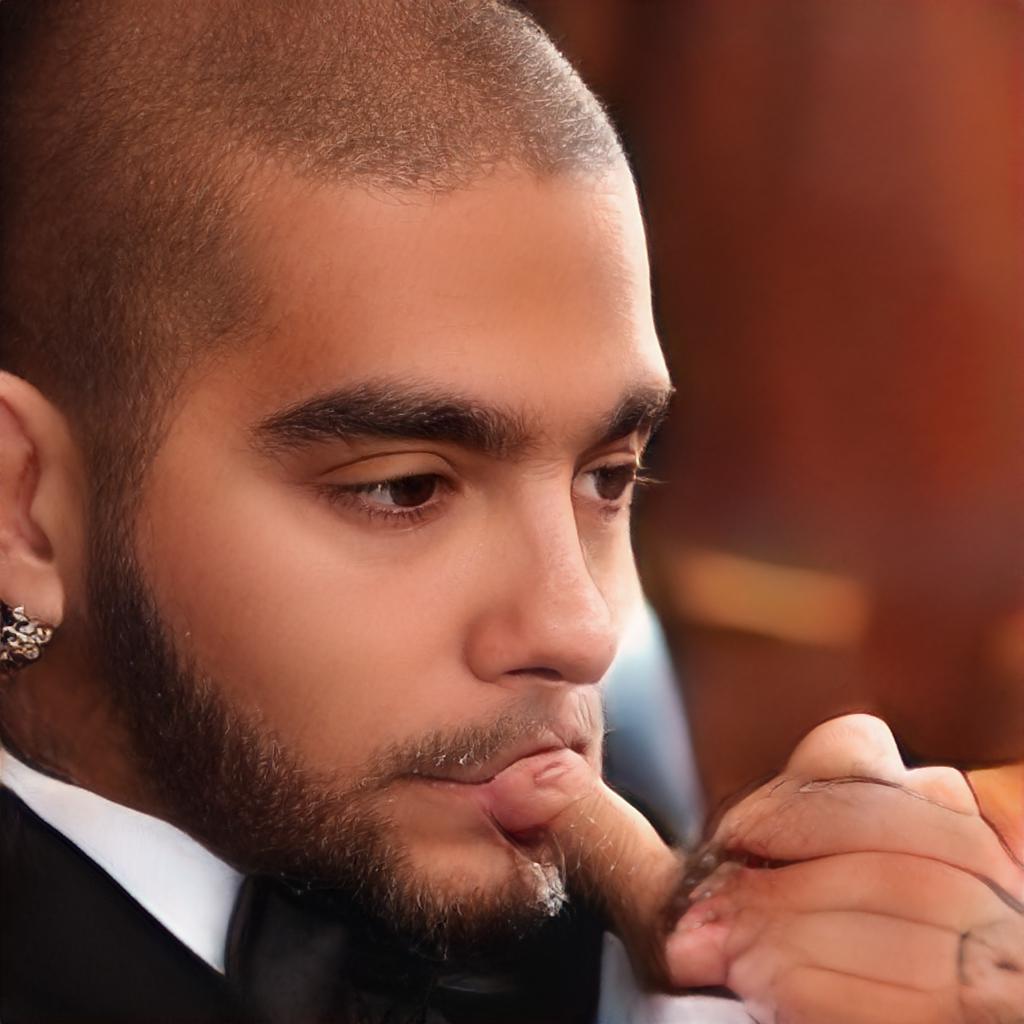} &
        \includegraphics[width=0.12\linewidth]{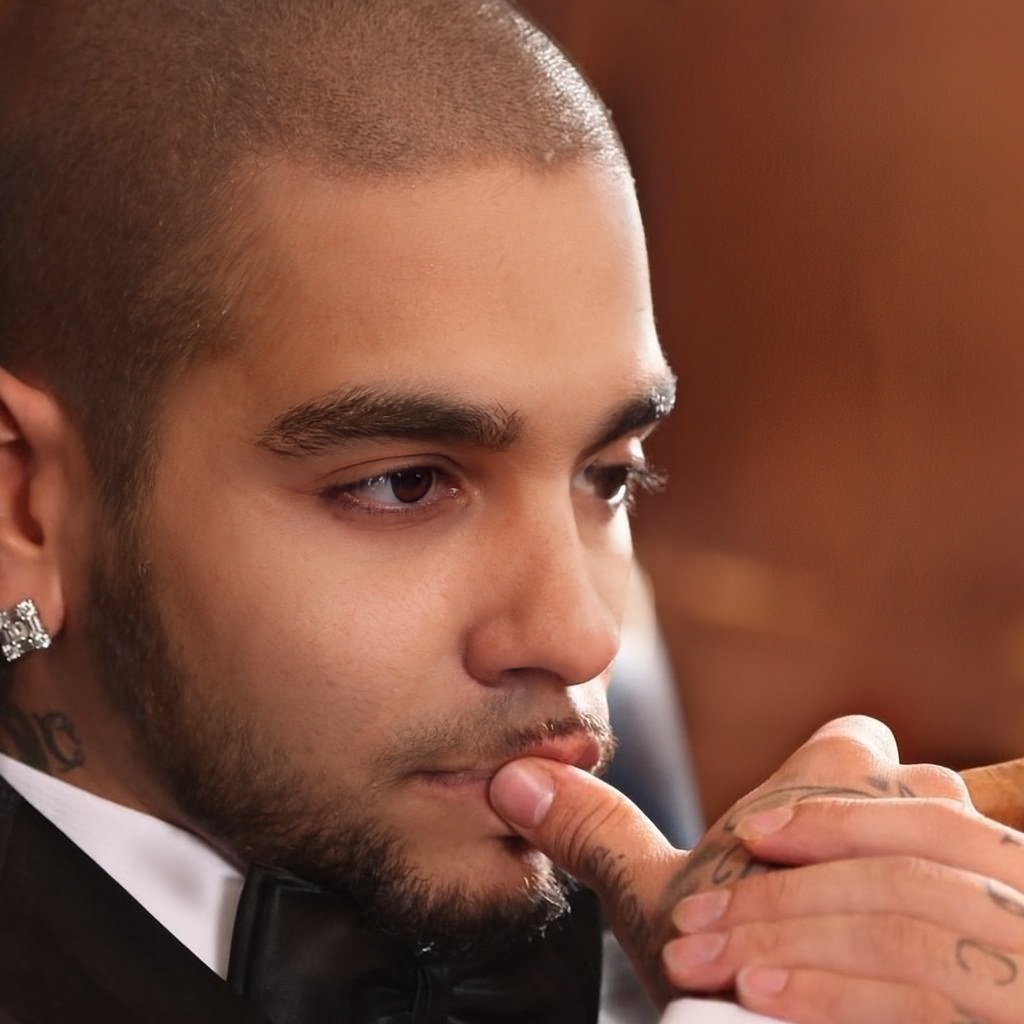} &
        \includegraphics[width=0.12\linewidth]{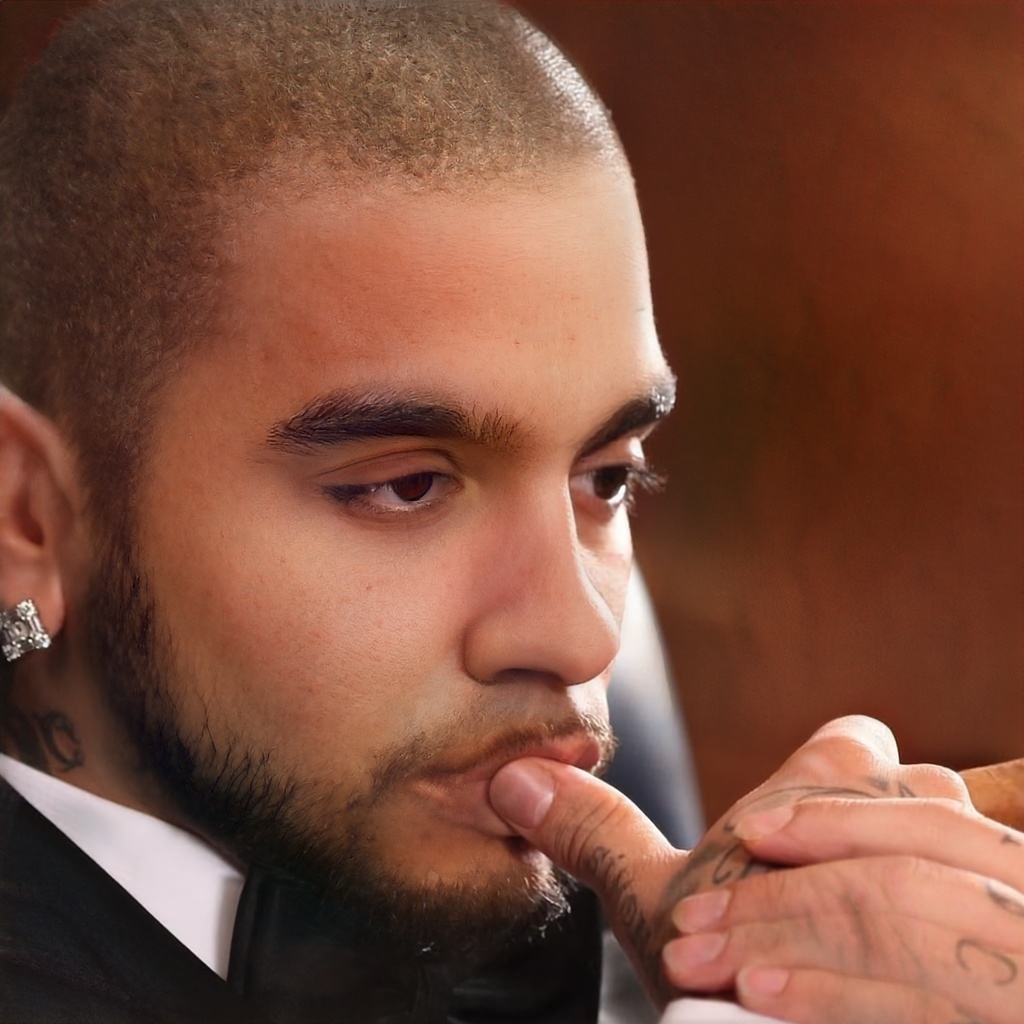}
    \end{tabular}
    \vspace{-0.5em}
    \caption{Comparison of GAN inversion quality on CelebAMask-HQ~\cite{CelebAMask-HQ} testing dataset. Our method produces the best results in the hats and backgrounds since we skip the generation of the out-of-domain contents. Meanwhile, we eliminate the spatial misalignment in the generator features to better compose out-of-domain content with generated content without ghosting artifacts.}
    \label{fig:inversion}
    \vspace{-1em}
\end{figure*}

\subsubsection{Mask Regularization}
We hope to maximize the area of $x_{in}$ to better utilize the GAN invertibility for the follow-up applications (e.g., face attribute manipulation). 
Thus, we train SAMM to produce $m$ with the maximum low-intensity area under the supervision of $L_{rec}$, since the region with low-intensity value in $M_{i, N}$ indicates only minor correction in $g_i$ is needed, which also implicitly defines ID areas in $x$. 
Inspired by \cite{bae2022furrygan}, we use a regularization loss $L_{mask}$ on $m$. $L_{mask}$ consists of the binary regularization $L_{bin}$ and the area regularization $L_{area}$:
\begin{gather}
    L_{bin}(M_{i, N}) = {\rm min}(M_{i, N}, (1-M_{i, N})), \\
    \resizebox{0.85\linewidth}{!}{$L_{area}(M_{i, N}, \phi_{area, i}) = min(0, \phi_{area} - \frac{1}{|M_{i, N}|}\sum M_{i, N})$},
\end{gather}
where $\phi_{area, i}$ is the expect OOD size in the $i$-th layer,
and $|M_{i, N}|$ is the pixel count of mask $M_{i, N}$. Finally, we have
\begin{equation}
\small
L_{mask} = \sum_{i=1}^{L} [\lambda_1 L_{bin}(M_{i, N}) + L_{area}(M_{i, N}, \phi_{area,i})],
\end{equation}
where $\lambda_1$ is the loss weight for $L_{bin}$.
In summary, our overall objective $L_{total}$ is:
\begin{equation}
    \label{eq:total_loss}
    L_{total} = L_{rec} + L_{adv} + L_{mask}.
\end{equation}

We found that if we loosen the $L_{mask}$ with a larger $\phi_{area}$ and a smaller $\lambda_1$, the predicted $m$ has a high value (close to 1) at most pixels except for the eyes and mouth area. 
Consequently, the reconstruction error of $\widehat{x}$ can be optimized to be low, while the editability must be harmed. As we hope to maintain the off-the-shelf editability of styleGAN generator, in this paper, we set $\phi_{area} = 0.3$ for $32^2, 64^2$ masks and $\phi_{area} = 0.25$ for $128^2, 256^2$ masks.


\begin{figure*}[t]
    \centering
    \begin{tabular}{@{}c@{\hspace{3mm}}c@{\hspace{3mm}}c@{\hspace{3mm}}c@{\hspace{3mm}}c@{\hspace{3mm}}c@{\hspace{3mm}}c@{}}
        \footnotesize{Input} & \footnotesize{Inversion}  & \footnotesize{+Age} & \footnotesize{+Beard} & \footnotesize{-Smile} & \footnotesize{+Thick eyebrows}
        \\
        \includegraphics[width=0.13\linewidth]{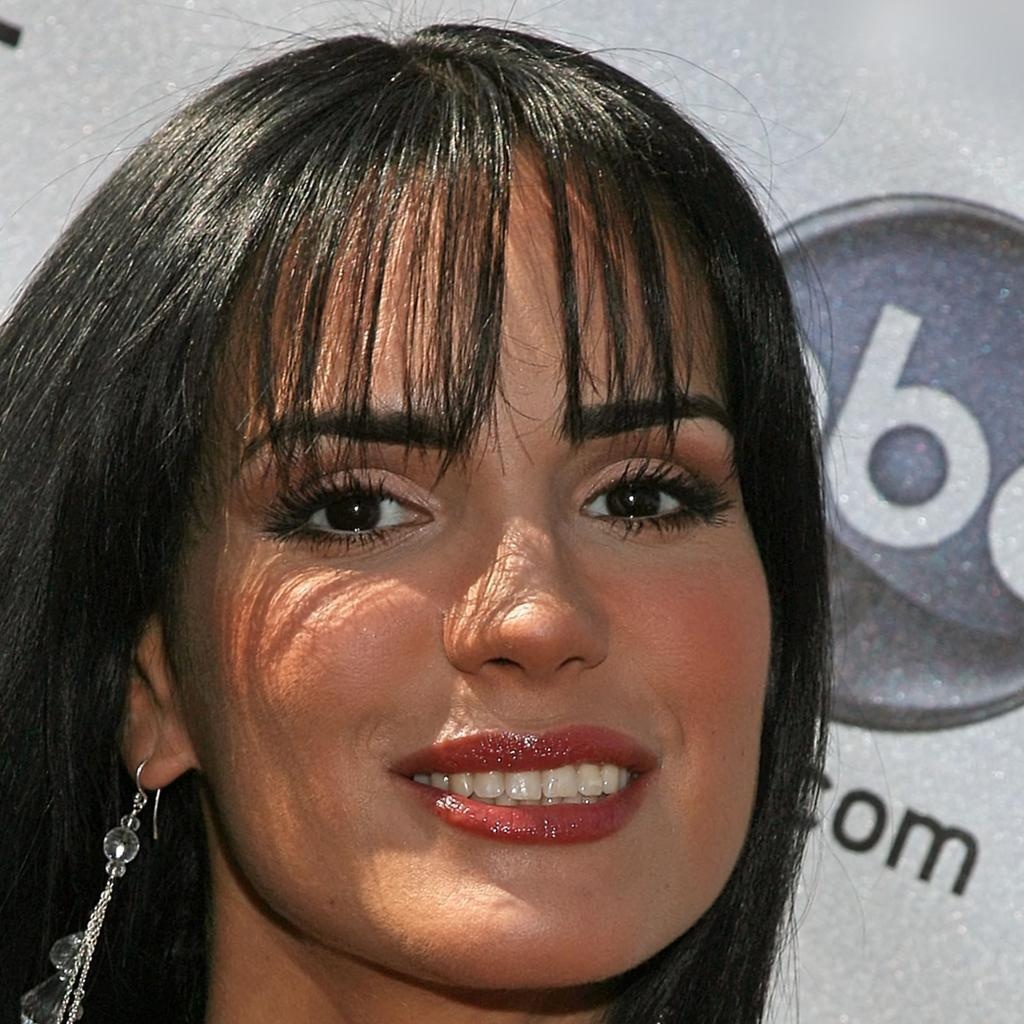} &
        \includegraphics[width=0.13\linewidth]{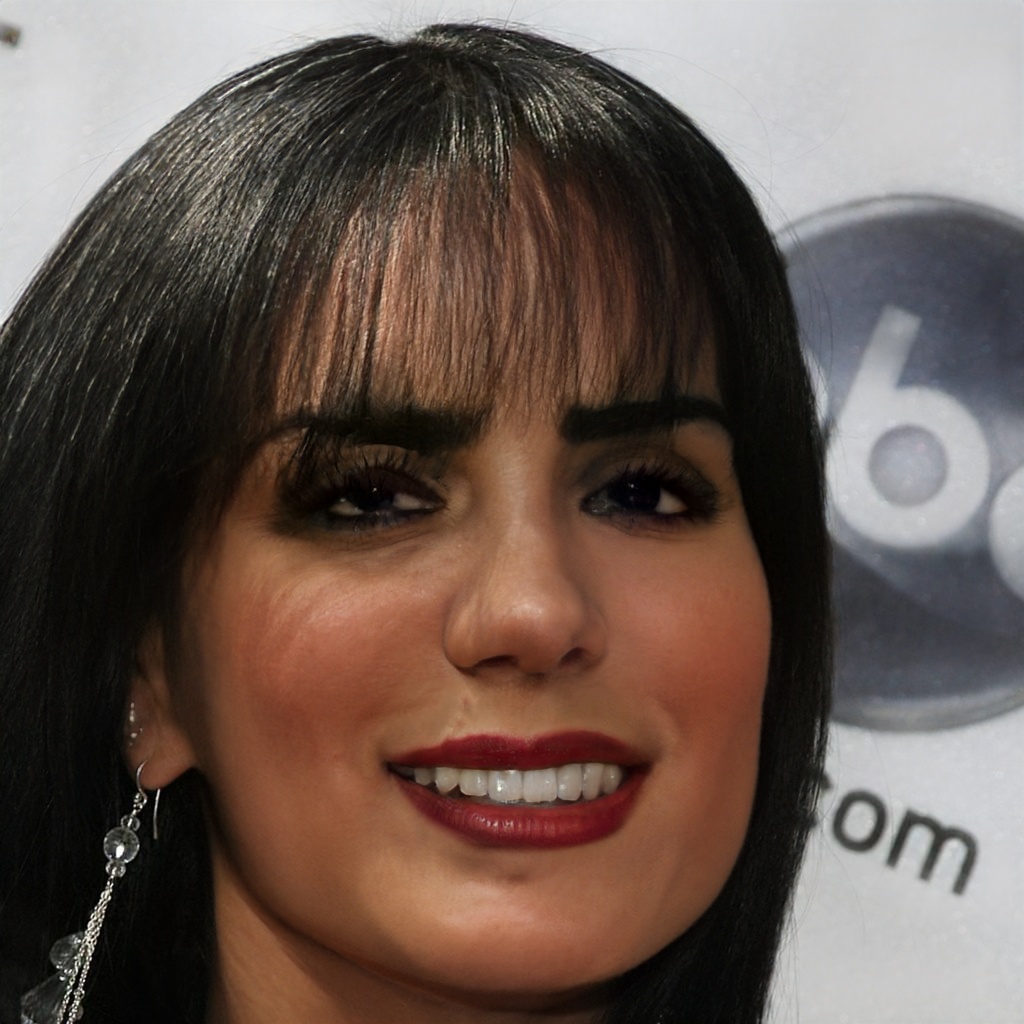} &
        \includegraphics[width=0.13\linewidth]{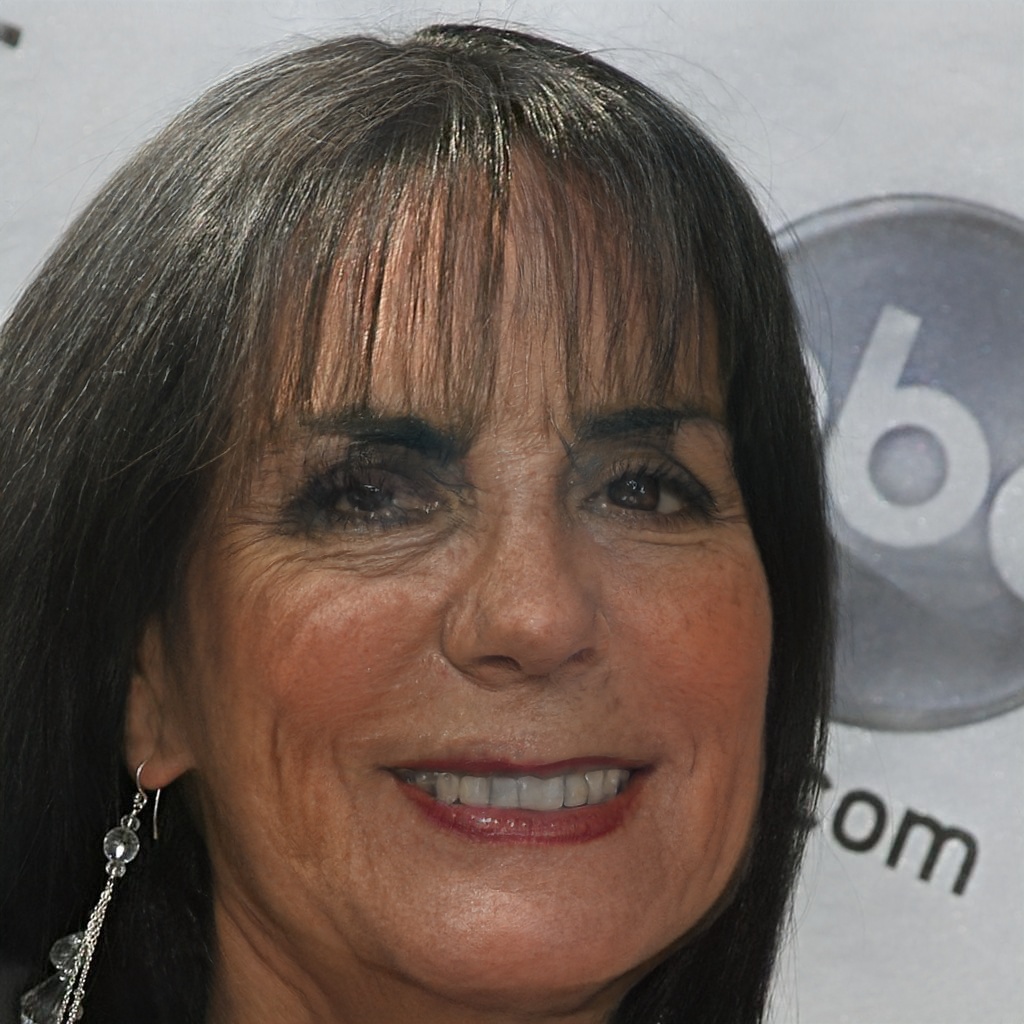} & 
        \includegraphics[width=0.13\linewidth]{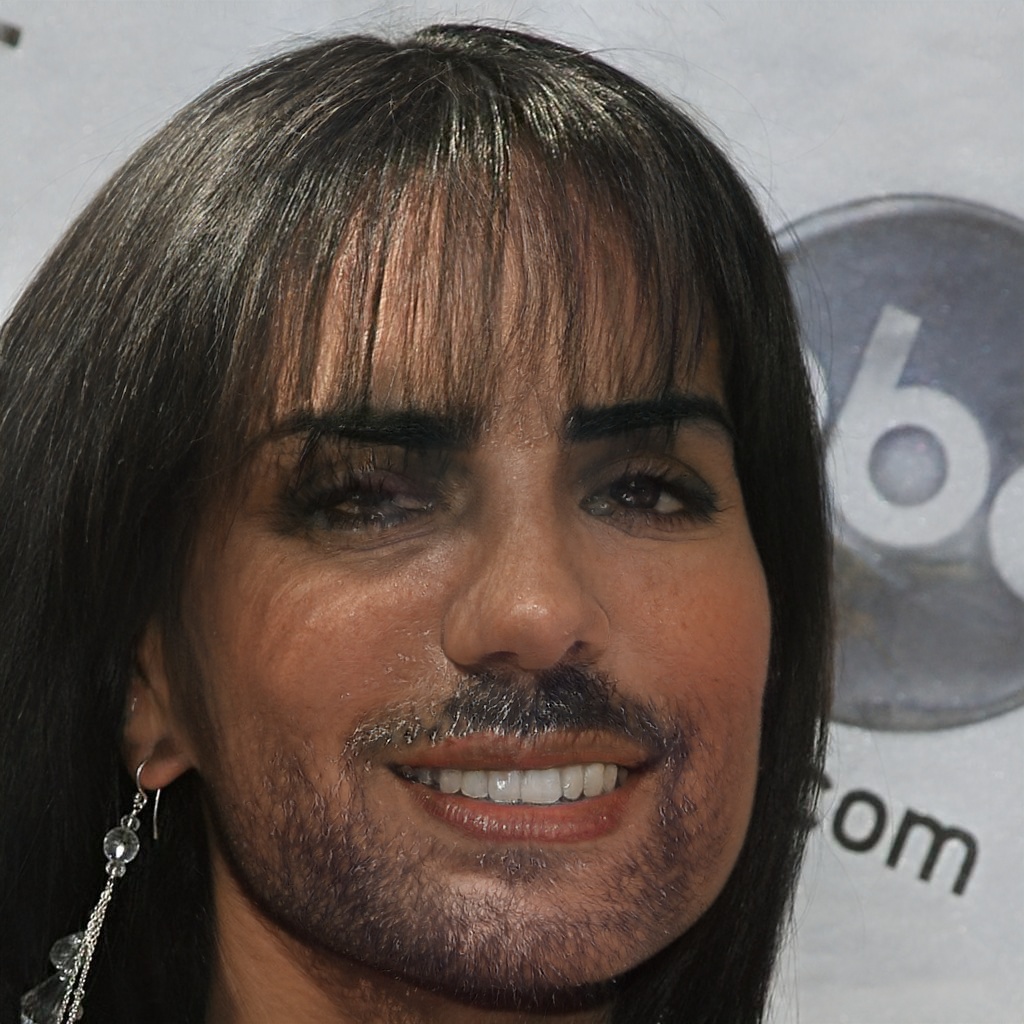} &
        \includegraphics[width=0.13\linewidth]{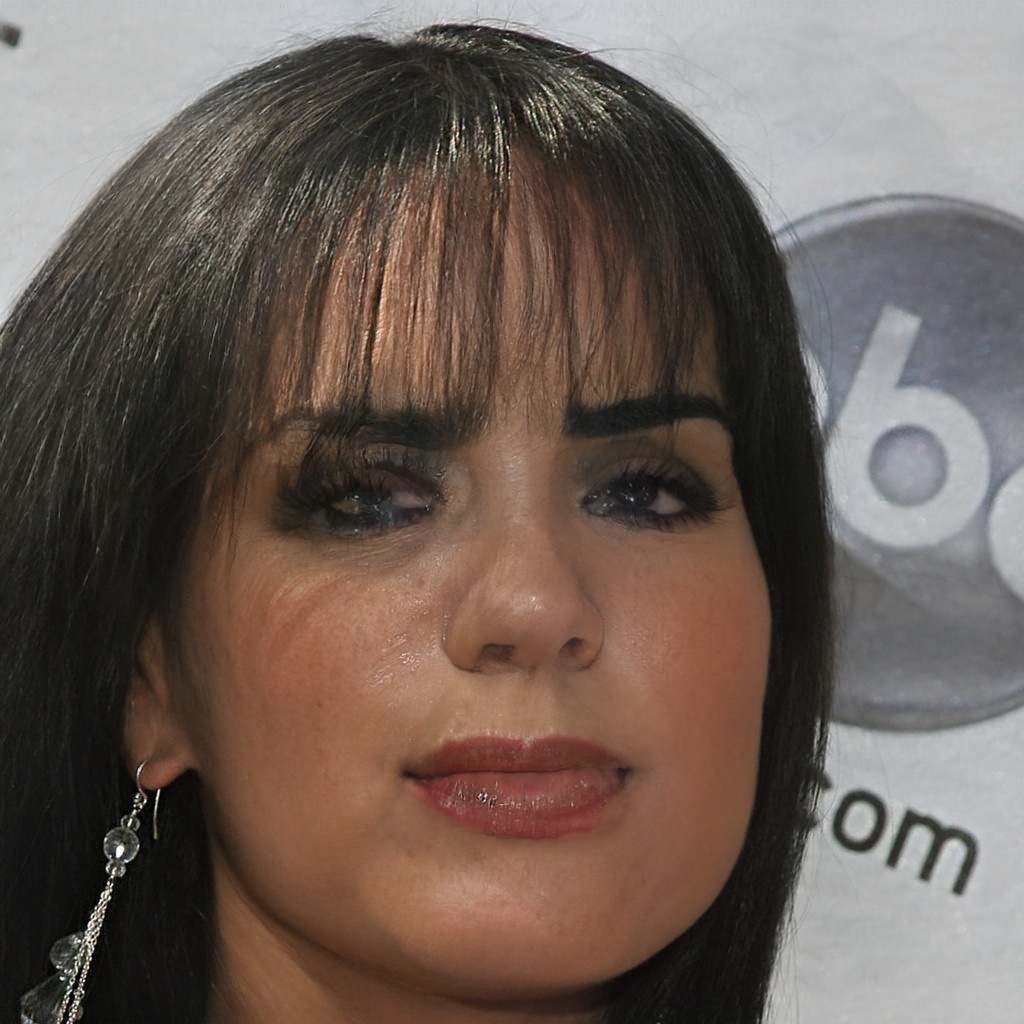} &
        \includegraphics[width=0.13\linewidth]{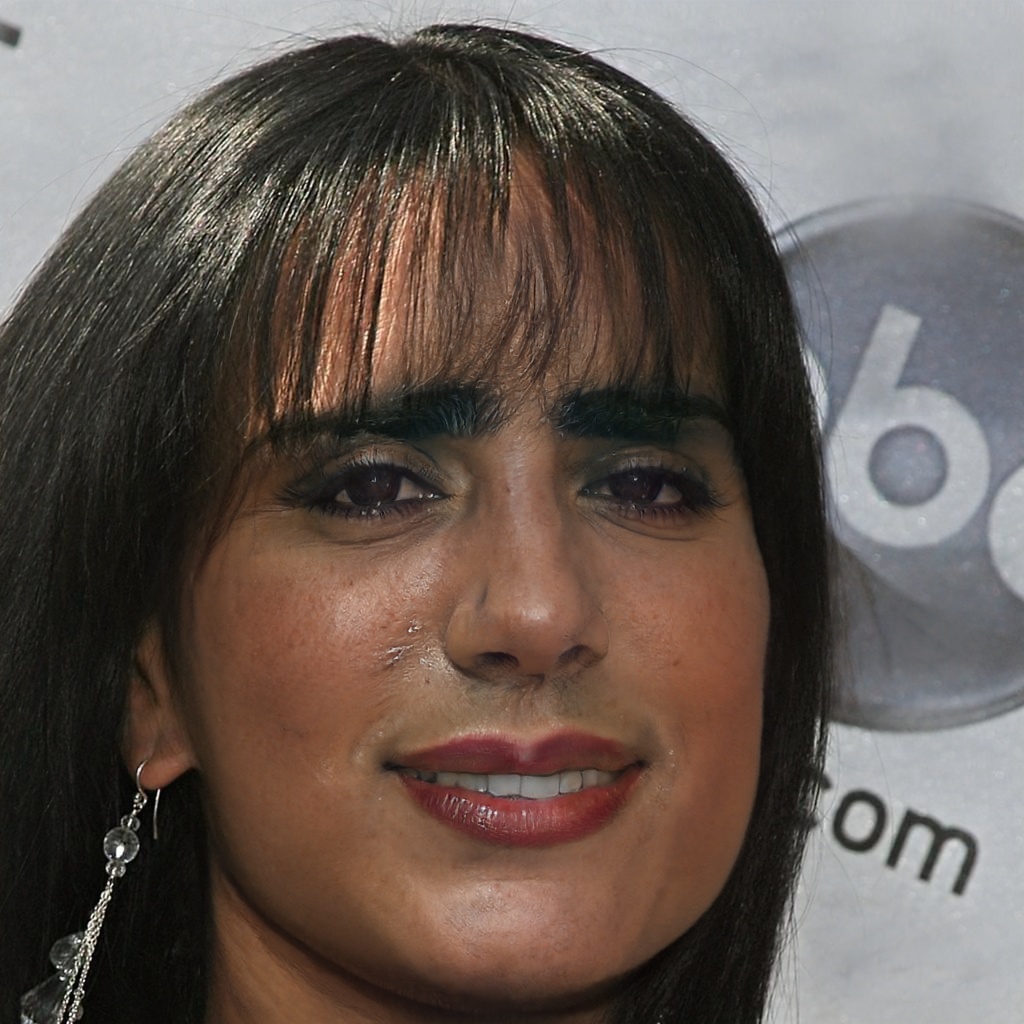}
        \\
        \includegraphics[width=0.13\linewidth]{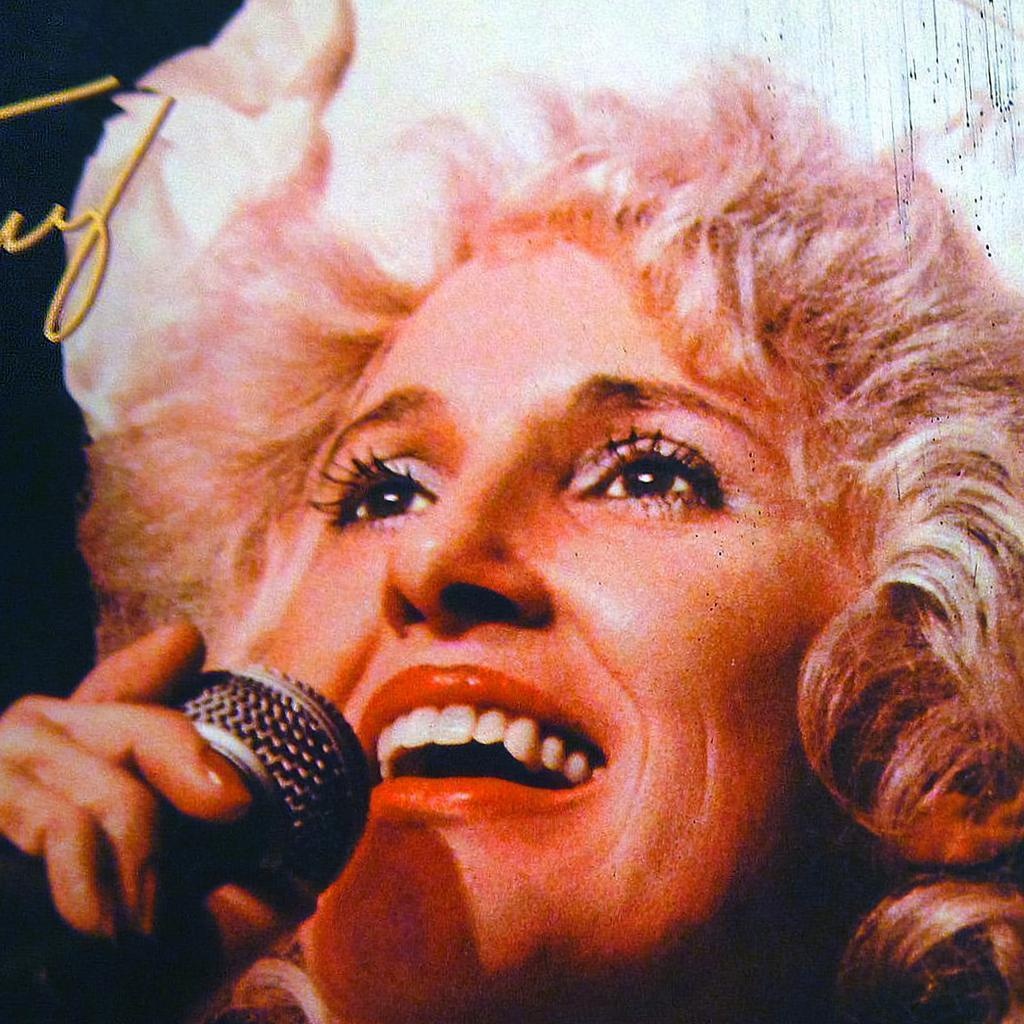} &
        \includegraphics[width=0.13\linewidth]{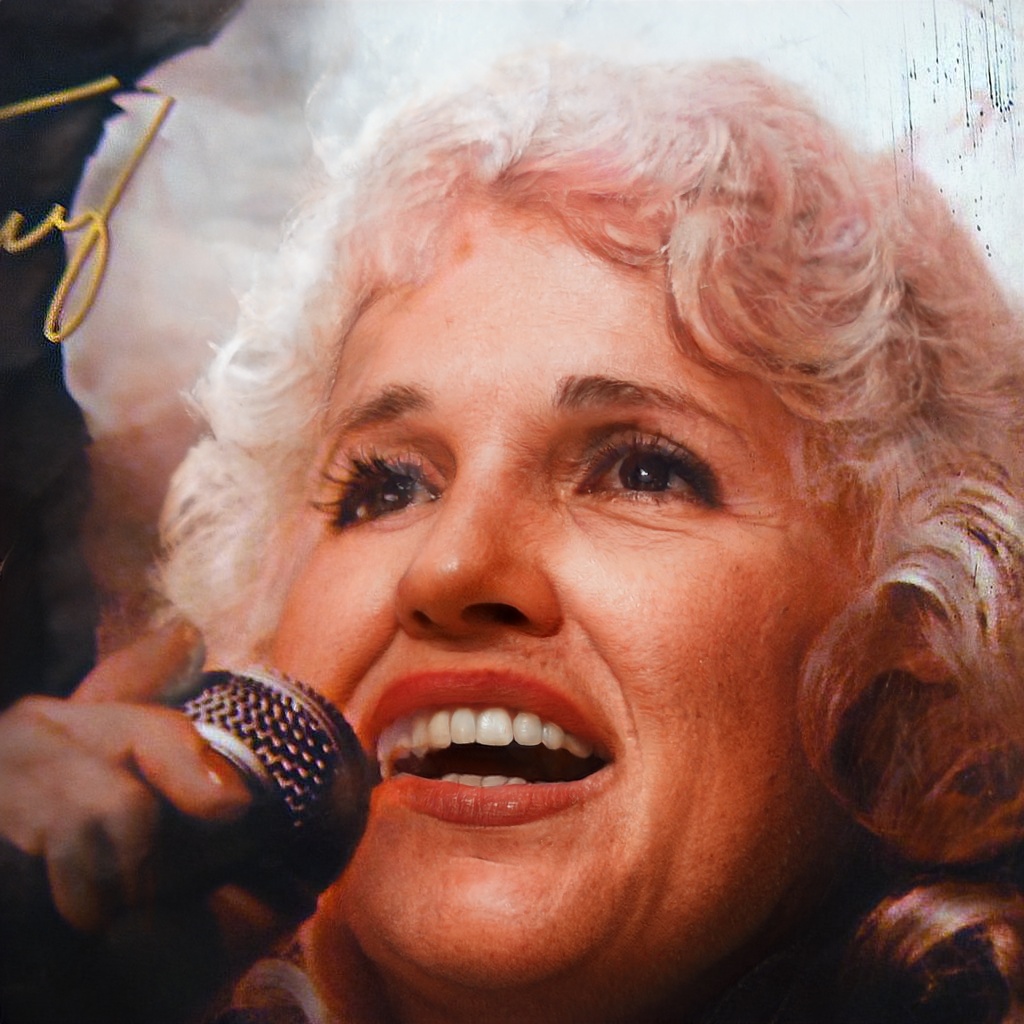} &
        \includegraphics[width=0.13\linewidth]{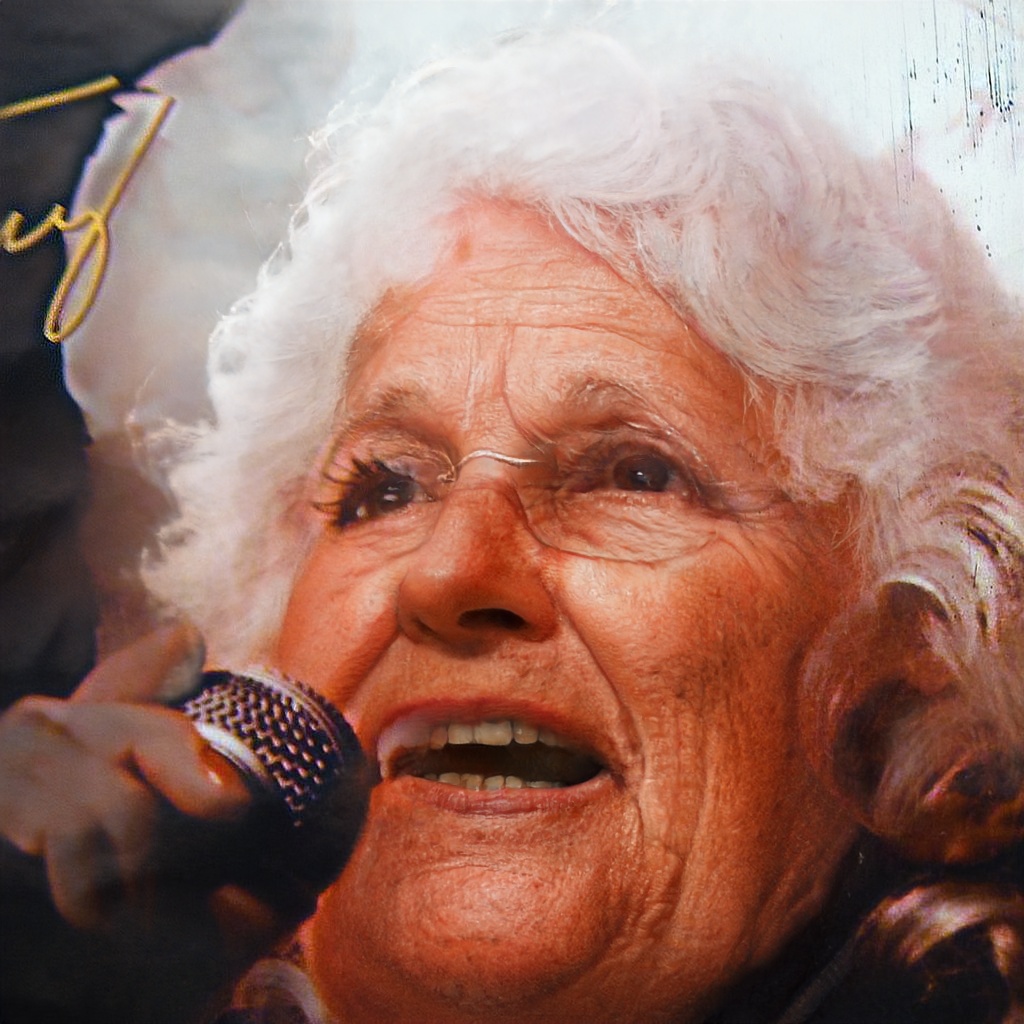} & 
        \includegraphics[width=0.13\linewidth]{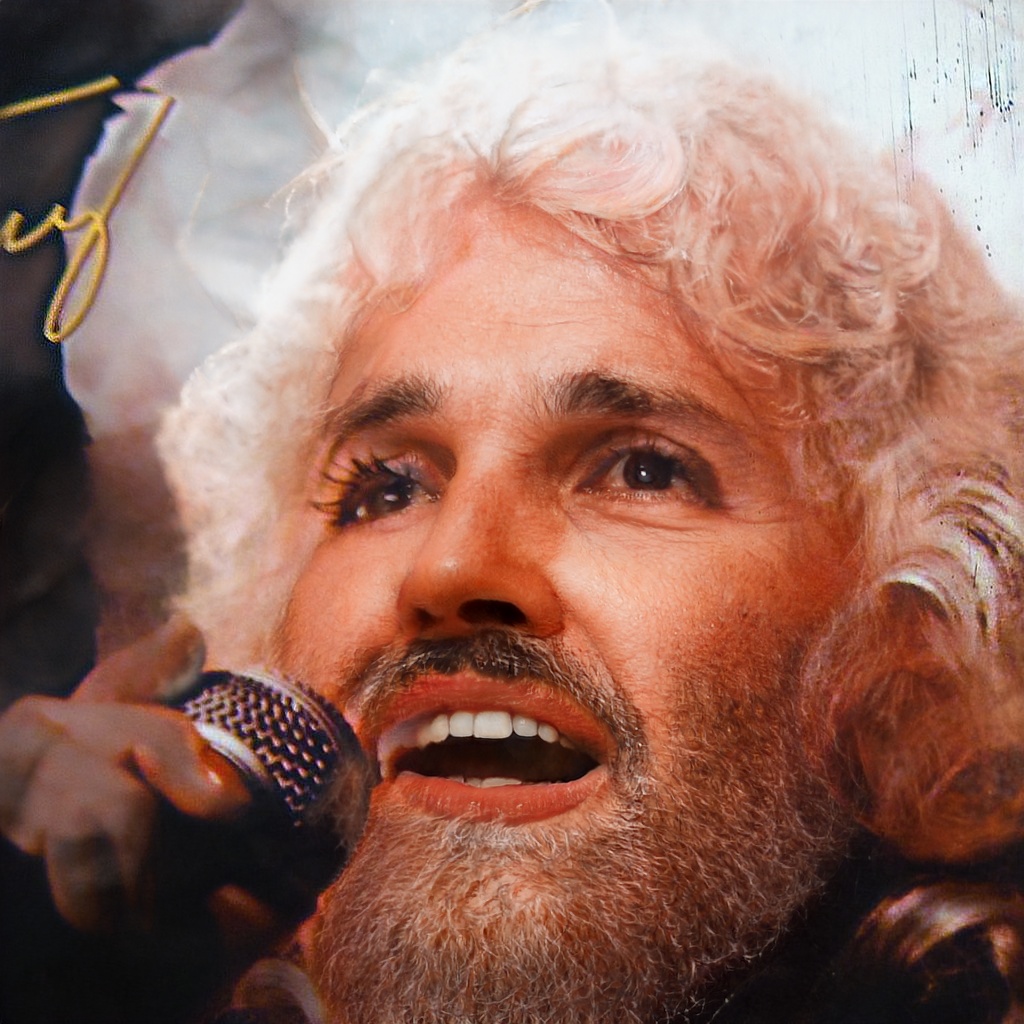} &
        \includegraphics[width=0.13\linewidth]{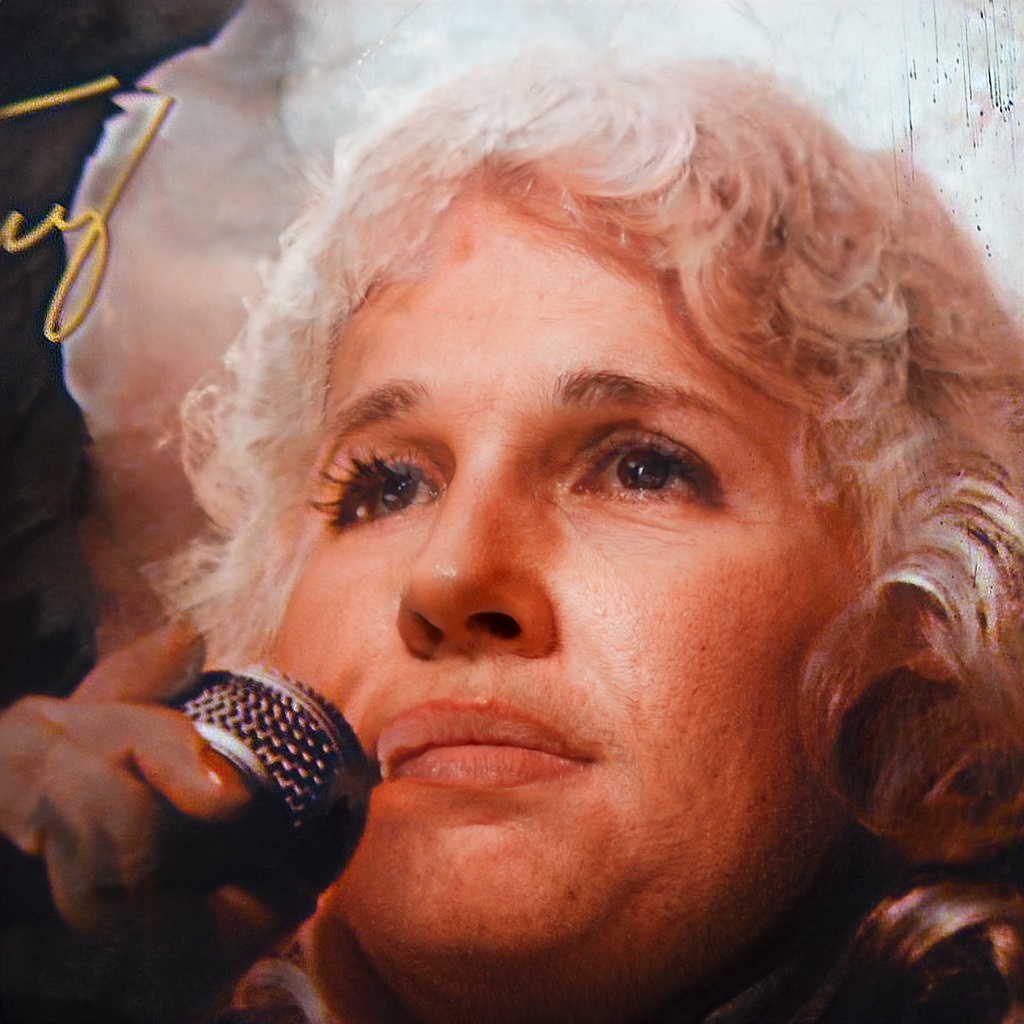} &
        \includegraphics[width=0.13\linewidth]{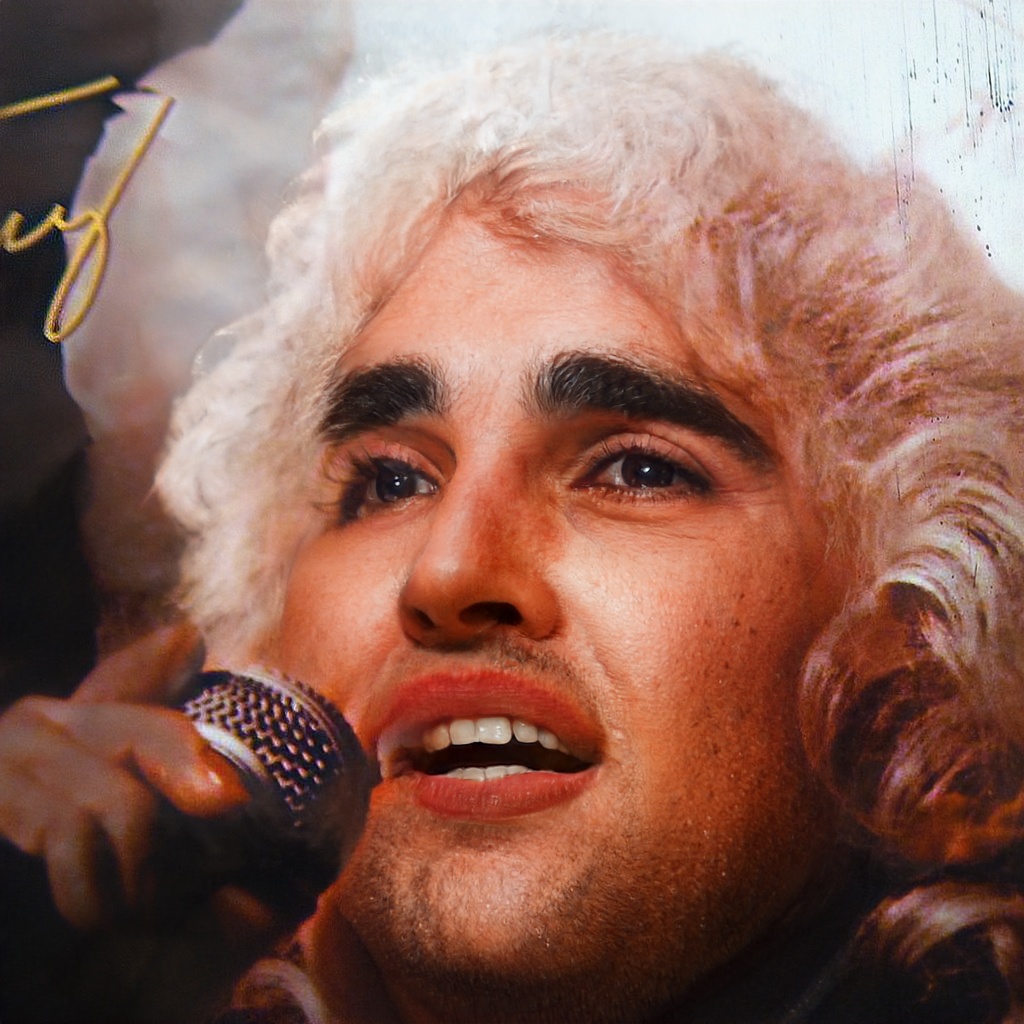}
        \\
        \includegraphics[width=0.13\linewidth]{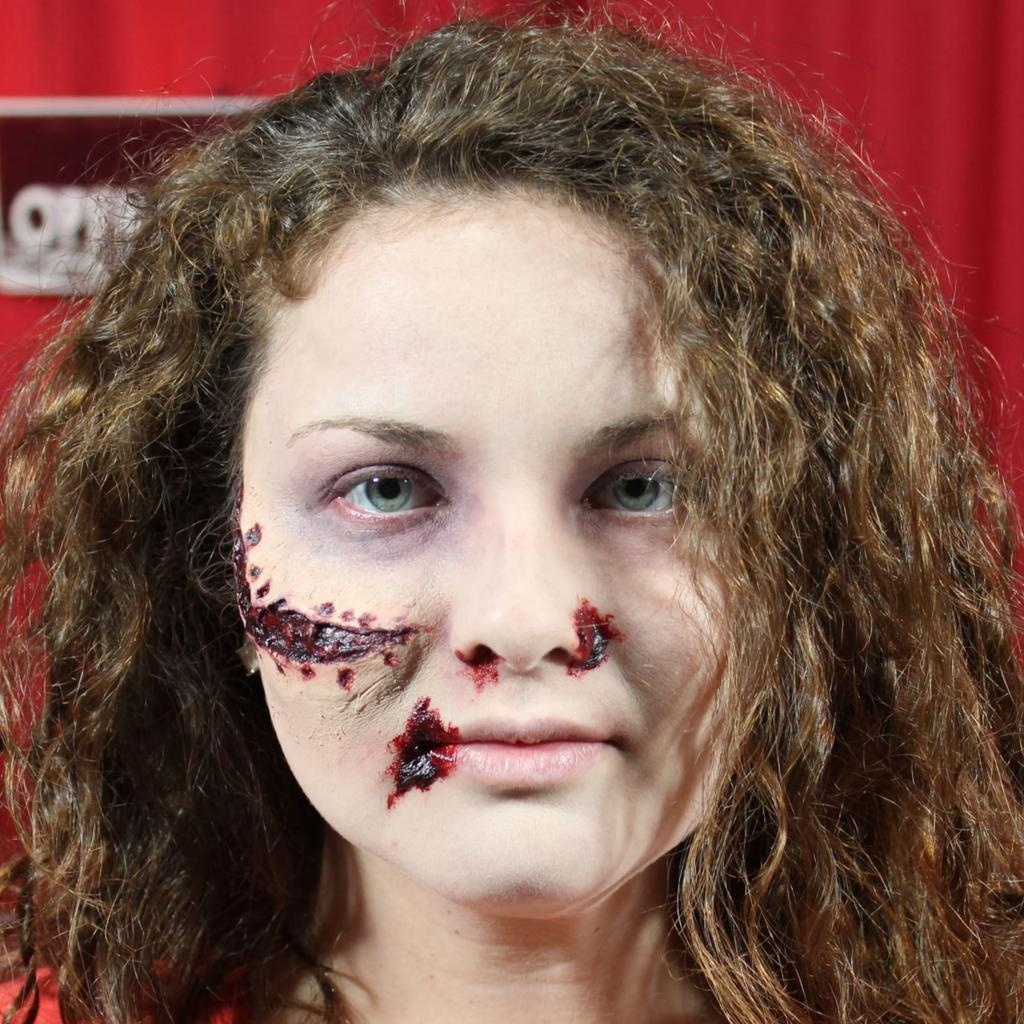} &
        \includegraphics[width=0.13\linewidth]{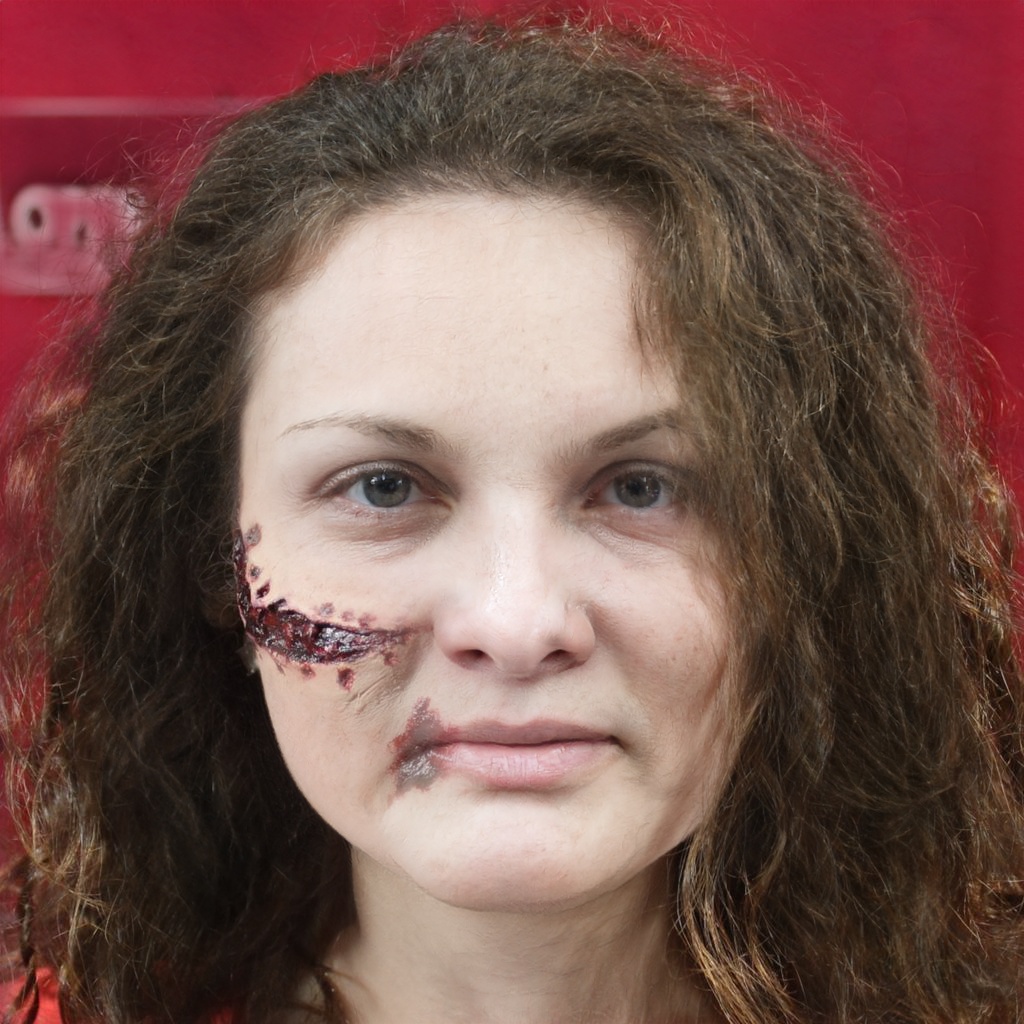} &
        \includegraphics[width=0.13\linewidth]{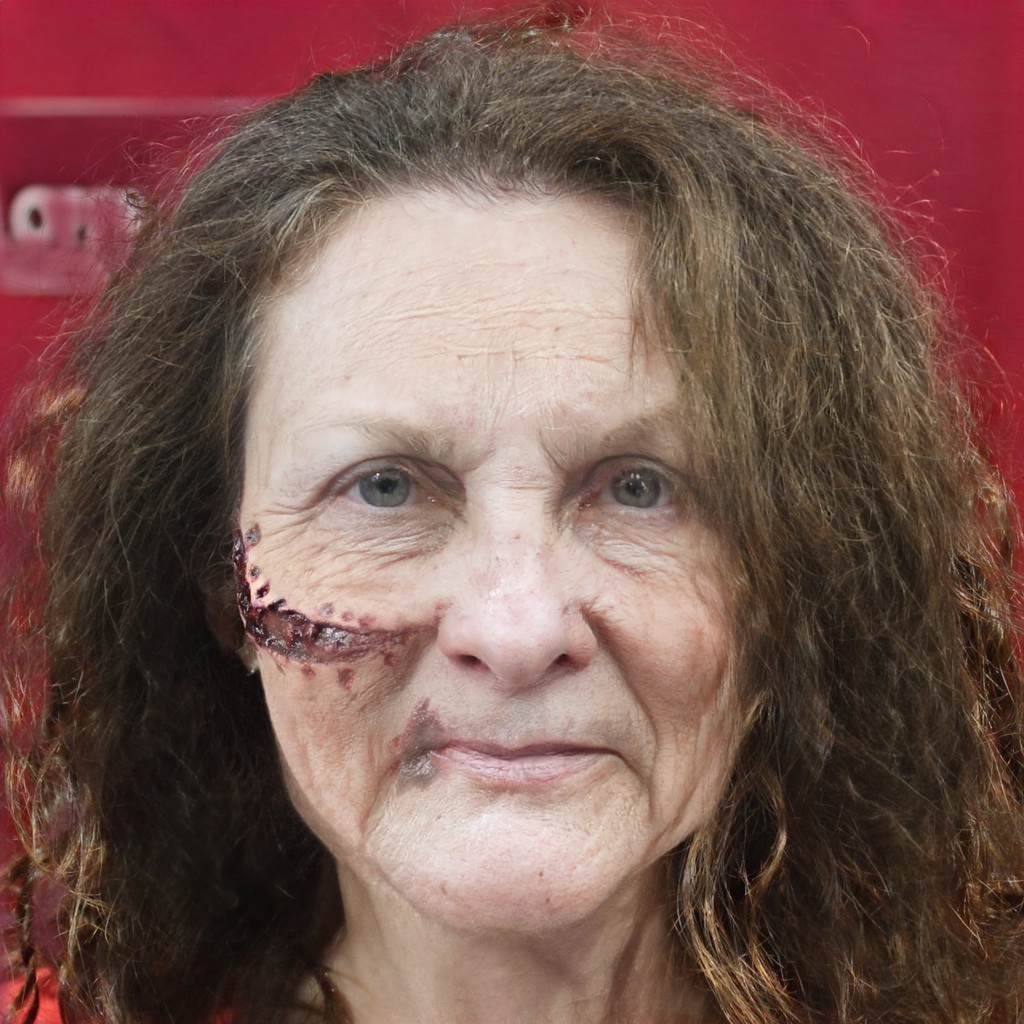} & 
        \includegraphics[width=0.13\linewidth]{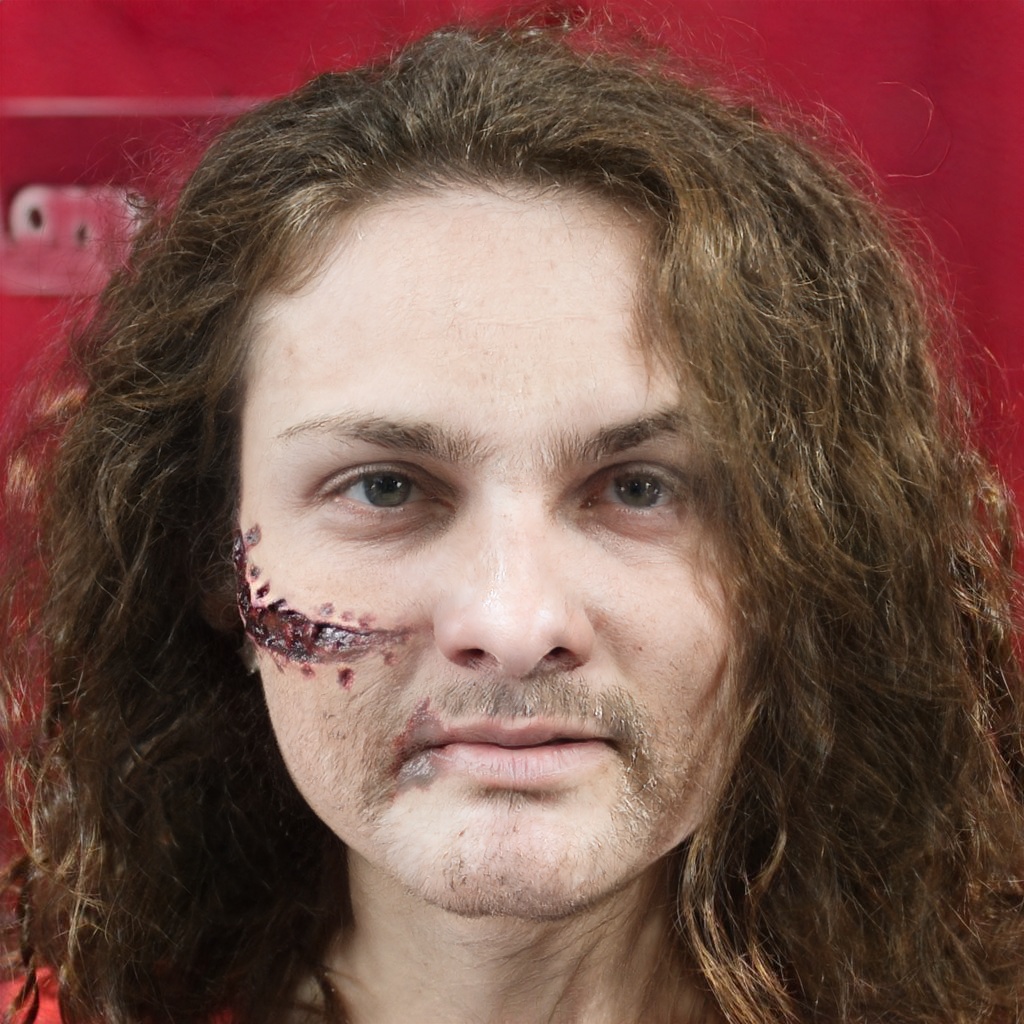} &
        \includegraphics[width=0.13\linewidth]{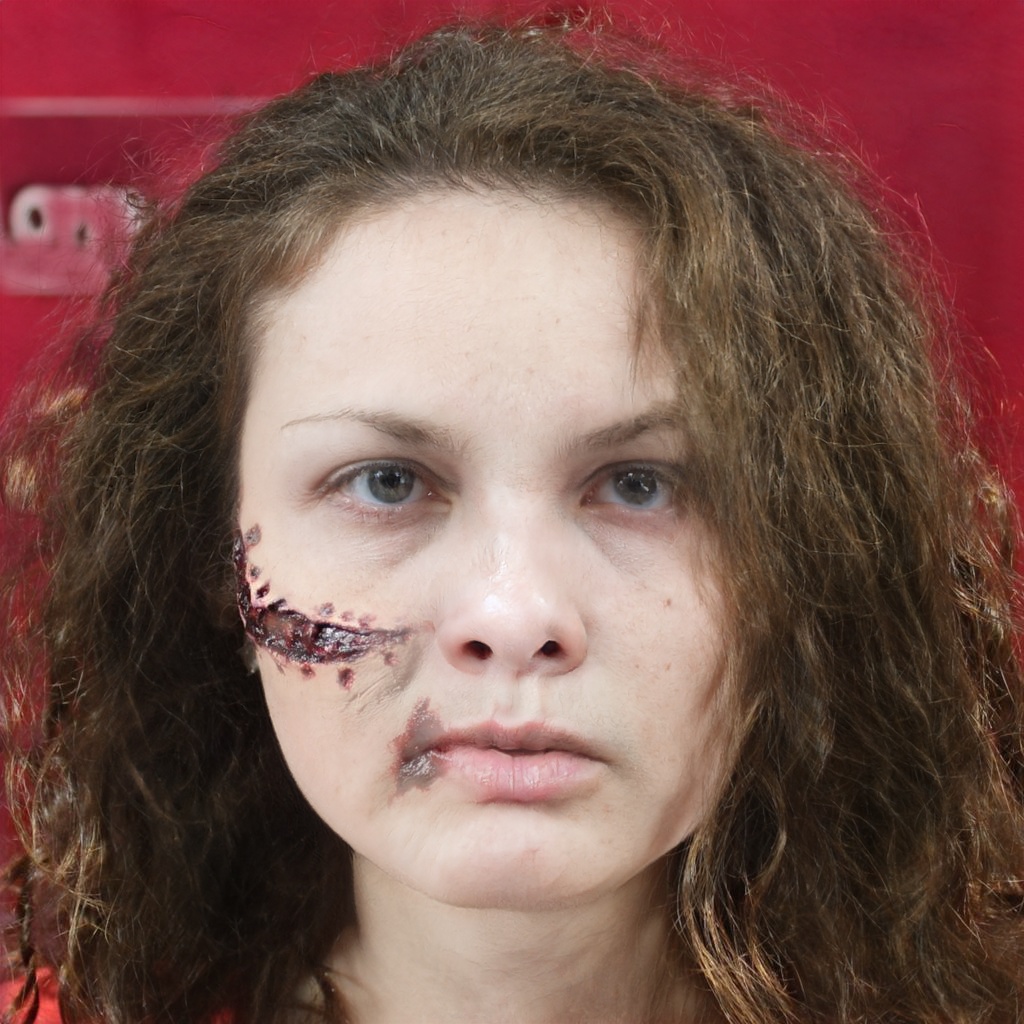} &
        \includegraphics[width=0.13\linewidth]{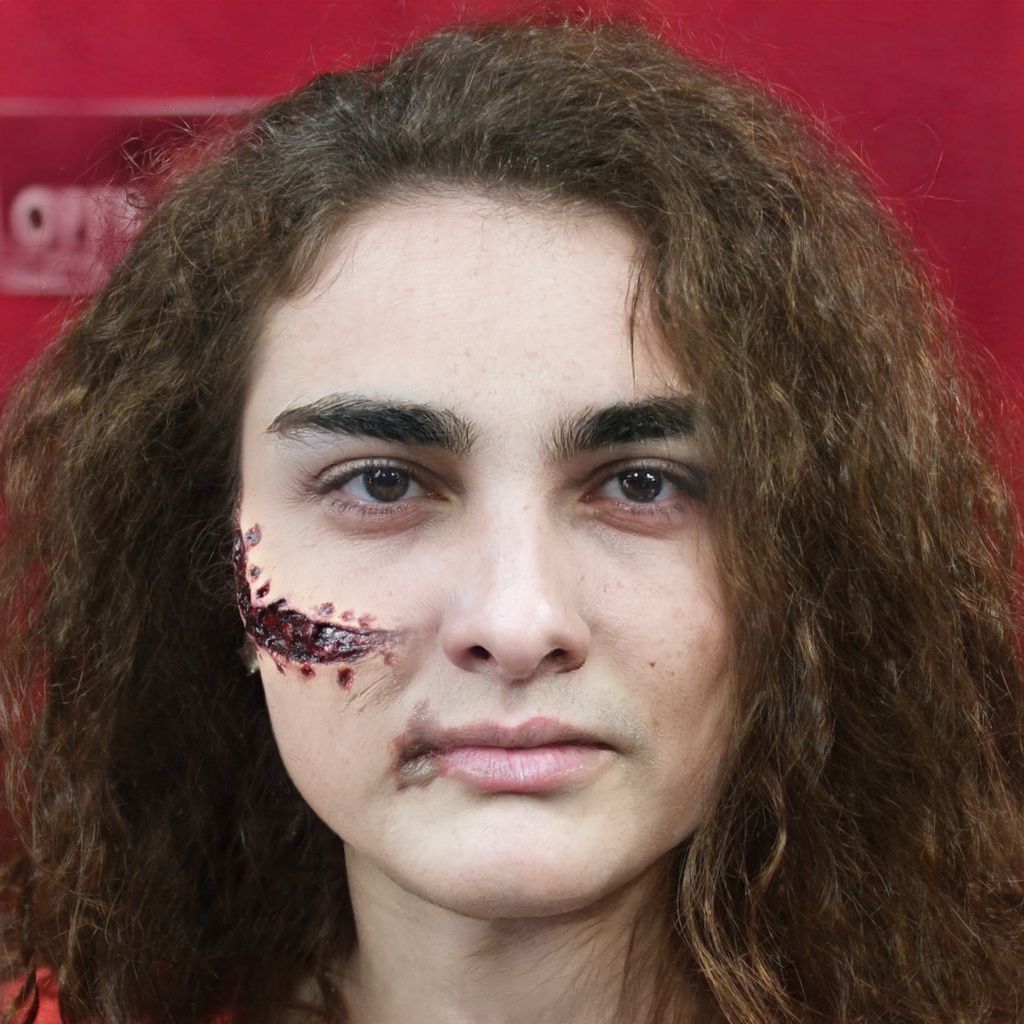} 
    \end{tabular}
    \vspace{-0.5em}
    \caption{Face manipulation results of our framework on CelebAMask-HQ~\cite{CelebAMask-HQ} dataset, please zoom in for detail.}
    \label{fig:editing}
    \vspace{-1.3em}
\end{figure*}

\begin{figure}[t]
    \centering
    \begin{tabular}{@{}c@{\hspace{2mm}}c@{\hspace{2mm}}c@{}}
        \scriptsize{Input} & 
        \scriptsize{Inversion} & 
        \scriptsize{$\rightarrow$``Pink hair"}
        \\
        \includegraphics[width=0.3\linewidth]{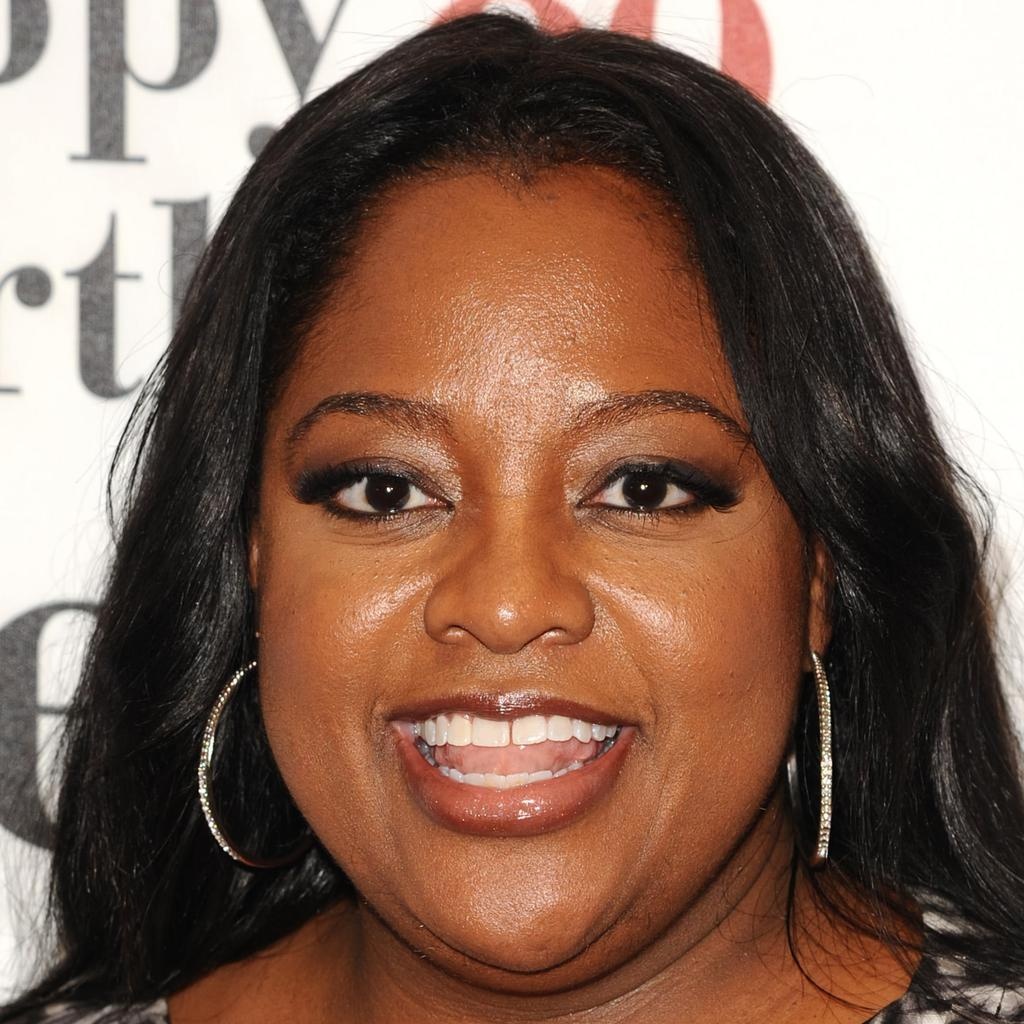} &
        \includegraphics[width=0.3\linewidth]{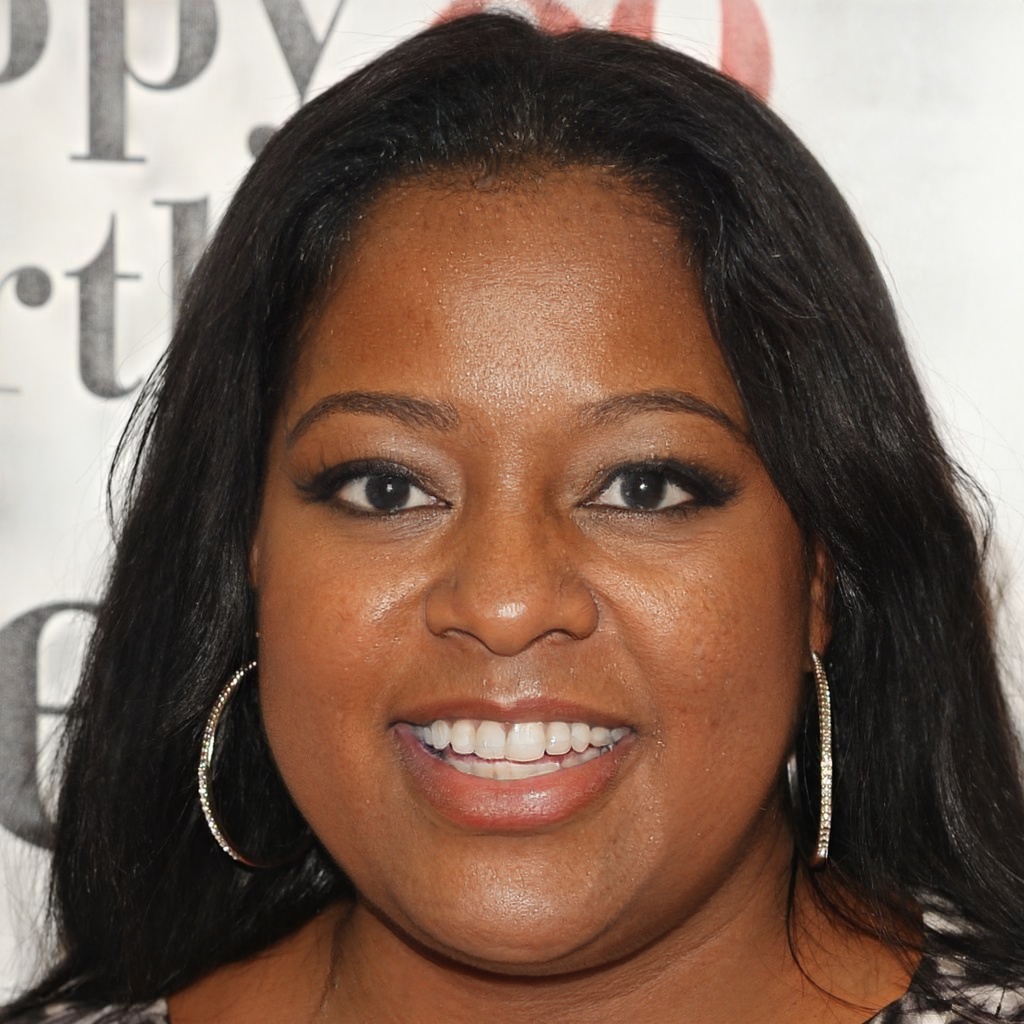} &
        \includegraphics[width=0.3\linewidth]{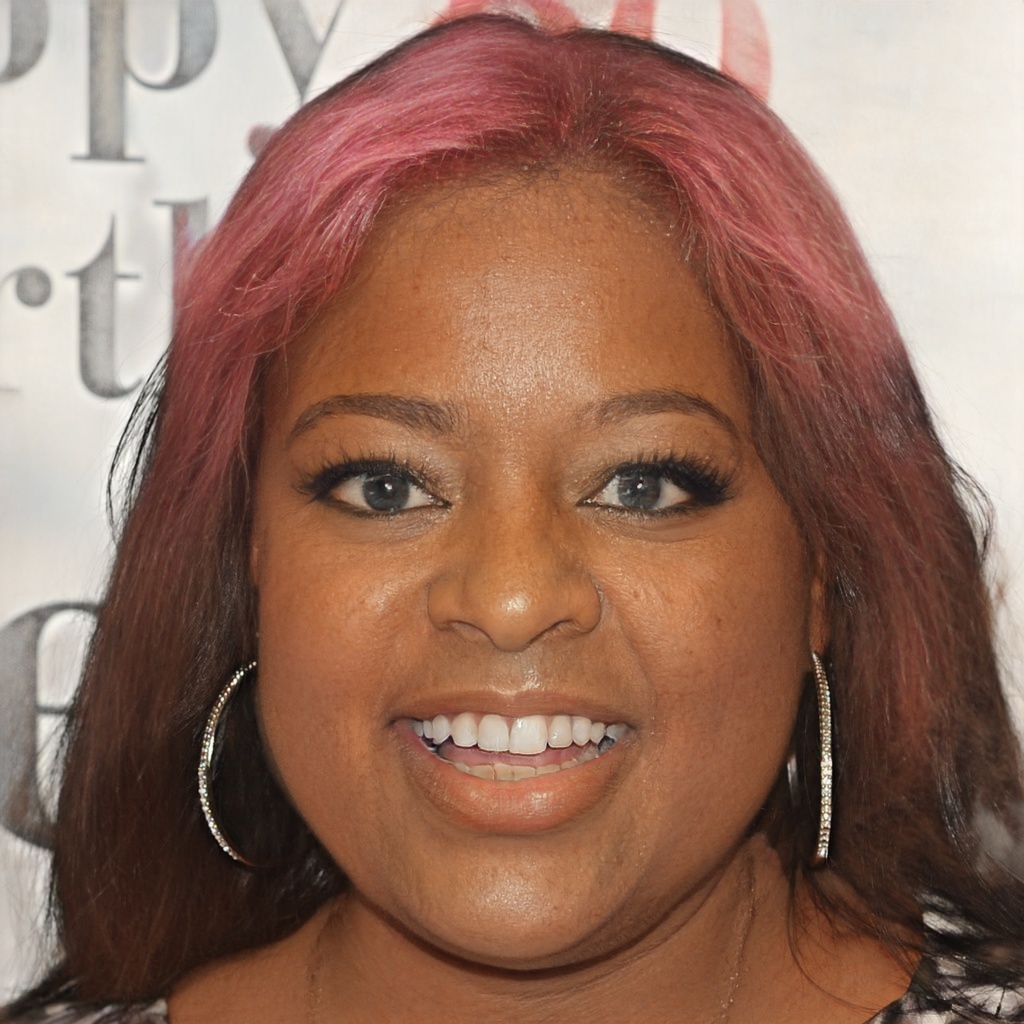}
        \\
    \end{tabular}
    \vspace{-0.5em}
    \caption{\textbf{Edit with text guidance.} Our method also support off-the-shelf attribute manipulation with text guidance~\cite{patashnik2021styleclip} for in-domain area (i.e., hair color). }
    \label{fig:clip_edit}
    \vspace{-1.3em}
\end{figure}

\section{Experiments}
In this paper, we adopt the official checkpoints of e4e~\cite{tov2021designing} or ReStyle~\cite{alaluf2021restyle} encoder $E$ and StyleGAN2~\cite{karras2020analyzing} (config-f) generator $G$, which were pretrained on the FFHQ~\cite{2018StyleGAN} dataset, and fixed their parameters. 
Then, we train our SAMM with $E$ and $G$ on FFHQ for GAN inversion to minimize $L_{total}$ (Eqn.~\ref{eq:total_loss}). 
Please refer to our supplement for more training details.

\begin{figure*}[tp]
    \centering
    \begin{tabular}{@{}c@{\hspace{1mm}}c@{\hspace{1mm}}c@{\hspace{1mm}}c@{\hspace{1mm}}c@{\hspace{1mm}}c@{\hspace{1mm}}c@{\hspace{1mm}}c@{}}
        \scriptsize{Input $\rightarrow$ \textbf{- Age}} & 
        \scriptsize{HFGI$_{e4e}$~\cite{wang2021high}}  &
        \scriptsize{HyperStyle~\cite{alaluf2021HyperStyle}}  &
        \scriptsize{SAM~\cite{parmar2022spatially}, iter=500}  &
        \scriptsize{FeatureStyle~\cite{xuyao2022}}  &
        \scriptsize{DiffCAM~\cite{song2022editing}}  &
        \scriptsize{Ours$_{e4e}$}
        \\
        \includegraphics[width=0.13\linewidth]{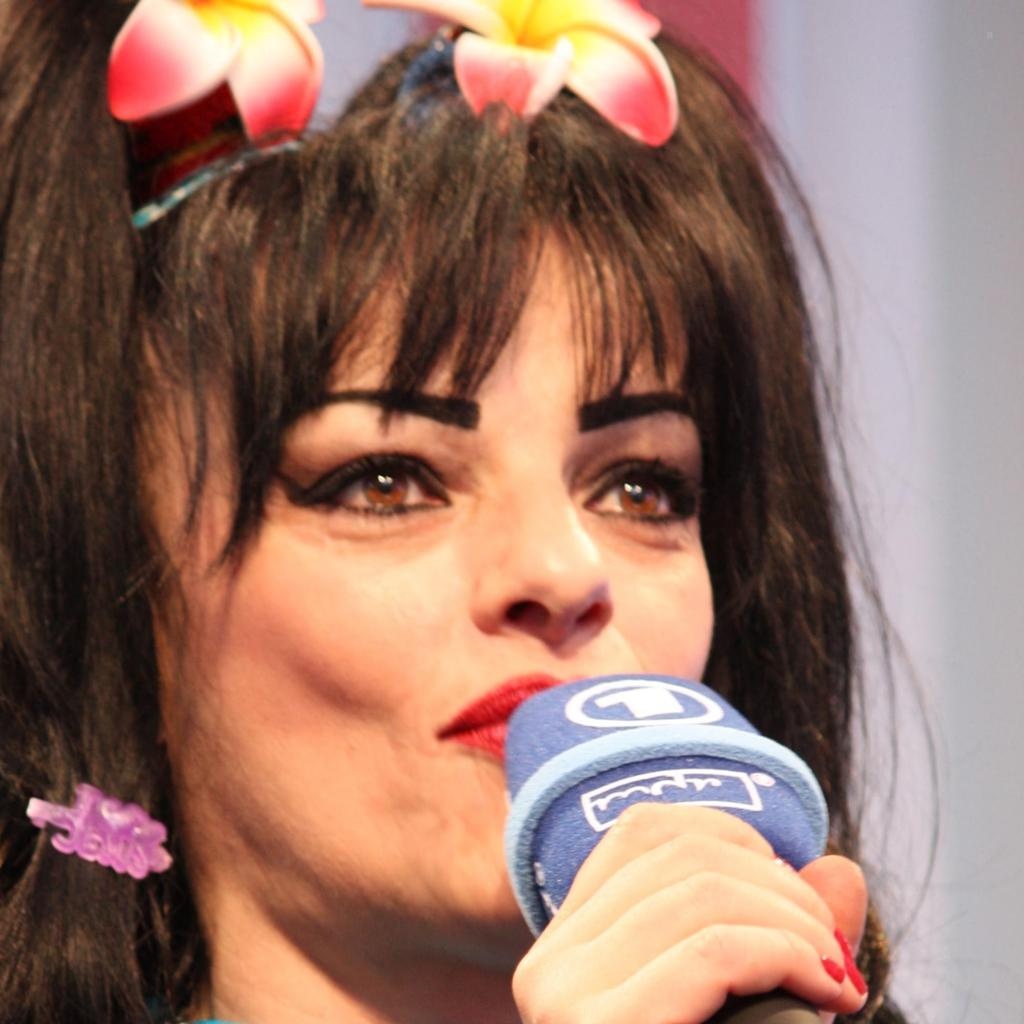} &
        \includegraphics[width=0.13\linewidth]{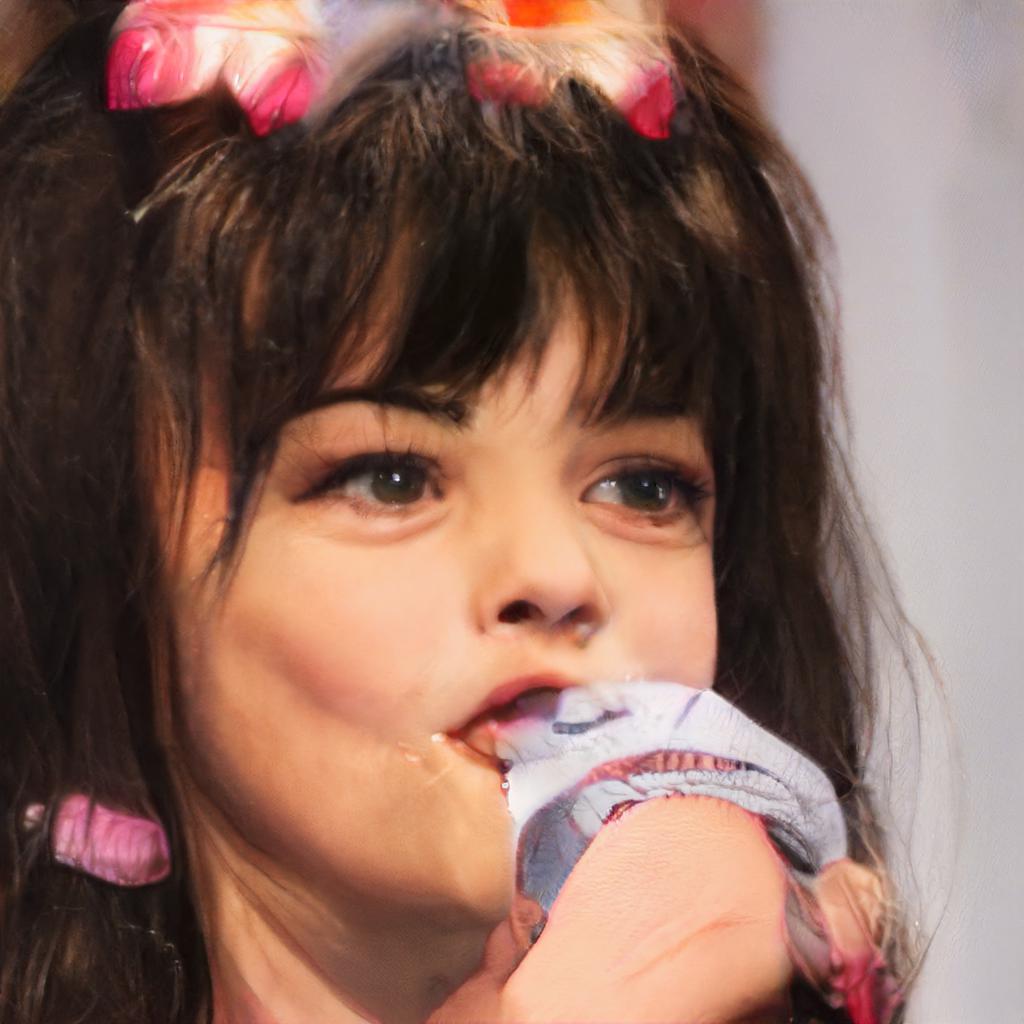} &
        \includegraphics[width=0.13\linewidth]{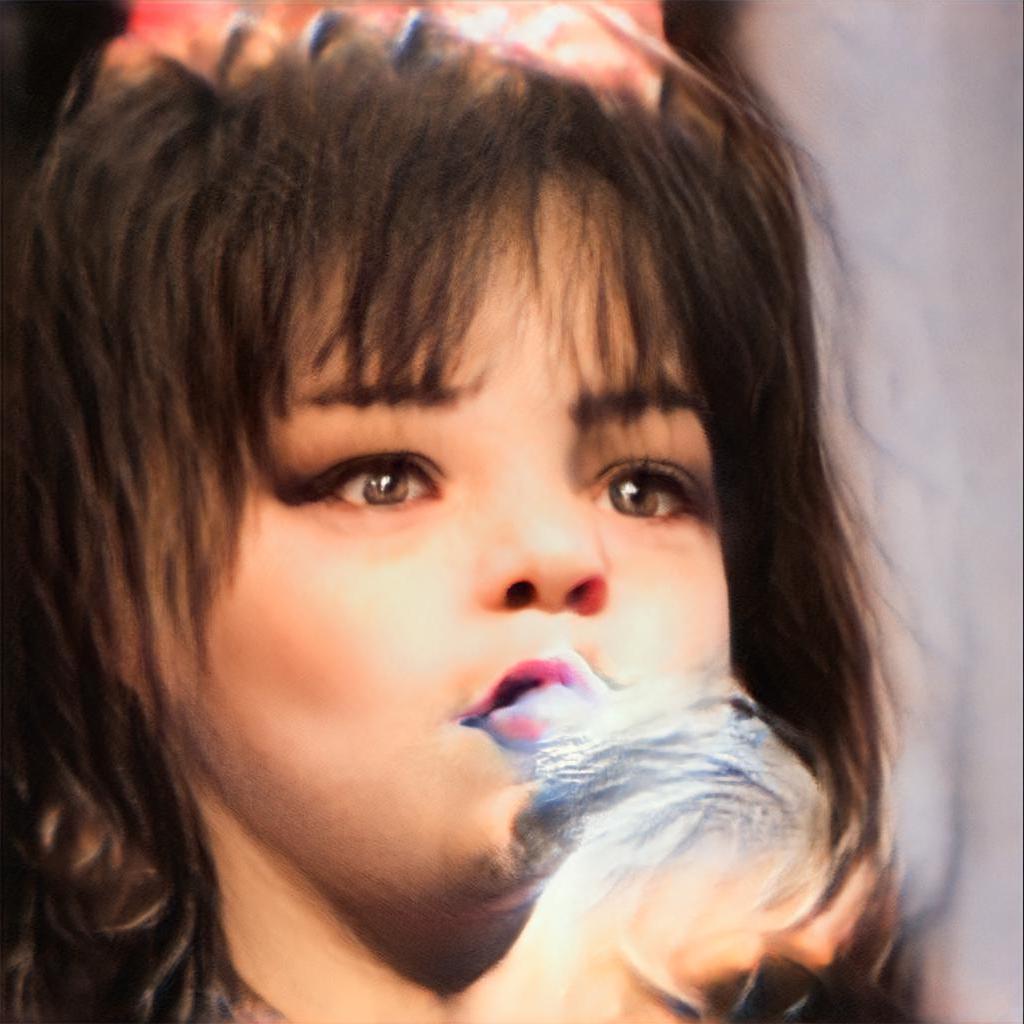} &
        \includegraphics[width=0.13\linewidth]{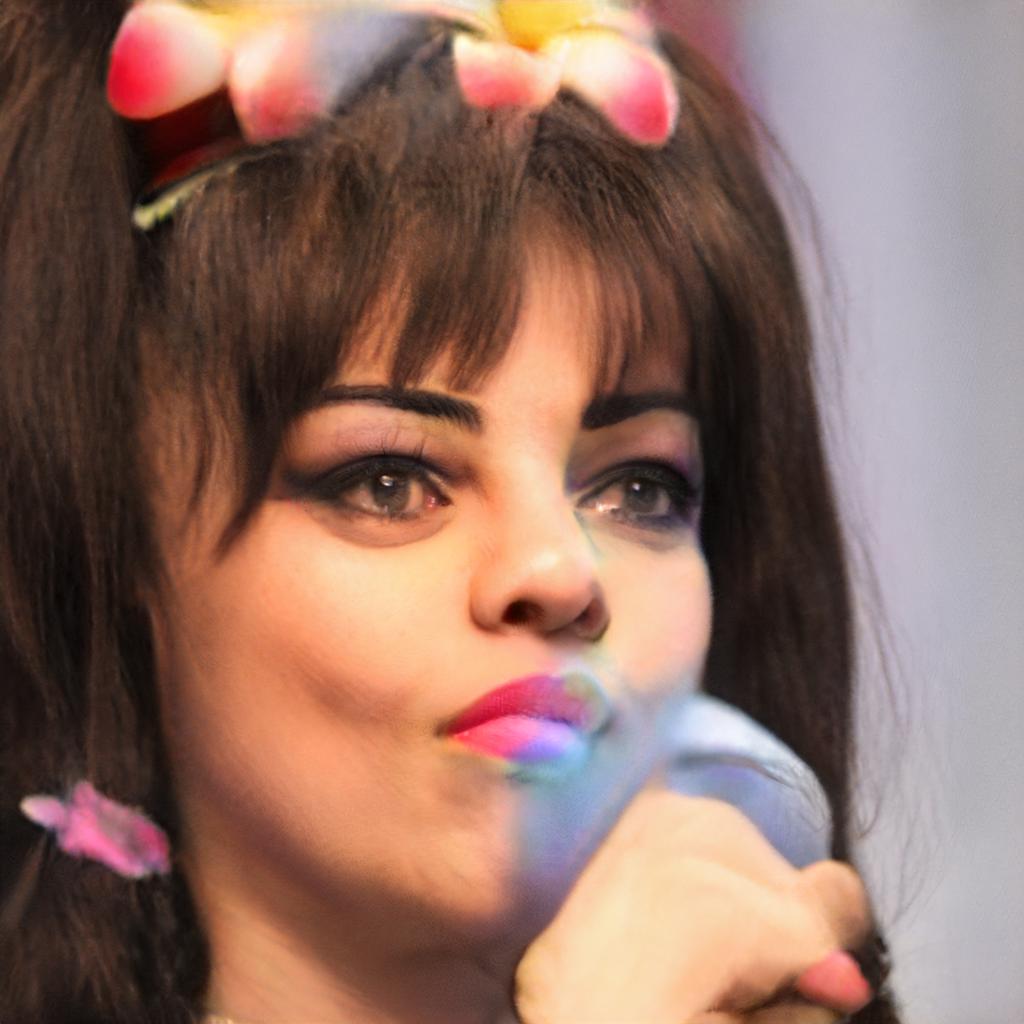} &
        \includegraphics[width=0.13\linewidth]{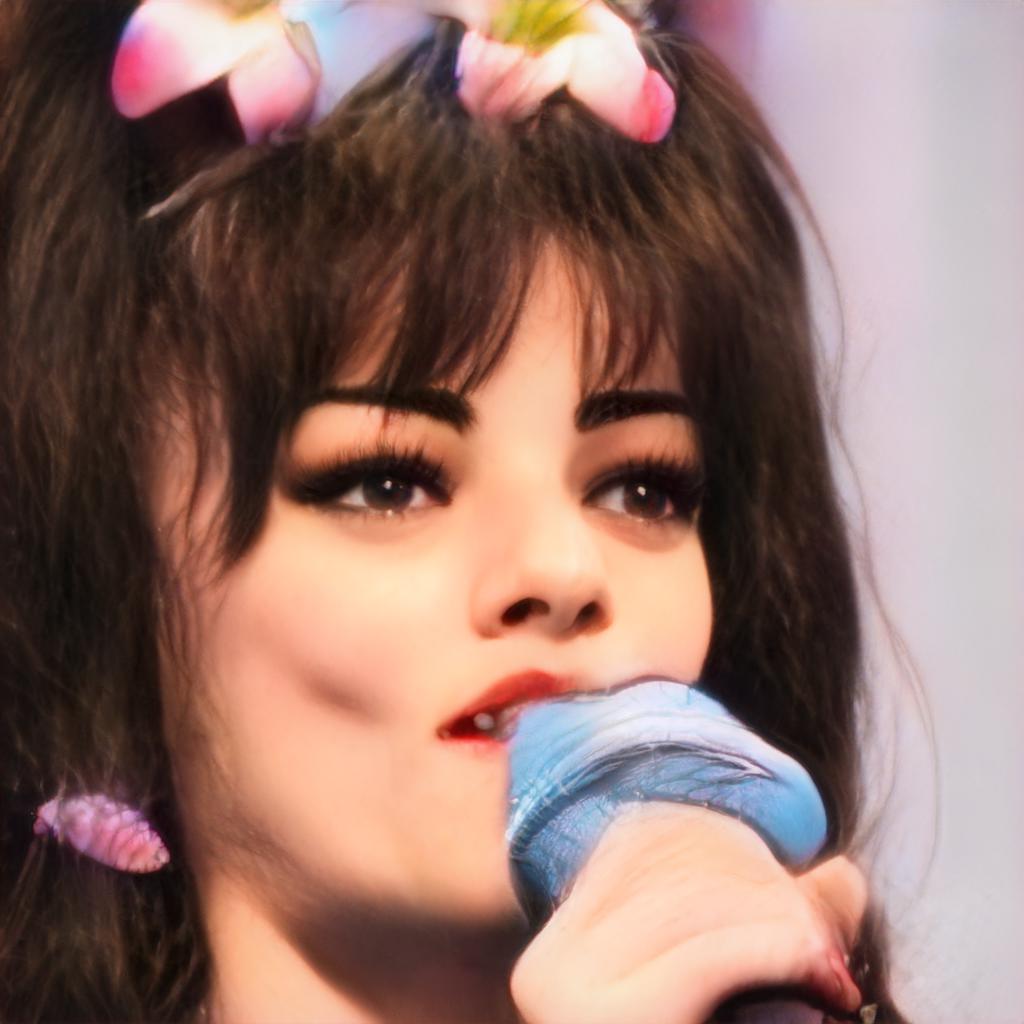} &
        \includegraphics[width=0.13\linewidth]{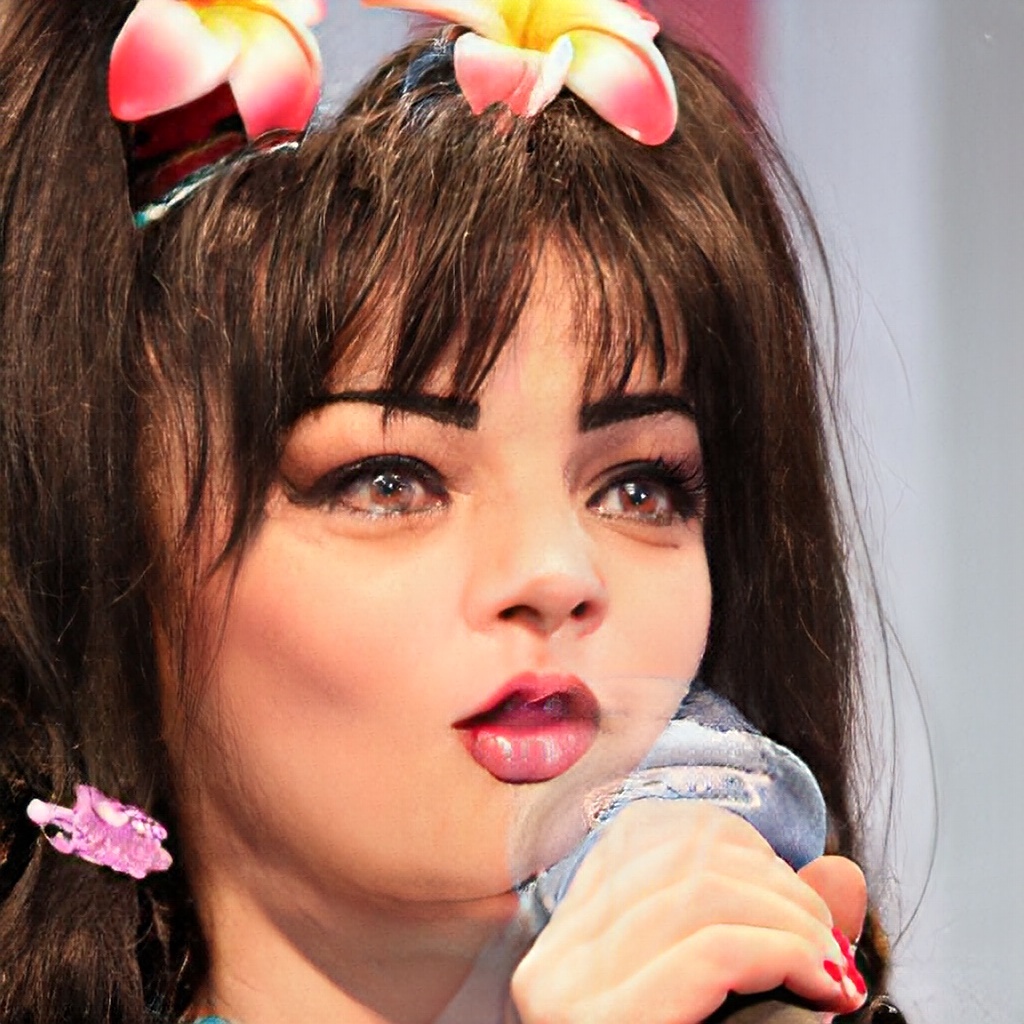} &
        \includegraphics[width=0.13\linewidth]{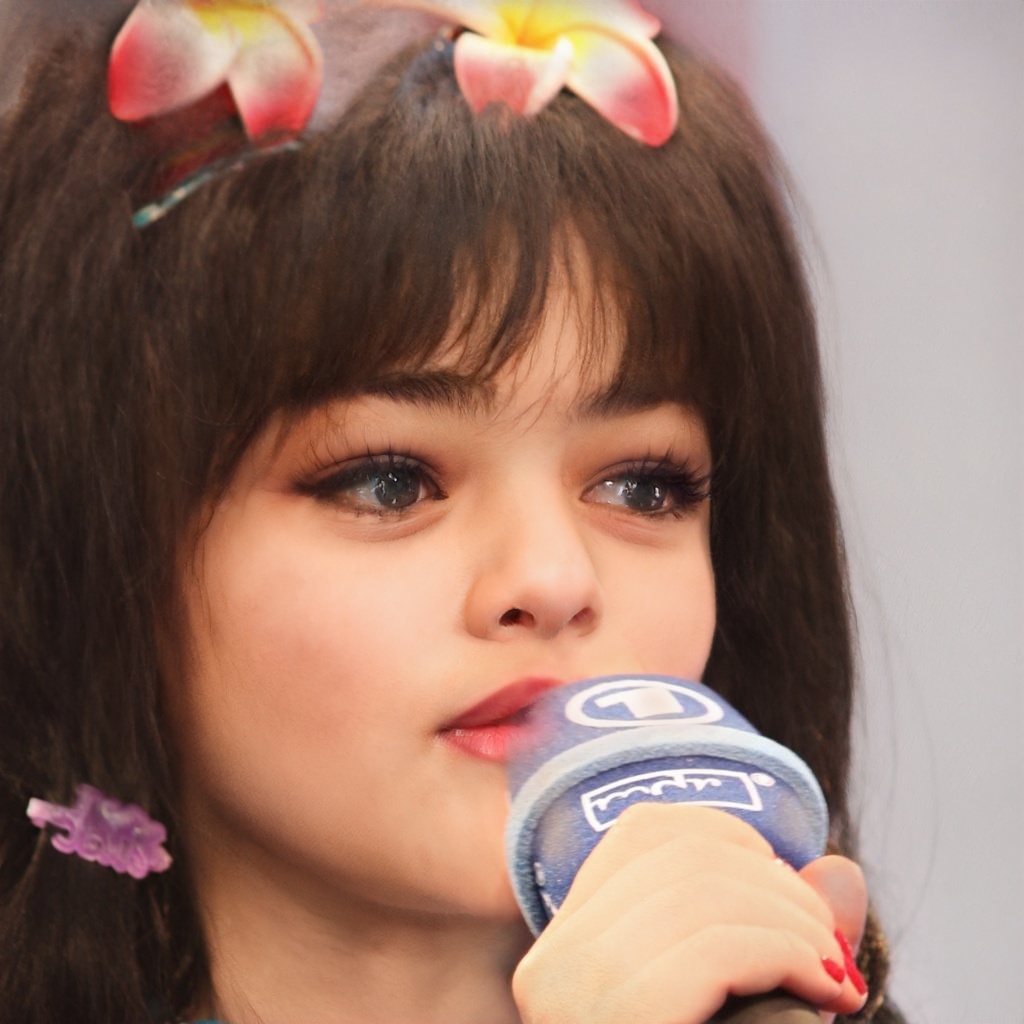}
        \\
        \scriptsize{$\rightarrow$ \textbf{+ Smile}}
        \\
        \includegraphics[width=0.13\linewidth]{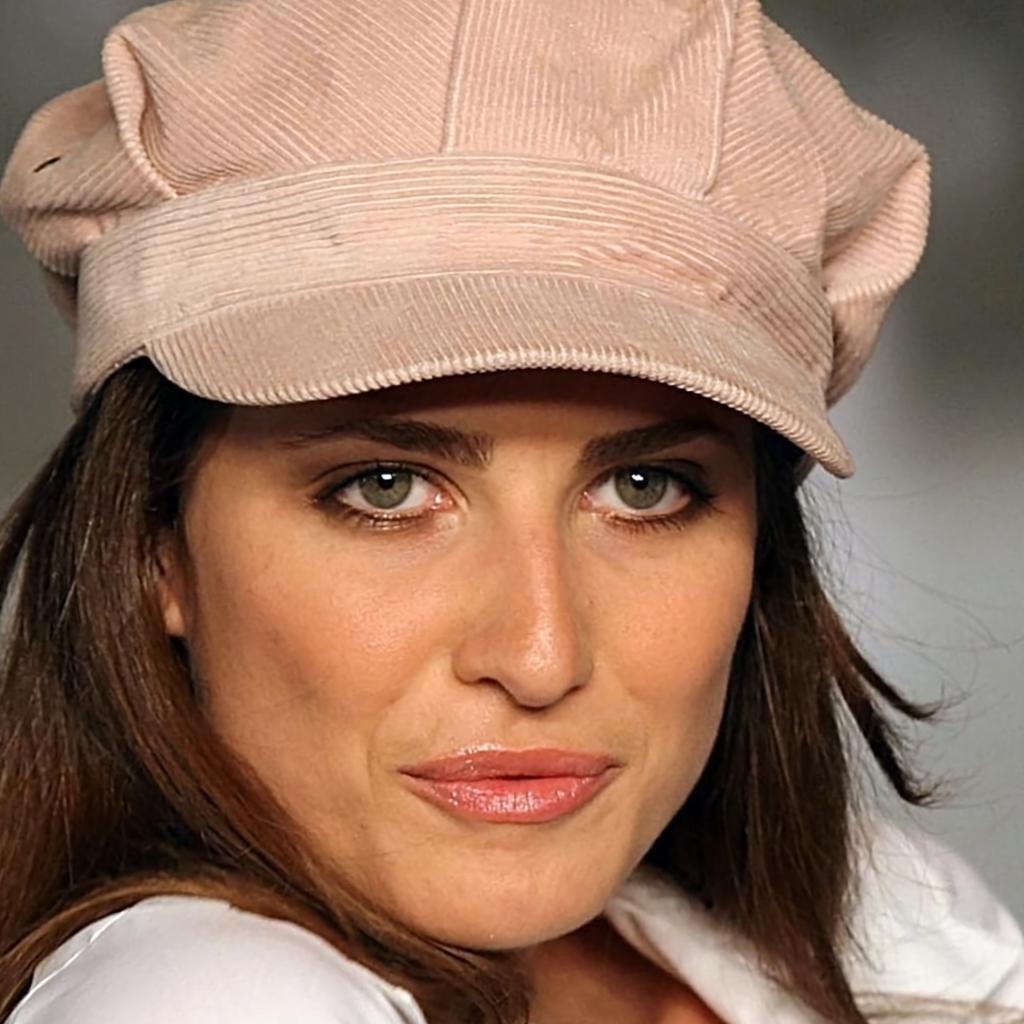} &
        \includegraphics[width=0.13\linewidth]{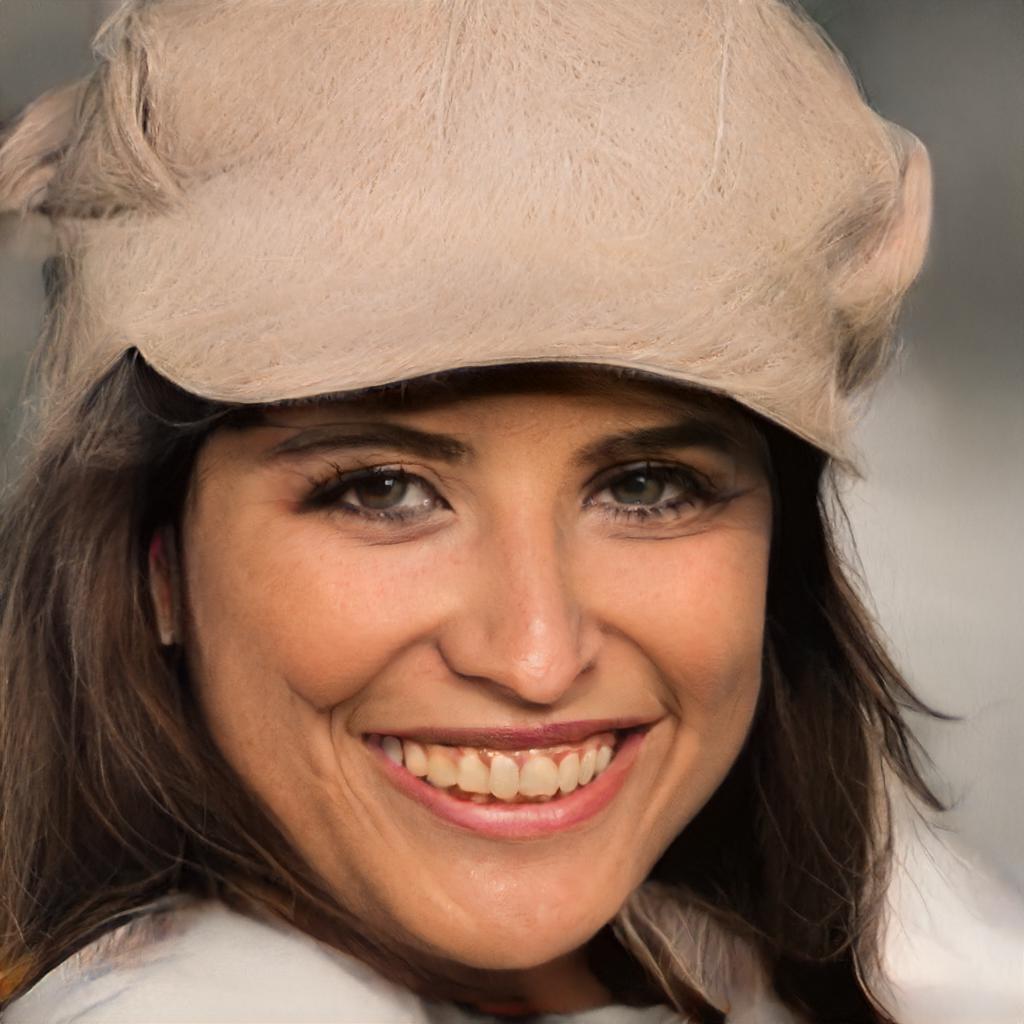} &
        \includegraphics[width=0.13\linewidth]{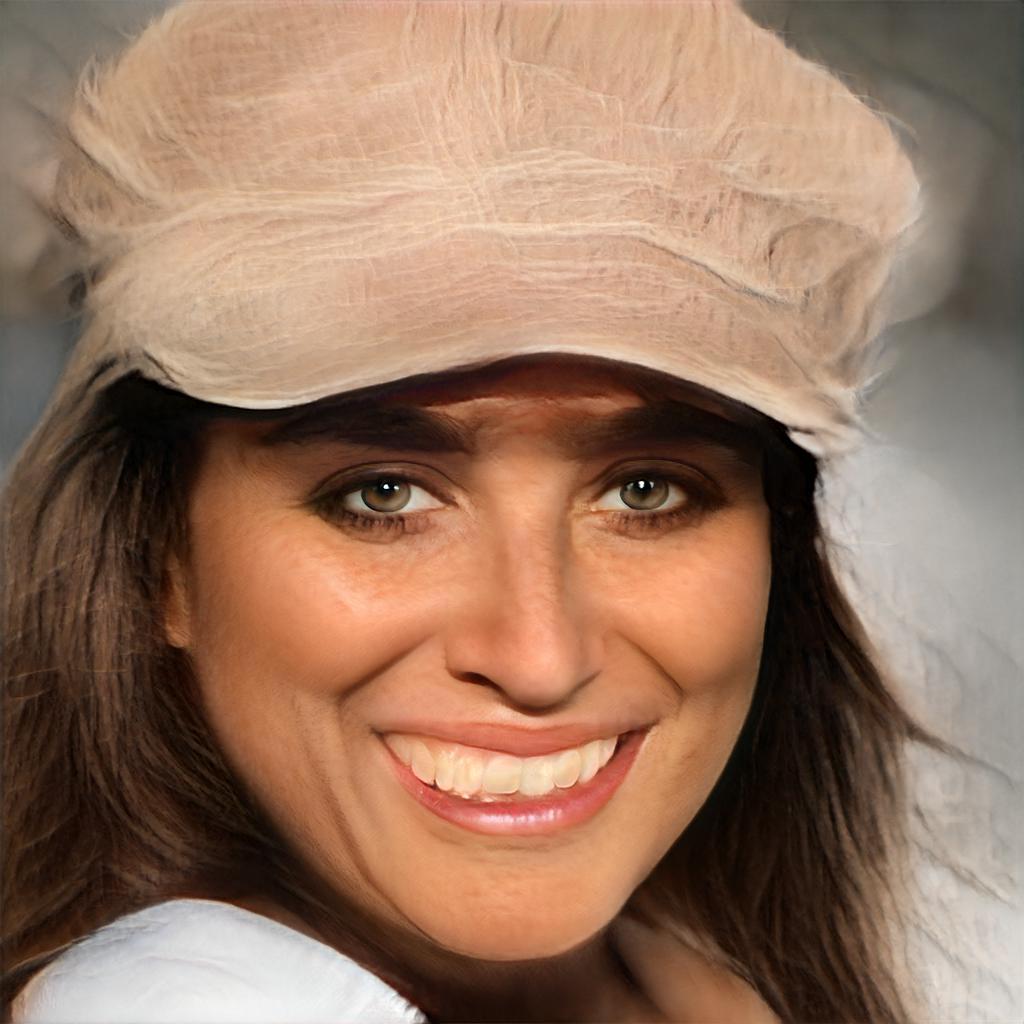} &
        \includegraphics[width=0.13\linewidth]{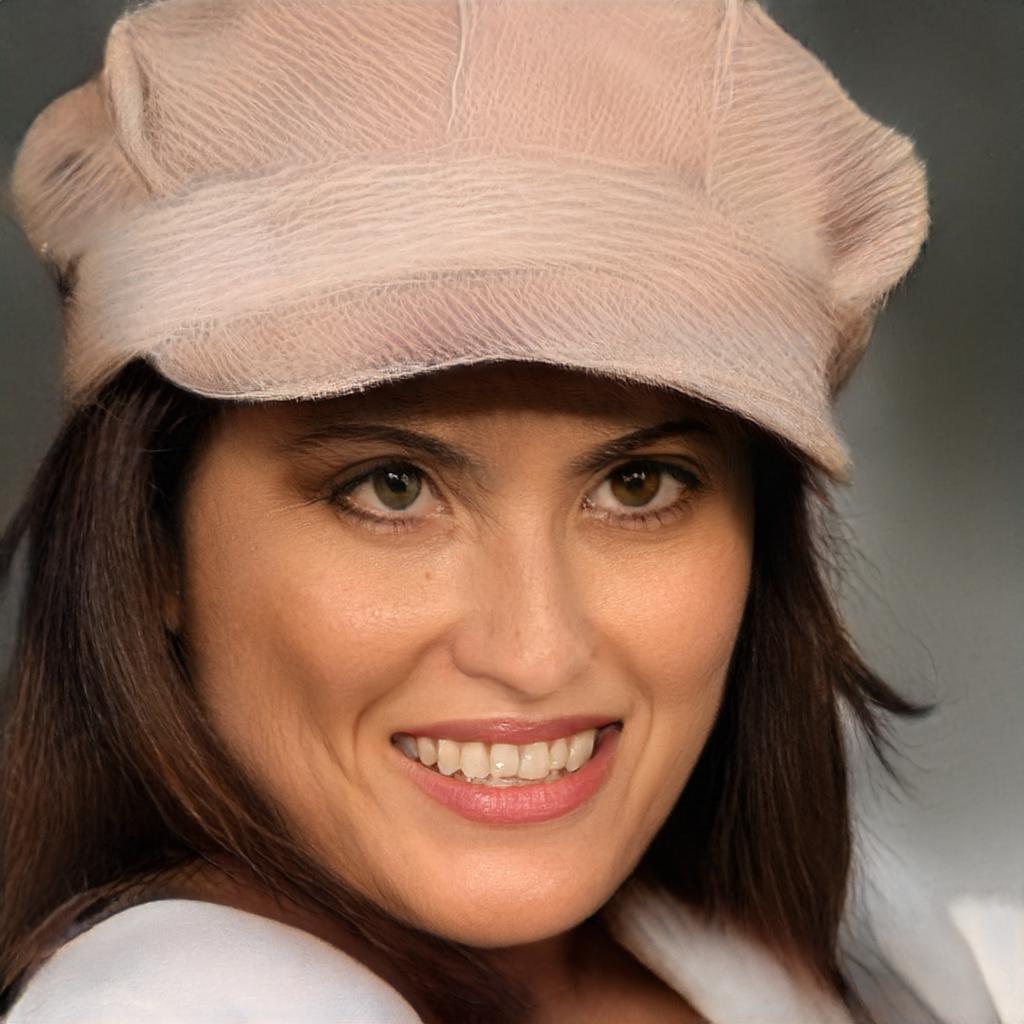} &
        \includegraphics[width=0.13\linewidth]{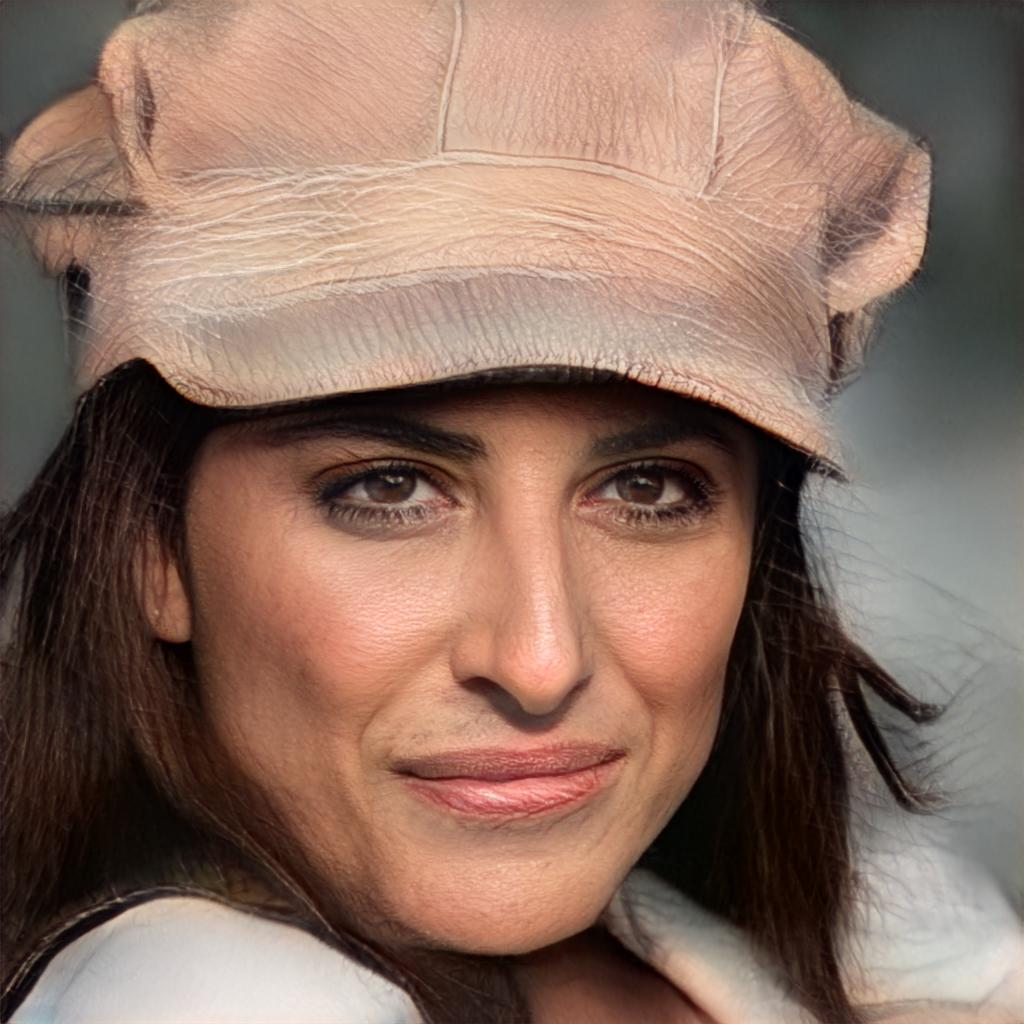} &
        \includegraphics[width=0.13\linewidth]{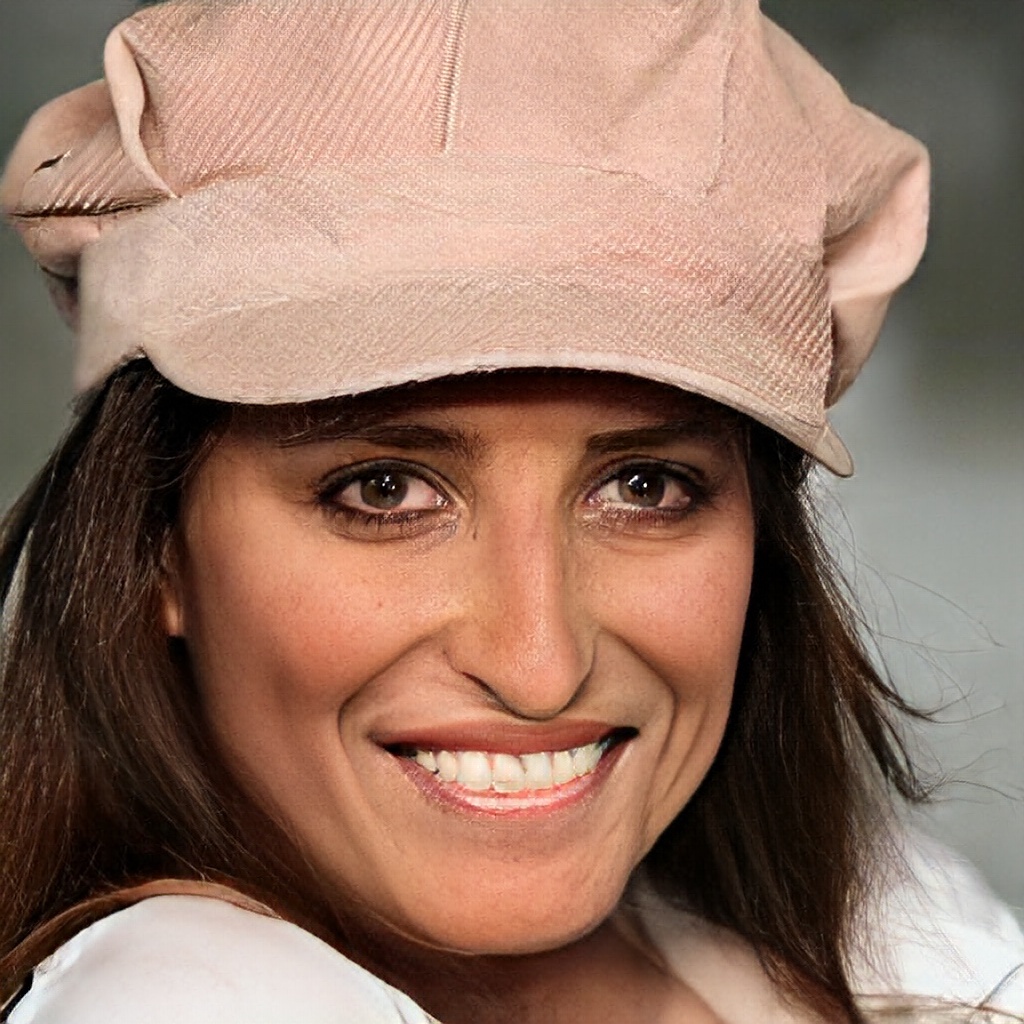} &
        \includegraphics[width=0.13\linewidth]{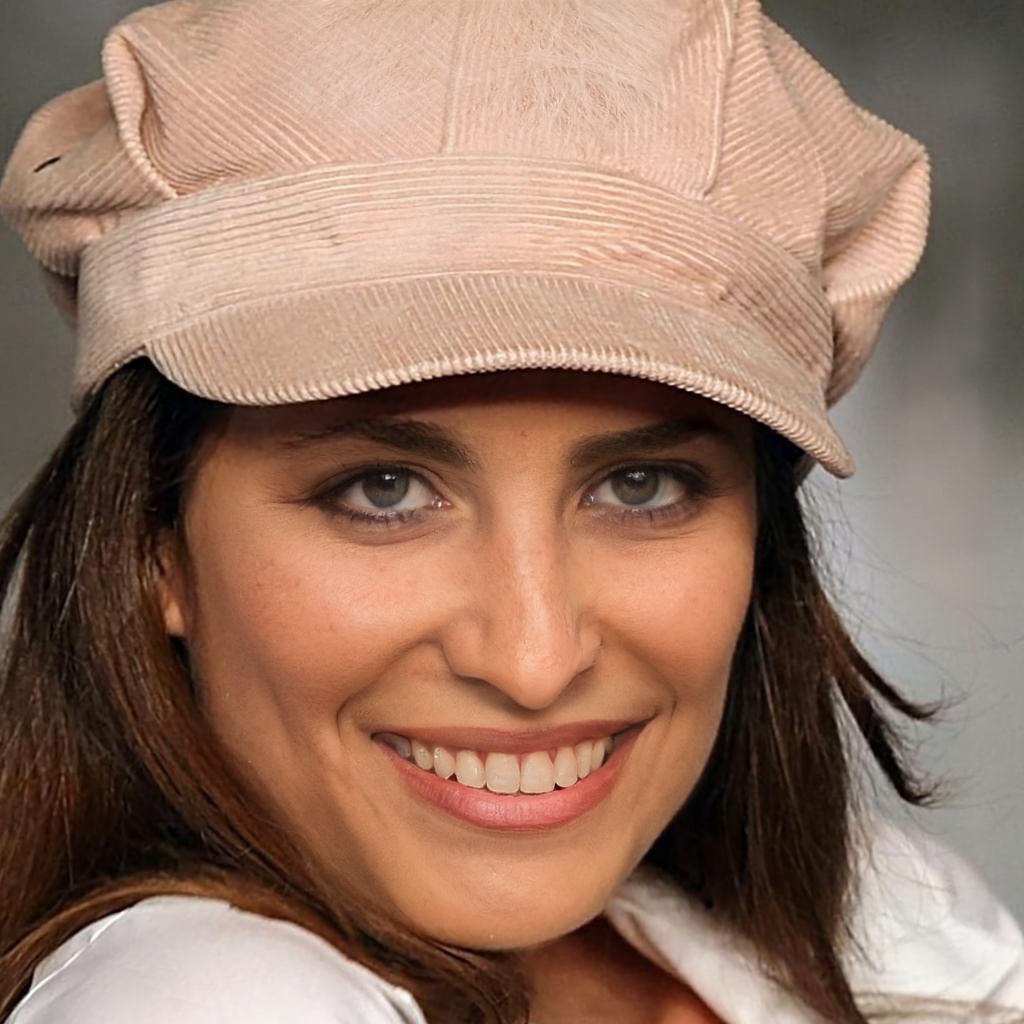}
        \\
        \scriptsize{$\rightarrow$ \textbf{+ Thick Eyebrows}}
        \\
        \includegraphics[width=0.13\linewidth]{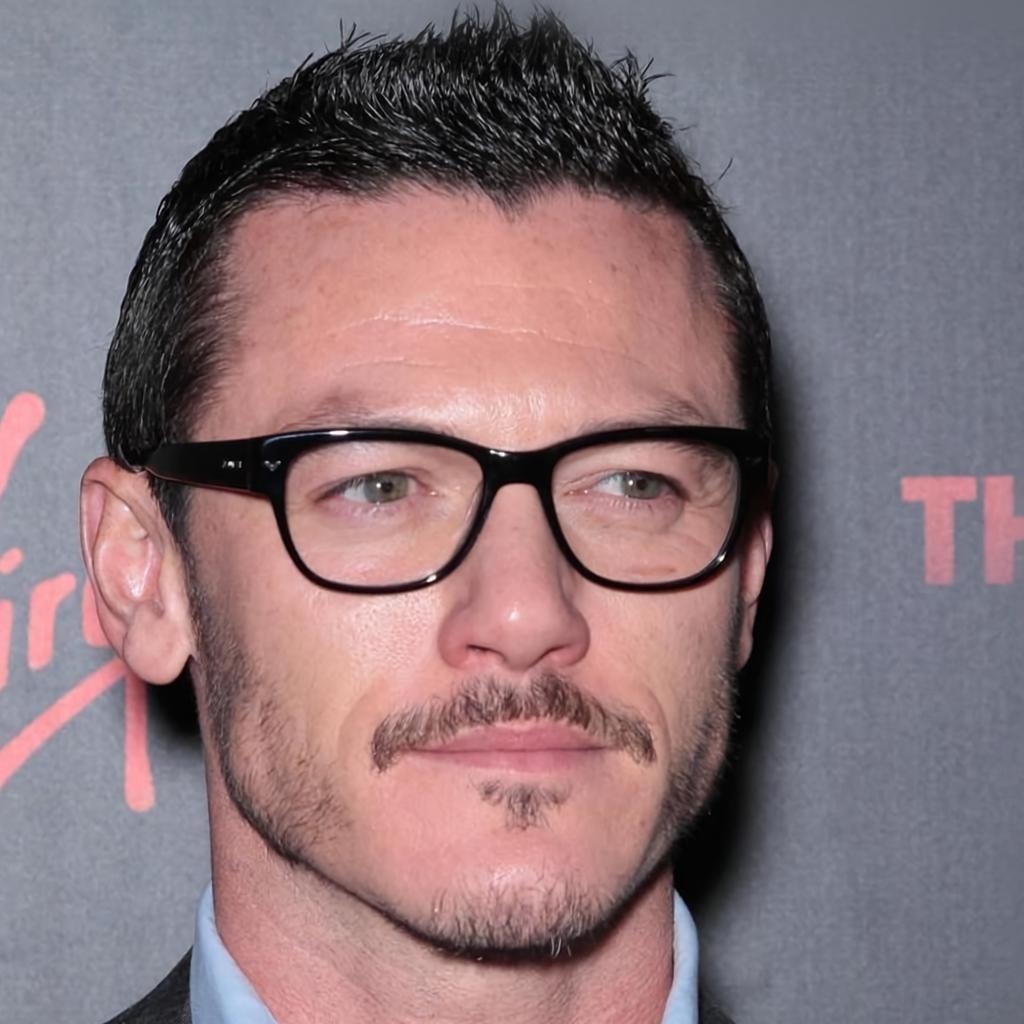} &
        \includegraphics[width=0.13\linewidth]{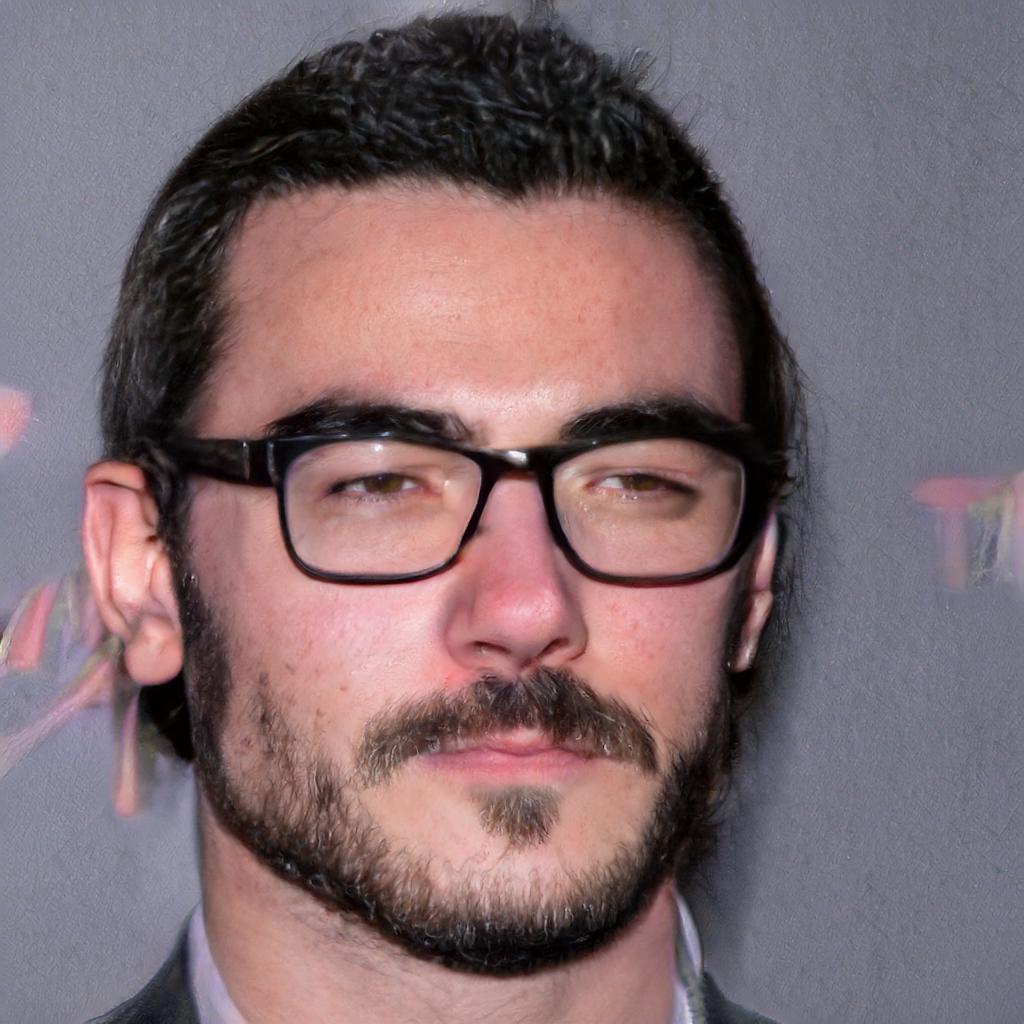} &
        \includegraphics[width=0.13\linewidth]{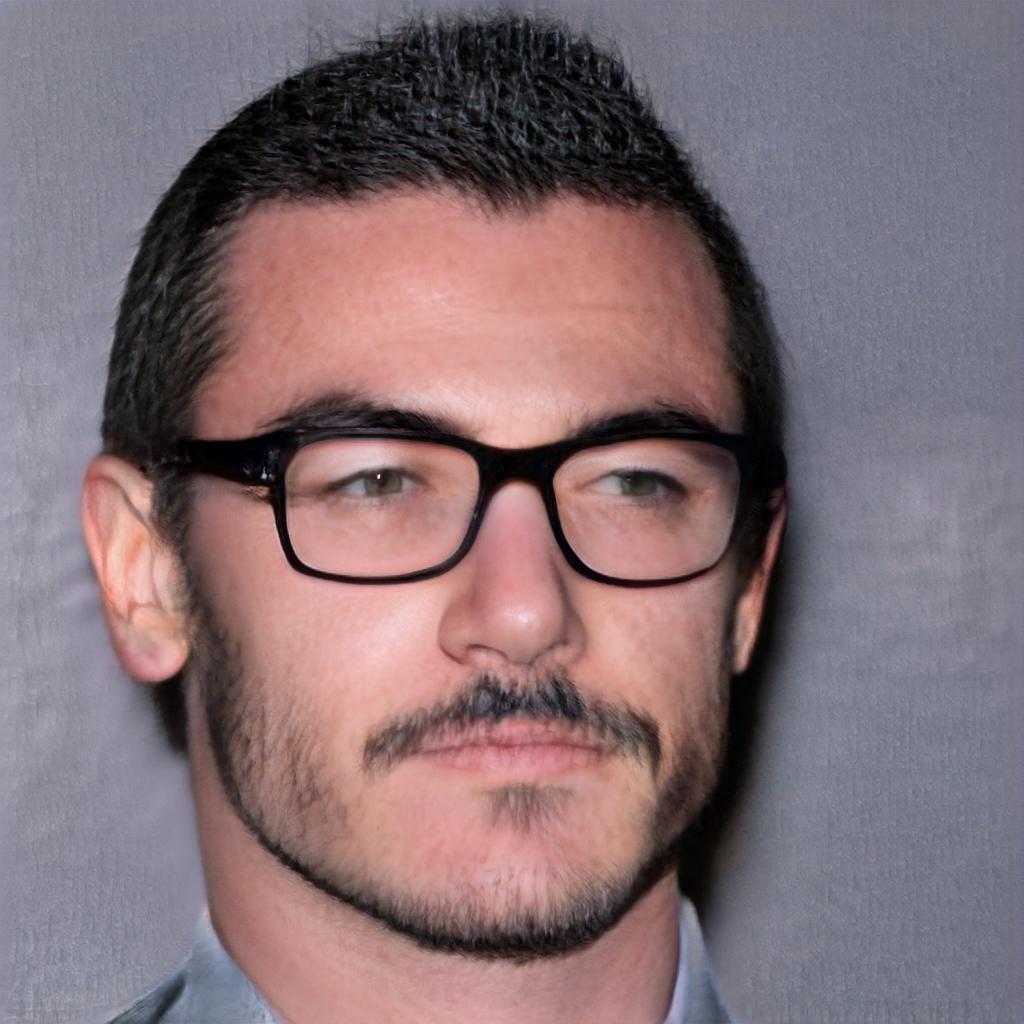} &
        \includegraphics[width=0.13\linewidth]{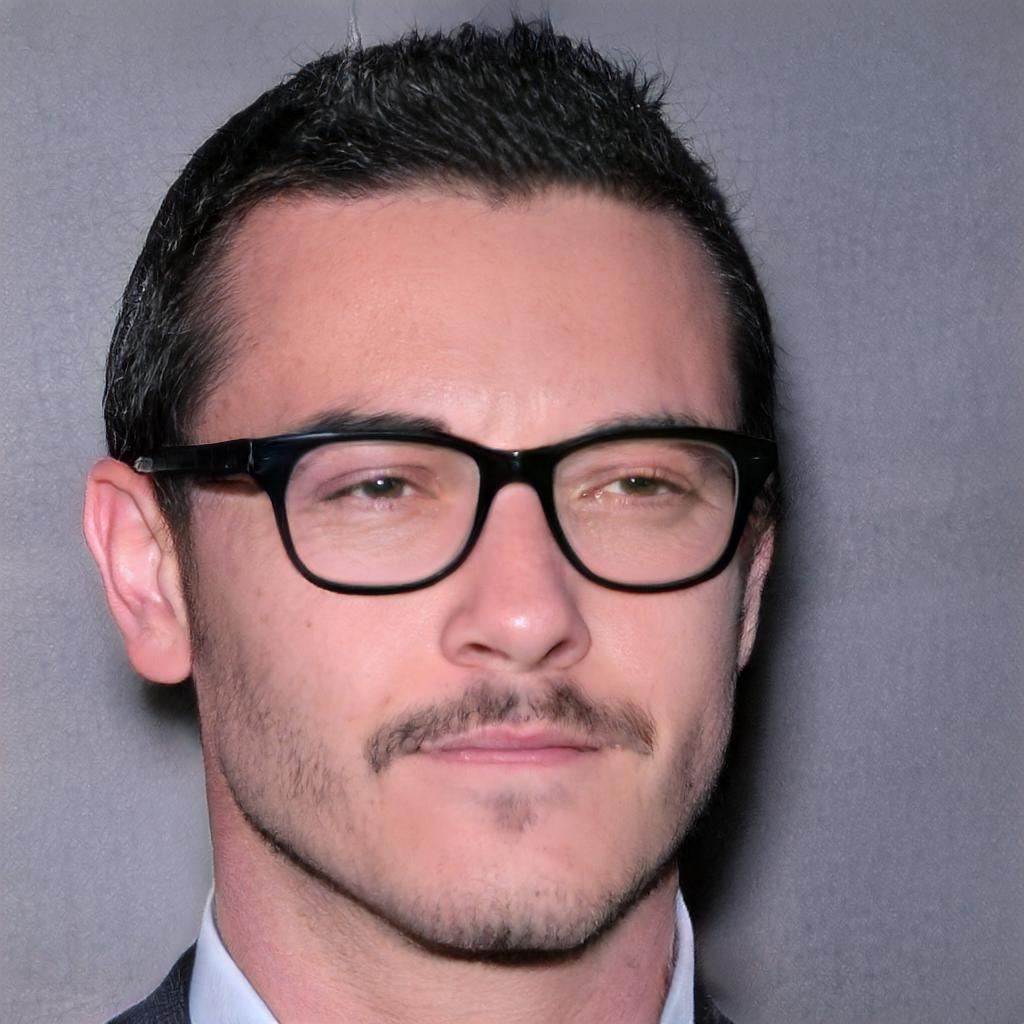} &
        \includegraphics[width=0.13\linewidth]{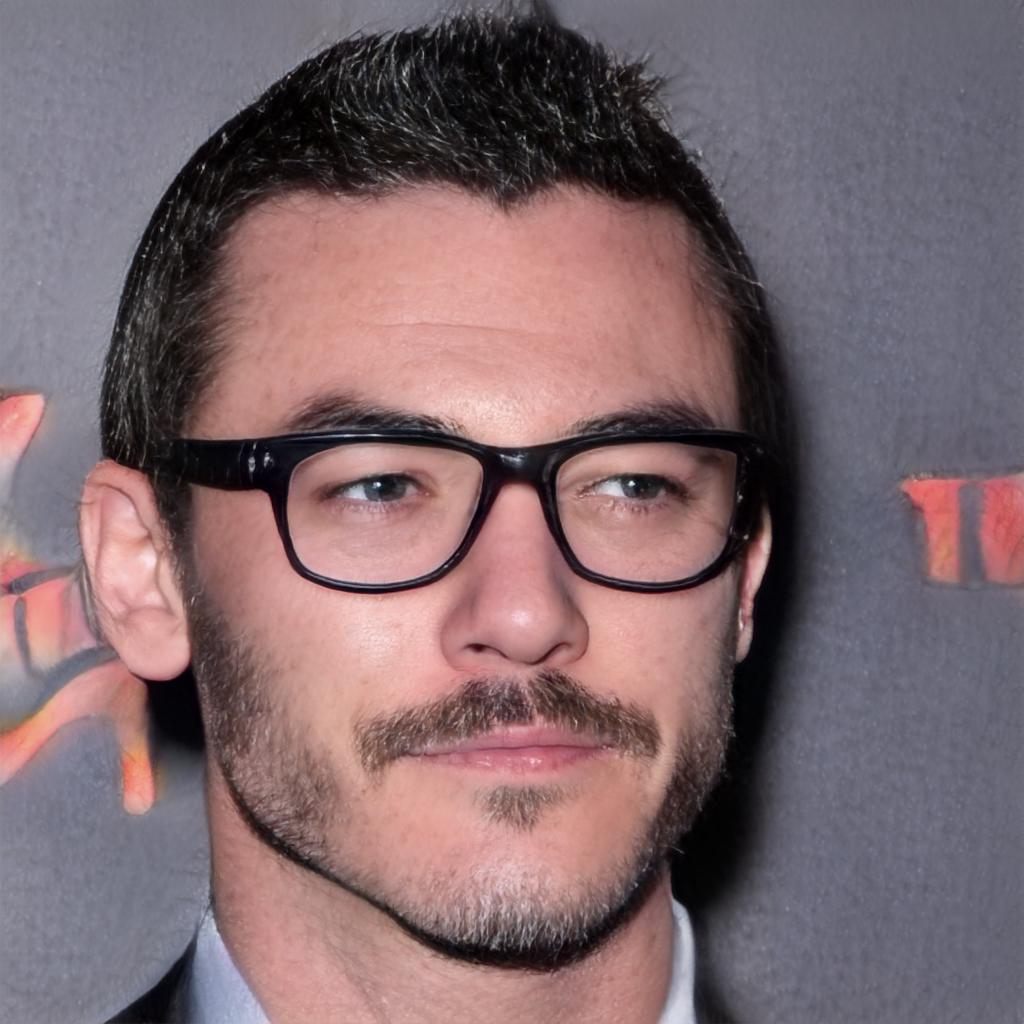} &
        \includegraphics[width=0.13\linewidth]{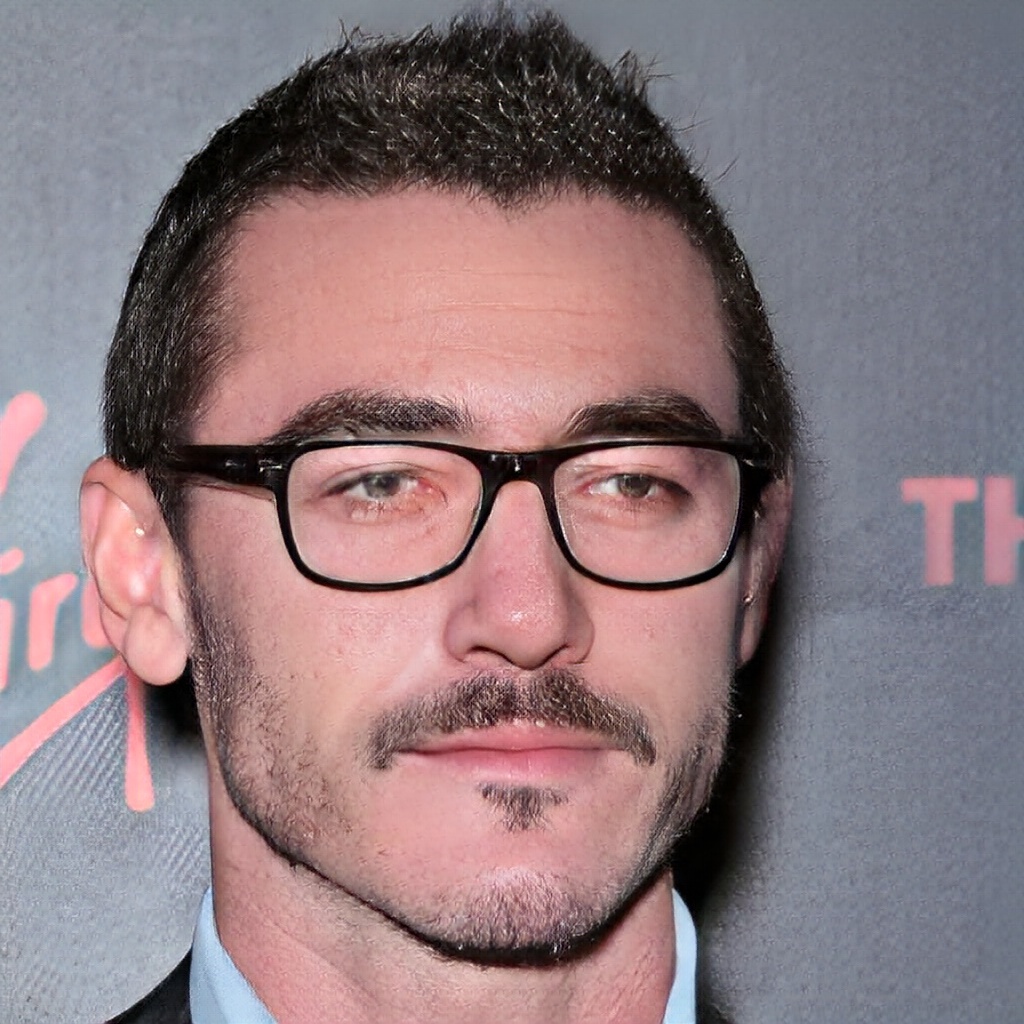} &
        \includegraphics[width=0.13\linewidth]{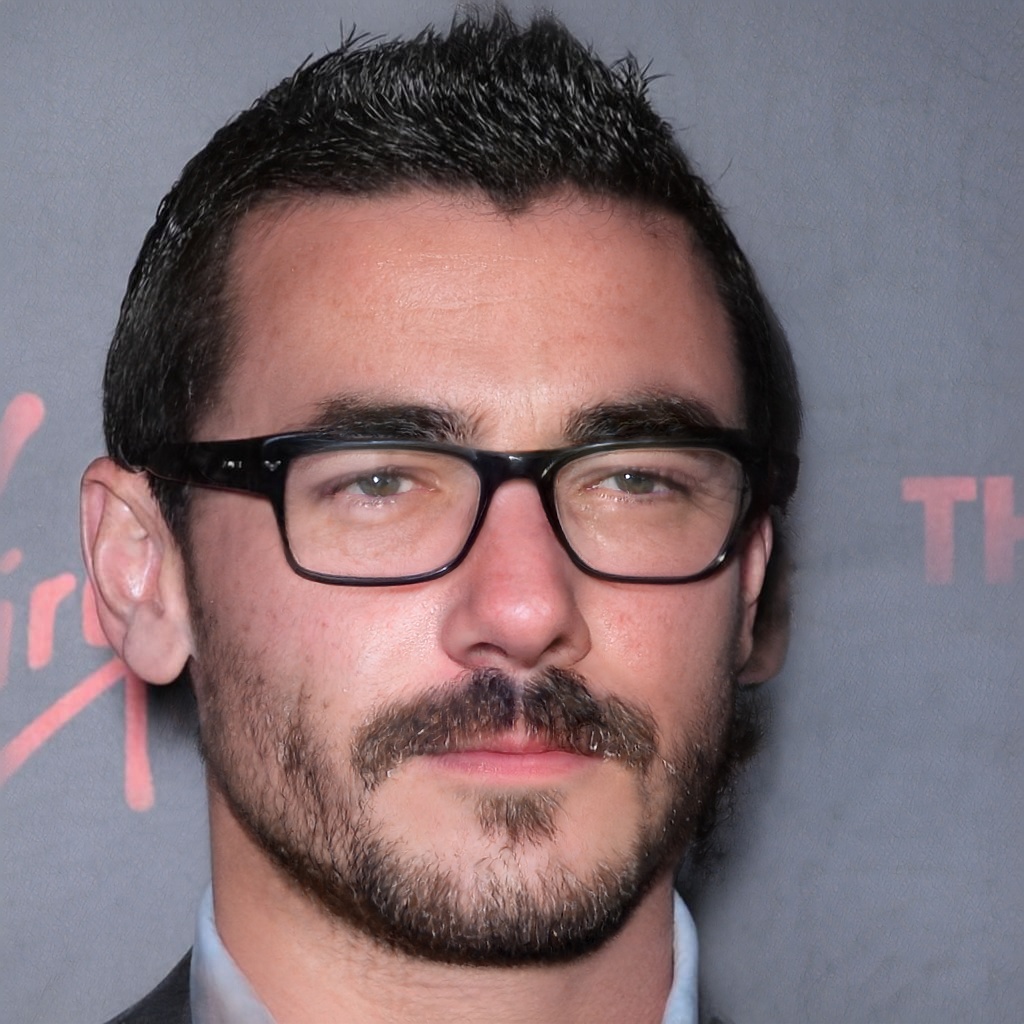}
    \end{tabular}
    \vspace{-0.7em}
    \caption{Comparison of attribute manipulation quality via GAN inversion. We apply the same edit on the latent vectors and generate the editing results with baseline methods and our method. Our method produces the best result with well-preserved out-of-domain content and high-fidelity generated content. Please zoom in for detail.}
    \label{fig:editing_compare}
    \vspace{-1.3em}
\end{figure*}

\begin{figure}[ht]
    \centering
    \begin{tabular}{@{}c@{}c@{}c@{}}
        \scriptsize{Input} & 
        \scriptsize{(a) Ours w/o spatial alignment} & 
        \scriptsize{(b) Ours}
        \\
        \multicolumn{3}{@{}c@{}}{\includegraphics[width=0.93\columnwidth]{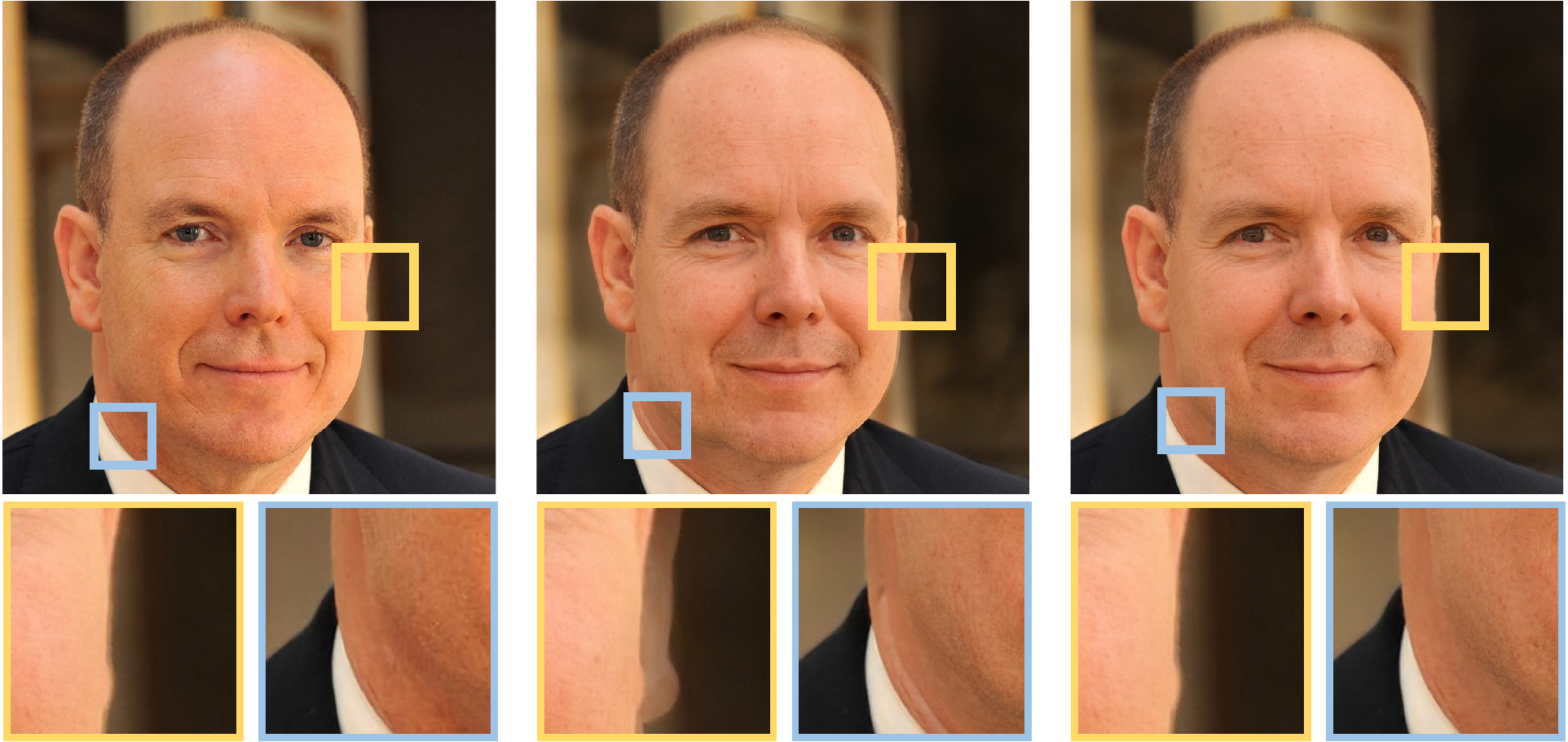}}
        \\
        \resizebox{0.31\columnwidth}{!}{~} &
        \resizebox{0.31\columnwidth}{!}{~} &
        \resizebox{0.31\columnwidth}{!}{~}
    \end{tabular}
    \vspace{-2em}
    \caption{\textbf{Ablation study on spatial alignment.} We generate (a) by skipping the grid sampling operation on $g_i$ in SAMM. Compared to the result of our full model (b), the ghosting artifact is evident in (a). Please zoom in for detail.
    }
    \label{fig:abla_SA}
    \vspace{-1.5em}
\end{figure}

\label{sec:eval}
\begin{table}[t]
\centering
\resizebox{1.0\linewidth}{!}{
    \begin{tabular}{@{}l@{\hspace{1mm}}c@{\hspace{1mm}}c@{\hspace{1mm}}c@{\hspace{1mm}}c@{\hspace{1mm}}c@{}}
        \hline
        Method & PSNR$\uparrow$ & SSIM$\uparrow$ & LPIPS$\downarrow$ & FID$\downarrow$ & Runtime(s)$\downarrow$
        \\
        \hline
        e4e~\cite{tov2021designing} & 20.30 & 0.665 & 0.350 & 34.98 & 0.145\\
        ReStyle~\cite{alaluf2021restyle} & 21.46 & 0.682 & 0.345 & 32.20 & 0.448\\
        HFGI$_{e4e}$~\cite{wang2021high} & 23.66 & 0.724 & 0.285 & 22.79 & 0.159 \\
        HyperStyle~\cite{alaluf2021HyperStyle}, iter=5 & 23.59 & 0.728 & 0.300 & 30.49 & 0.915 \\
        SAM~\cite{parmar2022spatially}, iter=10 & 21.95 & 0.684 & 0.311 & 27.38 & 1.199\\
        FeatureStyle~\cite{xuyao2022} & 25.24 & 0.717 & 0.188 & 16.86 & 0.181\\
        \hline
        Ours$_{e4e}$, N=1 & 26.97 & 0.892 & 0.147 & 13.33 & 0.166\\
        Ours$_{e4e}$, N=2 & {27.08} & {0.897} & {0.143} & {13.13} & 0.193\\ 
        Ours$_{e4e}$, N=3 & \textbf{27.24} & \textbf{0.900} & \textbf{0.139} & \textbf{12.73} & 0.218 \\
        Ours$_{ReStyle}$, N=2 & {25.37} & {0.813} & {0.236} & {17.68} & 0.602\\
        \hline
    \end{tabular}
}
\vspace{-0.5em}
\caption{Quantitative evaluation of GAN inversion quality on the first 1,000 images in the CelebAHQ-Mask~\cite{CelebAMask-HQ} testing dataset. The runtime is measured on a single nvidia RTX3090 GPU.}
\label{tab:quantitative_evaluation}
\vspace{-1em}
\end{table}


\begin{table}[t]
\centering
\resizebox{1.0\linewidth}{!}{
    \begin{tabular}{@{}l@{\hspace{1mm}}c@{\hspace{1mm}}c@{\hspace{1mm}}|l@{\hspace{1mm}}c@{\hspace{1mm}}c}
        \hline
        {Method} & {Inversion $\downarrow$} & {Editing $\downarrow$} &
        {Method} & {Inversion $\downarrow$} & {Editing $\downarrow$}
        \\
        \hline
        HyperStyle~\cite{alaluf2021HyperStyle}, iter=5 & 3.67 & 3.51 & 
        SAM~\cite{parmar2022spatially}, iter=10  & 3.79 & 3.56
        \\
        HFGI$_{e4e}$~\cite{wang2021high} & 3.19 & 4.13 & 
        DiffCAM~\cite{song2022editing}  & - & 3.75
        \\
        FeatureStyle~\cite{xuyao2022} & 2.68 & 4.01 & 
        Ours$_{e4e}$ & \textbf{1.64} & \textbf{2.04}
        \\
        \hline
    \end{tabular}
}
\vspace{-0.5em}
\caption{Average ranking of user's preference on the GAN inversion and attribute editing results of human face images. }
\label{tab:user_study}
\vspace{-2em}
\end{table}

\noindent \textbf{Face inversion.}
Following previous works~\cite{wang2021high, song2022editing, parmar2022spatially, xuyao2022}, we evaluate our model and state-of-the-art baseline methods on the first 1,000 images in the testing partition of CelebAHQ-Mask~\cite{CelebAMask-HQ} dataset, assessing the inversion quality. 
We measure the image reconstruction accuracy with PSNR and SSIM~\cite{wang2004image}, the perceptual distance with LPIPS~\cite{zhang2018perceptual}, and also the distribution distance between the reconstructed image and the source image that is represented with Fréchet inception distance~\cite{heusel2017gans} (FID).

For HyperStyle~\cite{alaluf2021HyperStyle}, we set the optimization iteration to 5. For SAM~\cite{parmar2022spatially}, it takes over 131 seconds to finish 1001 iterations of optimization to produce a result, which is too time-consuming for testing. To make a fair comparison, we set the optimization iteration of SAM to 10 in our experiment. Also, because DiffCAM~\cite{song2022editing} only works for attribute editing, we do not evaluate this model for face inversion.
As is shown in Tab.~\ref{tab:quantitative_evaluation}, our model outperforms previous methods, achieving the best restoration quality measure with different metrics.
In Fig.~\ref{fig:inversion}, we present multiple inversion results for comparison. Our method preserves meticulous details in the background, hats, and even the cigarette on the face by conducting the blending for the out-of-domain objects during the generation. 
We conduct a user study and ask the users to rank the face inversion and editing results based on the image faithfulness, the detail preservation of OOD contents, the overall visual satisfaction, and the editing quality. The average rankings are summarized in Tab.~\ref{tab:user_study}. 
Most participants rated both the results of face inversion and attribute manipulation of our framework as the best. 
Please refer to our supplement for more details of the user study and more visual results for comparison.

\noindent \textbf{Attribute manipulation.}
Also, we have compared our method with state-of-the-art methods~\cite{wang2021high, alaluf2021restyle, alaluf2021HyperStyle, parmar2022spatially} on face attribute manipulation~\cite{harkonen2020ganspace,patashnik2021styleclip}. 
In Fig.~\ref{fig:editing}, we apply off-the-shelf GAN editing approach~\cite{shen2020interfacegan, shen2020interfacegan} with our framework on CelebAHQ-Mask~\cite{CelebAMask-HQ} images, our method preserves the out-of-domain contents regardless of the editing direction. Moreover, as shown in Fig.~\ref{fig:clip_edit}, we also perform text-guided semantic editing on the hair color with the CLIP~\cite{radford2021learning,patashnik2021styleclip} model.
In Fig.~\ref{fig:teaser}\footnote{Image license: www.pexels.com/license} and Fig.~\ref{fig:editing_compare}, we compare our attribute editing performance with baseline approaches. Our framework produces high-fidelity editing results without undesired artifacts and provides the best editing quality against existing works. 
Compared with ~\cite{alaluf2021HyperStyle, parmar2022spatially, wang2021high, xuyao2022}, our work preserves more details in the microphone, hat, and background.

In addition, we notice an unnatural over-sharpened artifact in the result of DiffCAM~\cite{song2022editing} (Fig.~\ref{fig:teaser}(e)). We hypothesize their ghosting removal module introduces such an artifact. Instead, our SAMM module helps decrease the ID reconstruction error along with the generation of $\widehat{x_{in}}$. Thus we do not need a post-processing process after the blending. Furthermore, ours works better than ~\cite{song2022editing, parmar2022spatially} in cases with occlusions on faces. For more visual results, please refer to our supplement.

\subsection{Ablation Study}
\label{sec:ablation}
In this section, we conduct ablation studies on our spatial alignment and masking module.

\begin{table}[th]
\centering
\resizebox{1.0\linewidth}{!}{
    \begin{tabular}{@{}l@{\hspace{1mm}}c@{\hspace{1mm}}c@{\hspace{1mm}}c@{\hspace{1mm}}c@{\hspace{1mm}}c@{}}
        \hline
        Method & PSNR$\uparrow$ & SSIM$\uparrow$ & LPIPS$\downarrow$ & FID$\downarrow$ & Avg. OOD area(\%)$\downarrow$
        \\
        \hline
        Ours$_{e4e}$, w/o SA & {26.32} & {0.875} & {0.171} & 13.53 & 31.34\\
        Ours$_{e4e}$, N=2, skip SA & {26.68} & {0.888} & {0.156} & 14.26 & 23.97\\
        \hline
    \end{tabular}
}
\vspace{-0.5em}
\caption{Ablation of spatial and iterative alignment. The average OOD area is calculated by averaging the intensity of the predicted invertibility masks on the testing dataset.}
\label{tab:quantitative_evaluation_SA}
\vspace{-1em}
\end{table}

\noindent \textbf{Spatial alignment (SA).}
We study SA by training the model without SA (``Ours$_{e4e}$, w/o SA'') and by skipping the grid sampling operation in Algorithm~\ref{alg:samm} on our full model (``Ours$_{e4e}$, N=2, skip SA"). 
As shown in Tab.~\ref{tab:quantitative_evaluation_SA}, we found that the reconstruction quality drops compared to our full model for both settings. Additionally, the predicted average OOD area (AOA) is 7.37\% larger for ``Ours$_{e4e}$, w/o SA", while a lower AOA is vital to our framework since we hope to maximize the ID area in the decomposition to better facilitate downstream applications such as attribute editing.
In Fig.~\ref{fig:abla_SA} we provide some visual results of skipping the SA, where we detect undesired ghosting artifacts in the blending result. Please refer to our supplement for more analysis.

\noindent \textbf{Iterative alignment.}
In Sec.~\ref{sec:SAMM}, we introduce the iterative alignment strategy for flow and mask prediction. 
In Tab.~\ref{tab:quantitative_evaluation}, we quantitatively evaluate the inversion quality of our model with $N$=$\{1, 2, 3\}$, and the reconstruction performance increase with the counts of iterative alignment. Besides, we measure the AOA for different $N$. With $N$=2, $23.97\%$ of pixels in our testing dataset is recognized as the OOD partition and the number increases by $2.4\%$ and $0.65\%$ when $N$=1 and $N$=3. It suggests that the iterative alignment strategy also improves the editability of our framework. 
Also, we conduct experiments to investigate the usefulness of our loss function on the masks. Please refer to our supplement for more details.

\begin{figure}[t]
    \centering
    \begin{tabular}{@{}c@{\hspace{1mm}}c@{\hspace{1mm}}c@{\hspace{1mm}}c@{}}
        \scriptsize{Input} & 
        \scriptsize{e4e~\cite{tov2021designing} Inversion} & 
        \scriptsize{{Ours$_{e4e}$ Inversion}} & 
        \scriptsize{$\rightarrow${+Grass}}
        \\
        \includegraphics[width=0.24\linewidth]{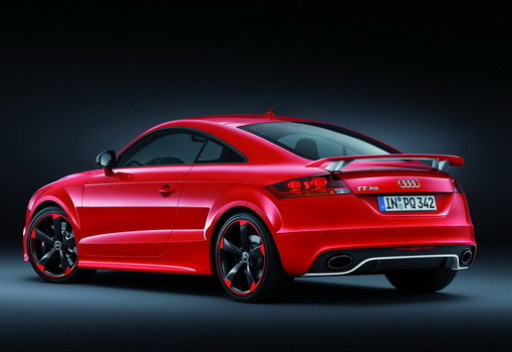} &
        \includegraphics[width=0.24\linewidth]{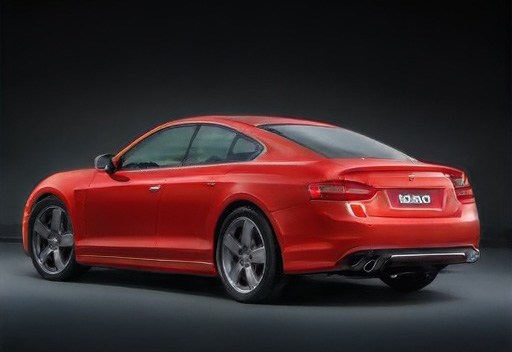} &
        \includegraphics[width=0.24\linewidth]{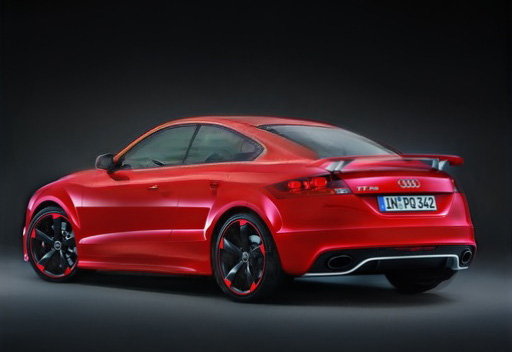} & 
        \includegraphics[width=0.24\linewidth]{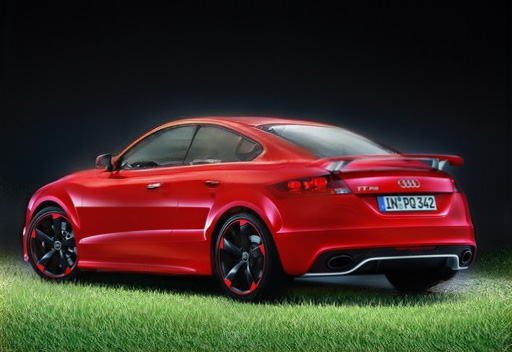}
        \\
        \vspace{-1.5em}
    \end{tabular}
    \caption{Our GAN inversion and manipulation on Standford Cars~\cite{KrauseStarkDengFei-Fei_3DRR2013} dataset.}
    \label{fig:cars}
    \vspace{-1.5em}
\end{figure}

\subsection{Extension to non-face domains}
Conceptually, our framework could be extended to other image domains such as cars for better OOD GAN inversion quality. As shown in Fig.~\ref{fig:cars}, our framework also improves the car inversion result, compared to e4e~\cite{tov2021designing}. We will leave more investigation of non-face domains to our future work.

\section{Conclusion}
\label{sec:conclusion}
In this paper, we propose a novel framework for photo-realistic OOD GAN inversion via invertibility decomposition on human face images.
We design the SAMM to predict the invertibility mask and the optical flows along with the generation process, which can work as a general plugin with different pre-trained encoders (i.e., e4e~\cite{tov2021designing}, ReStyle~\cite{alaluf2021restyle}). 
We only invert the ID area in the input image using the encoder and generator, then align the generated features with the input features for better reconstruction accuracy without compromising the editability.
Then, we can manipulate the attributes in the ID area with off-the-shelf approaches. 
Lastly, we seamlessly merge the OOD contents with the generated image. Our framework produces photo-realistic results and outperforms previous works in terms of both inversion accuracy and editing fidelity, as demonstrated in our experiments.

{\small
\bibliographystyle{ieee_fullname}
\bibliography{egbib}

\begin{thebibliography}{10}\itemsep=-1pt

\bibitem{abdal2019image2stylegan}
Rameen Abdal, Yipeng Qin, and Peter Wonka.
\newblock Image2stylegan: How to embed images into the stylegan latent space?
\newblock In {\em ICCV}, 2019.

\bibitem{abdal2020image2stylegan++}
Rameen Abdal, Yipeng Qin, and Peter Wonka.
\newblock Image2stylegan++: How to edit the embedded images?
\newblock In {\em CVPR}, 2020.

\bibitem{abdal2022clip2stylegan}
Rameen Abdal, Peihao Zhu, John Femiani, Niloy Mitra, and Peter Wonka.
\newblock Clip2stylegan: Unsupervised extraction of stylegan edit directions.
\newblock {\em ACM TOG}, 2022.

\bibitem{alaluf2021restyle}
Yuval Alaluf, Or Patashnik, and Daniel Cohen-Or.
\newblock Restyle: A residual-based stylegan encoder via iterative refinement.
\newblock In {\em ICCV}, 2021.

\bibitem{alaluf2021HyperStyle}
Yuval Alaluf, Omer Tov, Ron Mokady, Rinon Gal, and Amit~H. Bermano.
\newblock Hyperstyle: Stylegan inversion with hypernetworks for real image
  editing.
\newblock In {\em CVPR}, 2021.

\bibitem{bae2022furrygan}
Jeongmin Bae, Mingi Kwon, and Youngjung Uh.
\newblock Furrygan: High quality foreground-aware image synthesis.
\newblock In {\em ECCV}, 2022.

\bibitem{bai2022high}
Qingyan Bai, Yinghao Xu, Jiapeng Zhu, Weihao Xia, Yujiu Yang, and Yujun Shen.
\newblock High-fidelity gan inversion with padding space.
\newblock In {\em ECCV}, 2022.

\bibitem{chong2021stylegan}
Min~Jin Chong, Hsin-Ying Lee, and David Forsyth.
\newblock Stylegan of all trades: Image manipulation with only pretrained
  stylegan.
\newblock {\em arXiv}, 2021.

\bibitem{Collins20}
Edo Collins, Raja Bala, Bob Price, and Sabine S{\"u}sstrunk.
\newblock Editing in style: Uncovering the local semantics of {GANs}.
\newblock In {\em CVPR}, 2020.

\bibitem{deng2018arcface}
Jiankang Deng, Jia Guo, Xue Niannan, and Stefanos Zafeiriou.
\newblock Arcface: Additive angular margin loss for deep face recognition.
\newblock In {\em CVPR}, 2019.

\bibitem{feng2021understanding}
Ruili Feng, Deli Zhao, and Zheng-Jun Zha.
\newblock Understanding noise injection in gans.
\newblock In {\em ICML}, 2021.

\bibitem{gal2022stylegan}
Rinon Gal, Or Patashnik, Haggai Maron, Amit~H Bermano, Gal Chechik, and Daniel
  Cohen-Or.
\newblock Stylegan-nada: Clip-guided domain adaptation of image generators.
\newblock {\em ACM TOG}, 2022.

\bibitem{ha2016hypernetworks}
David Ha, Andrew Dai, and Quoc~V Le.
\newblock Hypernetworks.
\newblock In {\em ICLR}, 2016.

\bibitem{he2022gcfsr}
Jingwen He, Wu Shi, Kai Chen, Lean Fu, and Chao Dong.
\newblock Gcfsr: a generative and controllable face super resolution method
  without facial and gan priors.
\newblock In {\em CVPR}, 2022.

\bibitem{heusel2017gans}
Martin Heusel, Hubert Ramsauer, Thomas Unterthiner, Bernhard Nessler, and Sepp
  Hochreiter.
\newblock Gans trained by a two time-scale update rule converge to a local nash
  equilibrium.
\newblock In {\em NeurIPS}, 2017.

\bibitem{hu2022style}
Xueqi Hu, Qiusheng Huang, Zhengyi Shi, Siyuan Li, Changxin Gao, Li Sun, and
  Qingli Li.
\newblock Style transformer for image inversion and editing.
\newblock In {\em CVPR}, 2022.

\bibitem{harkonen2020ganspace}
Erik Härkönen, Aaron Hertzmann, Jaakko Lehtinen, and Sylvain Paris.
\newblock Ganspace: Discovering interpretable gan controls.
\newblock In {\em NeurIPS}, 2020.

\bibitem{johnson2016perceptual}
Justin Johnson, Alexandre Alahi, and Li Fei-Fei.
\newblock Perceptual losses for real-time style transfer and super-resolution.
\newblock In {\em ECCV}, 2016.

\bibitem{2018StyleGAN}
Tero Karras, Samuli Laine, and Timo Aila.
\newblock A style-based generator architecture for generative adversarial
  networks.
\newblock In {\em CVPR}, 2019.

\bibitem{karras2020analyzing}
Tero Karras, Samuli Laine, Miika Aittala, Janne Hellsten, Jaakko Lehtinen, and
  Timo Aila.
\newblock Analyzing and improving the image quality of stylegan.
\newblock In {\em CVPR}, 2020.

\bibitem{KrauseStarkDengFei-Fei_3DRR2013}
Jonathan Krause, Michael Stark, Jia Deng, and Li Fei-Fei.
\newblock 3d object representations for fine-grained categorization.
\newblock In {\em 3dRR-13}, 2013.

\bibitem{CelebAMask-HQ}
Cheng-Han Lee, Ziwei Liu, Lingyun Wu, and Ping Luo.
\newblock Maskgan: Towards diverse and interactive facial image manipulation.
\newblock In {\em CVPR}, 2020.

\bibitem{nitzan2022mystyle}
Yotam Nitzan, Kfir Aberman, Qiurui He, Orly Liba, Michal Yarom, Yossi
  Gandelsman, Inbar Mosseri, Yael Pritch, and Daniel Cohen-Or.
\newblock Mystyle: A personalized generative prior.
\newblock {\em arXiv}, 2022.

\bibitem{parmar2022spatially}
Gaurav Parmar, Yijun Li, Jingwan Lu, Richard Zhang, Jun-Yan Zhu, and
  Krishna~Kumar Singh.
\newblock Spatially-adaptive multilayer selection for gan inversion and
  editing.
\newblock In {\em CVPR}, 2022.

\bibitem{patashnik2021styleclip}
Or Patashnik, Zongze Wu, Eli Shechtman, Daniel Cohen-Or, and Dani Lischinski.
\newblock Styleclip: Text-driven manipulation of stylegan imagery.
\newblock In {\em ICCV}, 2021.

\bibitem{radford2021learning}
Alec Radford, Jong~Wook Kim, Chris Hallacy, Aditya Ramesh, Gabriel Goh,
  Sandhini Agarwal, Girish Sastry, Amanda Askell, Pamela Mishkin, Jack Clark,
  et~al.
\newblock Learning transferable visual models from natural language
  supervision.
\newblock In {\em ICML}, 2021.

\bibitem{richardson2021encoding}
Elad Richardson, Yuval Alaluf, Or Patashnik, Yotam Nitzan, Yaniv Azar, Stav
  Shapiro, and Daniel Cohen-Or.
\newblock Encoding in style: a stylegan encoder for image-to-image translation.
\newblock In {\em CVPR}, 2021.

\bibitem{roich2021pivotal}
Daniel Roich, Ron Mokady, Amit~H Bermano, and Daniel Cohen-Or.
\newblock Pivotal tuning for latent-based editing of real images.
\newblock {\em ACM TOG}, 2021.

\bibitem{shannon1959coding}
Claude~E Shannon et~al.
\newblock Coding theorems for a discrete source with a fidelity criterion.
\newblock {\em IRE Nat. Conv. Rec}, 1959.

\bibitem{shen2020interpreting}
Yujun Shen, Jinjin Gu, Xiaoou Tang, and Bolei Zhou.
\newblock Interpreting the latent space of gans for semantic face editing.
\newblock In {\em CVPR}, 2020.

\bibitem{shen2020interfacegan}
Yujun Shen, Ceyuan Yang, Xiaoou Tang, and Bolei Zhou.
\newblock Interfacegan: Interpreting the disentangled face representation
  learned by gans.
\newblock {\em IEEE TPAMI}, 2020.

\bibitem{song2022editing}
Haorui Song, Yong Du, Tianyi Xiang, Junyu Dong, Jing Qin, and Shengfeng He.
\newblock Editing out-of-domain gan inversion via differential activations.
\newblock In {\em ECCV}, 2022.

\bibitem{Yang2021GPEN}
Xuansong~Xie Tao~Yang, Peiran~Ren and Lei Zhang.
\newblock Gan prior embedded network for blind face restoration in the wild.
\newblock In {\em CVPR}, 2021.

\bibitem{tishby2015deep}
Naftali Tishby and Noga Zaslavsky.
\newblock Deep learning and the information bottleneck principle.
\newblock In {\em itw}, 2015.

\bibitem{tov2021designing}
Omer Tov, Yuval Alaluf, Yotam Nitzan, Or Patashnik, and Daniel Cohen-Or.
\newblock Designing an encoder for stylegan image manipulation.
\newblock {\em ACM TOG}, 2021.

\bibitem{wang2021high}
Tengfei Wang, Yong Zhang, Yanbo Fan, Jue Wang, and Qifeng Chen.
\newblock High-fidelity gan inversion for image attribute editing.
\newblock In {\em CVPR}, 2021.

\bibitem{wang2021towards}
Xintao Wang, Yu Li, Honglun Zhang, and Ying Shan.
\newblock Towards real-world blind face restoration with generative facial
  prior.
\newblock In {\em CVPR}, 2021.

\bibitem{wang2004image}
Zhou Wang, Alan~C. Bovik, Hamid~R. Sheikh, and Eero~P. Simoncelli.
\newblock Image quality assessment: from error visibility to structural
  similarity.
\newblock {\em IEEE TIP}, 2004.

\bibitem{wei2022e2style}
Tianyi Wei, Dongdong Chen, Wenbo Zhou, Jing Liao, Weiming Zhang, Lu Yuan, Gang
  Hua, and Nenghai Yu.
\newblock E2style: Improve the efficiency and effectiveness of stylegan
  inversion.
\newblock {\em TIP}, 2022.

\bibitem{wu2021stylespace}
Zongze Wu, Dani Lischinski, and Eli Shechtman.
\newblock Stylespace analysis: Disentangled controls for stylegan image
  generation.
\newblock In {\em CVPR}, 2021.

\bibitem{yang2021gan}
Tao Yang, Peiran Ren, Xuansong Xie, and Lei Zhang.
\newblock Gan prior embedded network for blind face restoration in the wild.
\newblock In {\em CVPR}, 2021.

\bibitem{xuyao2022}
Xu Yao, Alasdair Newson, Yann Gousseau, and Pierre Hellier.
\newblock A style-based gan encoder for high fidelity reconstruction of images
  and videos.
\newblock In {\em ECCV}, 2022.

\bibitem{zhang2018perceptual}
Richard Zhang, Phillip Isola, Alexei~A Efros, Eli Shechtman, and Oliver Wang.
\newblock The unreasonable effectiveness of deep features as a perceptual
  metric.
\newblock In {\em CVPR}, 2018.

\bibitem{zhu2022blind}
Feida Zhu, Junwei Zhu, Wenqing Chu, Xinyi Zhang, Xiaozhong Ji, Chengjie Wang,
  and Ying Tai.
\newblock Blind face restoration via integrating face shape and generative
  priors.
\newblock In {\em CVPR}, 2022.

\end{thebibliography}
}

\end{document}